\documentclass[10pt,twocolumn,letterpaper]{article}

\usepackage[pagenumbers]{cvpr}      %

\usepackage{booktabs}
\usepackage{multirow}
\usepackage{array}
\usepackage{colortbl}
\usepackage{placeins}
\newcolumntype{C}[1]{>{\centering\arraybackslash}p{#1}}
\definecolor{groupgray}{gray}{0.93}
\definecolor{floorgray}{gray}{0.45}
\definecolor{ourpink}{HTML}{D6337F}
\definecolor{supp}{HTML}{F9E8EF}
\newcommand{\pos}[1]{\textcolor{ourpink}{#1}}

\newcommand{\tms}[1]{{\scriptsize\textcolor{floorgray}{#1}}}

\extrafloats{100}
\definecolor{deltablue}{HTML}{3C78A8}
\definecolor{rowblue}{HTML}{EAF2F8}   %
\newcommand{\hd}[1]{\begin{tabular}[c]{@{}c@{}}#1\end{tabular}}
\newcommand{\dspd}[1]{{\scriptsize\textcolor{ourpink}{#1$\times$}}}
\newcommand{\spd}[1]{{\scriptsize\textcolor{floorgray}{#1$\times$}}}
\newcommand{\sd}[1]{{\tiny\textcolor{floorgray}{$\pm$#1}}}
\newcommand{\vsd}[2]{\vtop{\baselineskip=4.9pt\lineskiplimit=-\maxdimen\lineskip=0pt\halign{\hfil##\hfil\cr#1\cr{\tiny\textcolor{floorgray}{$\pm$#2}}\cr}\kern0.8pt}}   %
\newcommand{\kk}[1]{{\scriptsize\textcolor{floorgray}{#1}}}
\newcommand{\kkc}[1]{\makebox[\linewidth][c]{\kk{#1}}}

\definecolor{cvprblue}{rgb}{0.21,0.49,0.74}
\usepackage[pagebackref,breaklinks,colorlinks,allcolors=cvprblue]{hyperref}

\def\confName{CVPR}
\def\confYear{2027}

\title{DecoupAttn: Decoupled Sparse Attention\\ for Training-Free Acceleration of Video Diffusion}

\author{Yanjin Li$^{1}$\quad Bingbing Qiu$^{1}$\quad Qiang Huang$^{2}$\quad Rui Qian$^{2}$\\Yanfeng Li$^{3}$\quad Jian Wang$^{2}$\quad Xiaoying Tang$^{4}$\quad Mingda Zhang$^{4,*}$\\[3pt]
{\footnotesize $^{1}$University of Illinois Urbana-Champaign\quad $^{2}$Fudan University\quad $^{3}$Macao Polytechnic University}\\
{\footnotesize $^{4}$The Chinese University of Hong Kong, Shenzhen\quad $^{*}$Corresponding author: \texttt{mingdazhang@ieee.org}}}

\makeatletter
\def\@maketitle{
   \newpage
   \null
   \vskip 9.53pt
   \begin{center}
      {\Large \bf \@title \par}
      \vspace*{9pt}{
        \large
        \lineskip .5em
        \begin{tabular}[t]{c}\@author\end{tabular}
        \par
      }
      \vspace*{4.03pt}
   \end{center}
}
\makeatother
\hypersetup{pdftitle={DecoupAttn: Decoupled Sparse Attention for Training-Free Acceleration of Video Diffusion},
  pdfauthor={Yanjin Li, Bingbing Qiu, Qiang Huang, Rui Qian, Yanfeng Li, Jian Wang, Xiaoying Tang, Mingda Zhang}}

\renewcommand{\dbltopfraction}{0.85}
\begin{document}
\maketitle
\begin{abstract}
Sparse attention accelerates few-step video diffusion transformers by computing a few key blocks exactly and compensating for the rest.
However, existing sparse attention still faces key challenges, including \emph{identification overhead} from finding critical blocks at every denoising step, \emph{suboptimal budget allocation} that sets sparsity budgets uniformly or by local metrics, and \emph{cross-step error accumulation} as reused masks or clusters go stale.
Observing that the errors left on skipped blocks cancel each other once every key is compensated, we propose DecoupAttn, a new paradigm of decoupled sparse attention: semantic clusters compensate every key, and exact attention corrects their error on a few blocks fixed once offline by position.
To support this paradigm, we introduce Distortion Budget, a rate--distortion objective for budget allocation that places exact computation across timesteps, layers, and heads where it removes the most distortion that compensation leaves in the final latent.
Built on this budget, we further propose Reassign Cache, in which each step reassigns the drifting keys to cached centroids and refreshes compensation while the budgeted blocks stay fixed.
Moreover, DecoupAttn is training-free: a few calibration clips compile one plan offline, and inference runs standard block-sparse kernels with one fused compensation kernel.
Our code will be released after submission.
\end{abstract}

\begin{figure}[t]
\centering
\includegraphics[width=\linewidth]{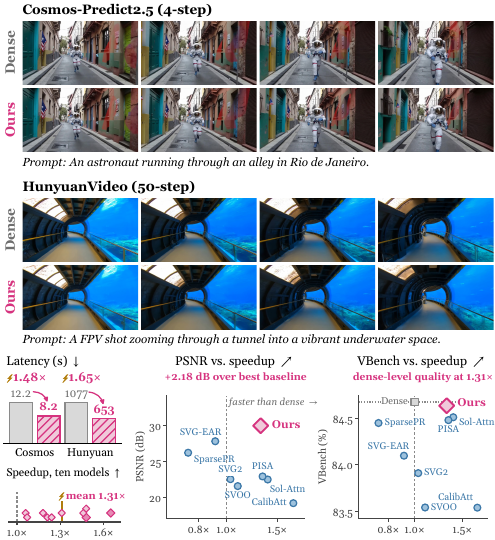}
\caption{\textbf{DecoupAttn keeps dense-attention quality at a lower latency.}
Frames from dense attention and DecoupAttn on Cosmos-Predict2.5 (4 steps) and HunyuanVideo (50 steps), with latency, speedup on all ten models (filled: these two), and PSNR and VBench against speedup for DecoupAttn and the seven baselines of Table~\ref{tab:main}, averaged over the ten models.}
\label{fig:teaser}
\end{figure}

\begin{figure*}[t]
\centering
\includegraphics[width=\linewidth]{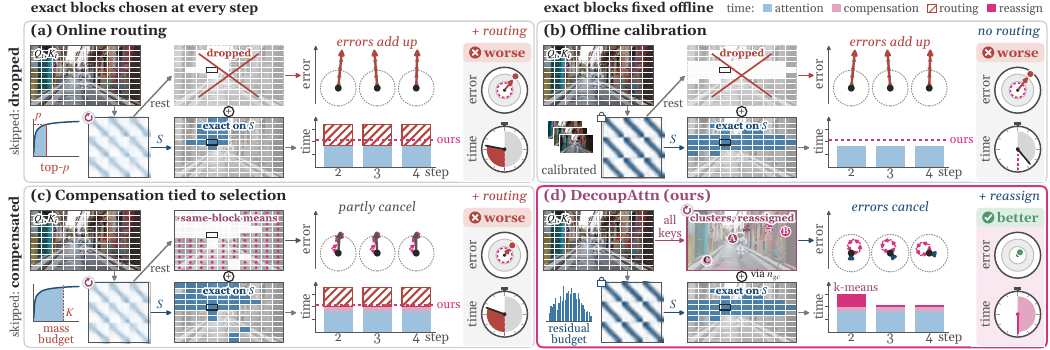}
\caption{\textbf{Three lines of sparse attention and DecoupAttn.}
(a) Online routing re-picks exact blocks each step and drops the rest, (b) offline calibration fixes them but drops the rest, and (c) compensation reuses them and reroutes every step.
(d) DecoupAttn fixes them offline and compensates every key by semantic clusters.
Right: schematic error and time against DecoupAttn (dashed), excess in red.}
\label{fig:paradigm}
\end{figure*}

\section{Introduction}
\label{sec:intro}

In recent years, video diffusion transformers have been applied to a wide range of generation tasks, from text-to-video synthesis~\cite{wan2025wan} to controlled generation~\cite{nvidia2025cosmos} and neural rendering~\cite{liang2025diffusionrenderer}, gradually moving from many-step samplers toward few-step distilled generators~\cite{yin2025causvid,huang2025selfforcing}.
Sparse attention makes these generators fast without retraining: since attention grows quadratically with video length, it computes a few key blocks exactly and compensates for the rest, keeping near-dense fidelity (Figure~\ref{fig:teaser}).
However, existing methods tie compensation to block selection: the groups that pick the exact blocks also summarize the skipped keys, so compensation is costly at every step of a few-step sampler and impossible once the selection is fixed offline.

Sparse attention for video generation has followed three main lines of work, as shown in Figure~\ref{fig:paradigm}.
First, online routing (dynamic sparse attention) identifies critical blocks at every step from the current queries and keys, by scoring pooled blocks~\cite{zhang2025spargeattn,xu2025xattention} or clustering tokens by semantics~\cite{yang2025svg2,tan2026adacluster}, treating sparsity as a property of each input.
Second, offline calibration (static sparse attention) fixes sparse patterns before deployment, measured on a few prompts~\cite{yehezkel2026calibatt} or derived from how attention decays with distance~\cite{li2025radial}, treating sparsity as a property of the model.
Third, compensation approximates the skipped blocks instead of dropping them, from cluster centroids~\cite{zhou2026svgear}, block-wise Taylor expansions~\cite{li2026pisa}, or fitted residuals~\cite{taghavi2026sparsepr}, recovering much of the attention that selection leaves out.

However, these methods still face three challenges.
\textbf{(i) Identification overhead.}
Online routing and the compensation built on its groups identify critical blocks at every denoising step~\cite{zhang2025spargeattn,xu2025xattention,xi2025svg,yang2025svg2,tan2026adacluster,li2026solattn,zhao2026scope,zhou2026svgear,li2026pisa,zhang2026pasa,xu2026recap,liu2025rectified,taghavi2026sparsepr}, an overhead that can offset the gains of sparsity in few-step samplers, while static patterns fixed in advance~\cite{yehezkel2026calibatt,li2025radial,zhang2025sta} discard the remaining blocks and incur information loss.
\textbf{(ii) Suboptimal budget allocation.}
Existing methods set sparsity budgets across heads, layers, and timesteps uniformly or by local metrics such as retained attention mass~\cite{yehezkel2026calibatt,yang2025svg2,luo2026svoo,zhang2026pasa,zhou2026svgear}, or by the effect of dropping blocks on the output~\cite{zheng2026heart,zhou2026scalingattention,xia2026allot}.
Retained attention mass, however, does not determine the output error~\cite{taghavi2026sparsepr}, which depends on the residual left after compensation and on how unevenly later layers and steps amplify it, so exact computation is spent where it barely changes the final video.
\textbf{(iii) Cross-step error accumulation.}
To avoid redundant recomputation, recent methods reuse masks, clusters, or earlier centroids across steps, or predict masks ahead~\cite{shmilovich2025liteattention,liu2026losa,liu2026moddit,yang2025svg2,lee2026hypervattention,luo2026svoo,chen2026mcsparse,xu2026recap}, but attention patterns evolve across denoising steps, so the reused groups go stale and their errors accumulate, while refreshing them repeats the identification.

Our key observation is that once every key is compensated, the residuals of the skipped blocks point in many directions and cancel each other like dithered quantization errors~\cite{schuchman1964dither}, so blocks fixed once offline match blocks re-selected at every step.
Based on this observation, we propose DecoupAttn, a new paradigm of decoupled sparse attention: semantic clusters compensate every key, and exact attention corrects their residual on a few blocks fixed once offline by position (Figure~\ref{fig:paradigm}d).
Selecting blocks costs nothing at inference, and the residual of every fixed block can be priced offline.
Few-step samplers amplify this residual unevenly, so we introduce Distortion Budget, a rate--distortion objective for budget allocation that prices every timestep, layer, and head by the distortion the residual leaves in the final latent.
Built on this budget, we further propose Reassign Cache, which limits the growth of these residuals by reassigning the drifting keys to cached centroids at every step, while the blocks and the budget stay fixed.
DecoupAttn thus moves block identification and budget allocation into one plan compiled offline with no training, and inference only refreshes the clusters, running standard block-sparse kernels and one fused compensation kernel.

We evaluate DecoupAttn on ten video generation, controlled generation, and neural rendering models, where it outperforms online-routing, offline-calibrated, and compensation-based methods on nine of them and delivers a substantial speedup over dense attention.

\section{Related Work}
\label{sec:related}

\paragraph{Summarizing skipped attention.}
Transformers summarize attention by clusters, near- and far-field splits, or sparse-plus-low-rank estimators~\cite{vyas2020clustered,nguyen2021fmmformer,roy2021routing,li2026momentkv,chen2021scatterbrain,zheng2023eva}, and video sparse attention compensates the skipped keys by cluster or block means~\cite{zhou2026svgear,liu2025rectified}, Taylor terms~\cite{li2026pisa,zhang2026pasa}, or fitted or cached residuals~\cite{taghavi2026sparsepr,xu2026recap,chen2026mcsparse}.
Masks frozen early or predicted ahead~\cite{liu2026losa,liu2026moddit} drop the keys outside them, and routers that reuse centroids~\cite{yang2025svg2,lee2026hypervattention,luo2026svoo} still let the clusters pick the blocks.
DecoupAttn summarizes the skipped keys by semantic clusters that never pick blocks.

\paragraph{Allocating computation by sensitivity.}
Coders and quantizers allocate bits by rate--distortion cost or layer sensitivity~\cite{shoham1988bitallocation,sullivan1998rdo,dong2019hawq}, and sparse attention sets density by retained attention mass, profiling, sampler curvature, or block error~\cite{yehezkel2026calibatt,yang2025svg2,luo2026svoo,zhang2026pasa,zhou2026svgear}, or by each unit's effect on the output~\cite{zheng2026heart,zhou2026scalingattention,xia2026allot}.
Distortion Budget instead prices the residual left after compensation, not the error of dropping a block, by its effect on the final latent.

\section{Preliminaries}
\label{sec:prelim}

\noindent\phantomsection\label{res:def1}\textbf{Definition 1: Block-sparse attention call.}
A self-attention call at denoising step $t$, layer $l$, and head $h$ maps the queries $q_i$, keys $k_j$, and values $v_j$ of its $L$ tokens to outputs $y_i$, with the tokens partitioned by position into query tiles $g$ and key blocks $b$, the units of block-sparse kernels~\cite{ye2025flashinfer}:
\begin{equation}
y_i=\frac{N_i}{Z_i}=\frac{\sum_b N_{ib}}{\sum_b Z_{ib}},\quad N_{ib}=\sum_{j\in b}e^{s_{ij}}v_j,\quad Z_{ib}=\sum_{j\in b}e^{s_{ij}},
\label{eq:dense}
\end{equation}
where $s_{ij}=q_i^\top k_j/\sqrt d$ and $d$ is the head dimension.
A sparse call computes $N_{ib}$ and $Z_{ib}$ exactly on a block set $\mathcal S_g$ and summarizes the rest.
Existing compensation writes $\hat N_i=\sum_{b\in\mathcal S_g}N_{ib}+\sum_{b\notin\mathcal S_g}\tilde N_{ib}$ with $\tilde N_{ib}$ from the statistics of block $b$ itself, and every change of $\mathcal S_g$ rebuilds the compensation terms~\cite{li2026pisa,zhou2026svgear}.

\noindent\phantomsection\label{res:def2}\textbf{Definition 2: Sparse attention configuration.}
A sampler runs $T$ denoising steps.
Each head of each call is a unit $u=(t,l,h)$ of the unit set $\mathcal U$, and the heads of a call $(t,l)$ share one compensation estimator.
A configuration, or plan, $\pi=(\{o_u\}_{u\in\mathcal U},\{\eta_{tl}\})$ assigns every unit an option and every call an estimator, which determines the final latent and the density:
\begin{equation}
\pi\;\Rightarrow\;\hat x_0(\pi),\qquad X(\pi)=\frac{\sum_u w_uX_u(o_u)}{\sum_u w_u},
\label{eq:config}
\end{equation}
where $o_u\in\mathcal O$ is the unit's option (Eq.~\eqref{eq:options}), $\eta_{tl}\in\{\textsc{Full},\textsc{Centroid}\}$ the call's estimator (Eq.~\eqref{eq:comp}), $X_u$ the density of an option, the fraction of blocks it executes with every block padded to the kernel block size and the \textsc{Dense} option counted as one, and $w_u$ the number of sampler steps the unit represents.
The first step and the first layer always stay dense and lie outside $\mathcal U$, and the estimator adds a fixed overhead to $X$ (Appendix~\ref{app:th-cost}).

\noindent\textbf{Problem statement.}
Given a pretrained video diffusion transformer with frozen weights and a few calibration clips, we seek a plan that is fixed before deployment, that keeps the density within a target $\rho$, and that keeps the final latent closest to the dense one:
\begin{equation}
\pi^\star=\arg\min_{\pi}\;\mathbb E_x\big\|\hat x_0(\pi)-x_0\big\|_2^2\;\;\text{s.t.}\;\;X(\pi)\le\rho,
\label{eq:problem}
\end{equation}
where $x$ is the input (prompt, conditions, and seed) and $x_0$ the dense final latent for the same input.

\section{Method}
\label{sec:method}

\begin{figure*}[t]
\centering
\includegraphics[width=\linewidth]{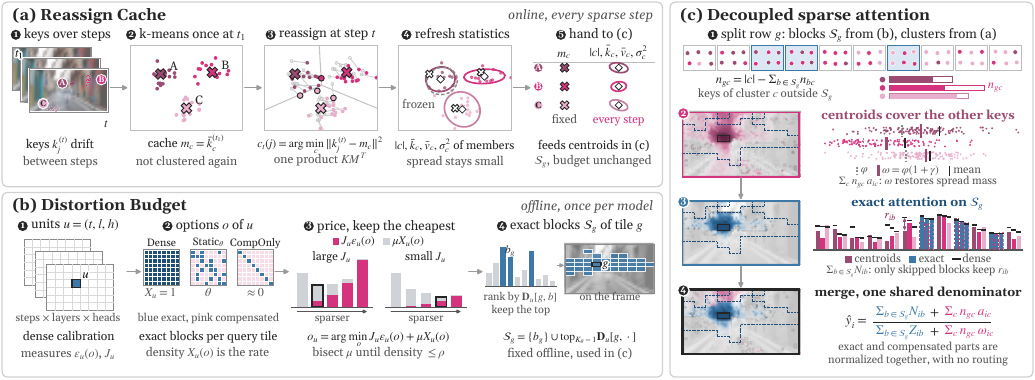}
\caption{\textbf{How DecoupAttn computes one attention call} (Cosmos-Predict2.5).
(a) Reassign Cache clusters the keys once, caches the centroids, and at every later step moves each key to its nearest cached centroid and refreshes the cluster statistics, (b) Distortion Budget prices the options of every unit offline and keeps the cheapest, which fixes the exact blocks \(\mathcal S_g\), and (c) exact attention on \(\mathcal S_g\) and centroid compensation of every other key meet only through the counts \(n_{gc}\) and share one softmax denominator (pink: approximation, blue: exact).}
\label{fig:method}
\end{figure*}

As Figure~\ref{fig:method} shows, DecoupAttn centers on one quantity, the residual that compensation leaves on each key block: the decoupled estimator makes this residual the only error that reaches the output and keeps it small (Section~\ref{sec:decoup}), Distortion Budget spends exact computation where the residual most distorts the final latent (Section~\ref{sec:budget}), and Reassign Cache limits its growth as keys drift (Section~\ref{sec:reassign}).

\subsection{Decoupled Sparse Attention}
\label{sec:decoup}

To keep compensation when the selection is fixed offline, decoupled sparse attention follows the difference estimators of survey sampling~\cite{sarndal1992model}: a cheap algorithm predicts the contribution of every key, and exact attention on a few blocks replaces the prediction there (Figure~\ref{fig:method}c).

\noindent\textbf{Centroid compensation with two corrections.}
At the first sparse step, the keys of every head are clustered by k-means into clusters $c$, and every later call summarizes each cluster by its centroid with two second-order corrections computed from the cluster's current members:
\begin{equation}
\omega_{ic}=\phi_{ic}\,(1+\gamma_{ic}),\qquad a_{ic}=\omega_{ic}\,\bar v_c+\phi_{ic}\,\frac{Hq_i}{\sqrt d},
\label{eq:comp}
\end{equation}
where $\phi_{ic}=e^{q_i^\top\bar k_c/\sqrt d}$ is the centroid weight, and $\bar k_c$, $\bar v_c$, and $\sigma_c^2$ are the cluster's mean key, mean value, and per-dimension key variance.
The \emph{variance correction} $\gamma_{ic}=\frac{1}{2d}\sum_p q_{ip}^2\sigma_{cp}^2$ is the diagonal part of the second-order term of $e^{q_i^\top k/\sqrt d}$ around $\bar k_c$ and restores this part of the mass that a centroid misses on a spread-out cluster (Lemmas~\hyperref[res:A3]{A.3} and~\hyperref[res:A4]{A.4}).
The within-cluster cross-covariance $H=\frac1L\big(\sum_j v_jk_j^\top-\sum_c|c|\,\bar v_c\bar k_c^\top\big)$, pooled over the clusters of a head, gives their shared \emph{first-order value correction} (\hyperref[res:A5]{Lemma~A.5}).
The \textsc{Centroid} estimator drops both corrections.

\noindent\textbf{Decoupled estimator.}
Each query tile $g$ computes its exact blocks $\mathcal S_g$ in full and covers every other key with its cluster centroid, and both parts share one softmax denominator:
\begin{equation}
\hat y_i=\frac{\hat N_i}{\hat Z_i}=\frac{\sum_{b\in\mathcal S_g}N_{ib}+\sum_c n_{gc}\,a_{ic}}{\sum_{b\in\mathcal S_g}Z_{ib}+\sum_c n_{gc}\,\omega_{ic}},
\label{eq:decoup}
\end{equation}
where $a_{ic}$ and $\omega_{ic}$ are what one member of cluster $c$ contributes to the numerator and the denominator (Eq.~\eqref{eq:comp}).
The count $n_{gc}=|c|-\sum_{b\in\mathcal S_g}n_{bc}$, with $n_{bc}=|b\cap c|$, keeps only the members of $c$ that the exact blocks do not already cover.
The exact blocks are fixed offline from position and calibration statistics, the clusters come from the keys alone, and the counts $n_{gc}$ are the only place the two meet, so no block identification runs at inference.

\noindent\textbf{Compensate every key, correct some blocks.}
Substituting $n_{gc}$ into Eq.~\eqref{eq:decoup} splits the estimate into two steps.
First, the centroids compensate all keys as if no block were exact.
Then, on each exact block, the exact value replaces the centroids' prediction, as in Scatterbrain and EVA~\cite{chen2021scatterbrain,zheng2023eva}:
\begin{equation}
\begin{gathered}
\hat N_i=\sum_c |c|\,a_{ic}+\sum_{b\in\mathcal S_g}\Delta N_{ib},\\
\hat y_i-y_i=-\frac{1}{\hat Z_i}\sum_{b\notin\mathcal S_g}\underbrace{\big(\Delta N_{ib}-y_i\,\Delta Z_{ib}\big)}_{r_{ib}},
\end{gathered}
\label{eq:residual}
\end{equation}
where $\Delta N_{ib}=N_{ib}-\sum_c n_{bc}a_{ic}$ and $\Delta Z_{ib}=Z_{ib}-\sum_c n_{bc}\omega_{ic}$ are the gaps between exact value and centroid prediction on block $b$, the residual $r_{ib}$ is the error block $b$ passes to the output, and skipped blocks leave error.

\noindent\textbf{Why fixed blocks suffice.}
By Eq.~\eqref{eq:residual}, the error is a sum of block residuals, so its size depends on the estimated mass, on how large the residuals are, and on how they align:
\begin{equation}
\|\hat y_i-y_i\|_2^2=\frac{1}{\hat Z_i^2}\Big\|\sum_{b\notin\mathcal S_g}r_{ib}\Big\|_2^2=\frac{\kappa_iR_i}{\hat Z_i^2},
\label{eq:coherence}
\end{equation}
where $R_i=\sum_{b\notin\mathcal S_g}\|r_{ib}\|_2^2$ is the residual energy of the skipped blocks, and the coherence ratio $\kappa_i=\|\sum_{b\notin\mathcal S_g}r_{ib}\|_2^2/R_i$, set to one when $R_i=0$, is one for mutually orthogonal residuals and grows as they align.
Dropped blocks each remove their whole attention and push the output the same way, whereas a compensated residual comes from its members' deviations from their centroids, which sum to zero over the cluster, and the residuals of the skipped blocks cancel each other like dithered quantization errors~\cite{schuchman1964dither} (\hyperref[res:A10]{Theorem~A.10}).
With $\hat Z_i\approx Z_i$ (\hyperref[res:A2]{Corollary~A.2}) and $\kappa_i$ kept small, the residual energy $R_i$ is what Sections~\ref{sec:budget} and~\ref{sec:reassign} then control.

\noindent\phantomsection\label{res:prop1}\textbf{Proposition 1.}
\emph{For count-based compensation, decoupled sparse attention equals compensating every key and correcting the residual of each exact block, and its error is determined by the summed residual of the other blocks for any blocks and clusters, while for blocks fixed offline the deviation parts of these residuals add like random errors and nearly cancel each other.}
(Proof in Appendices~\ref{app:prop1} and~\ref{app:th-coh}, evidence in Section~\ref{sec:exp-dither}.)

\subsection{Distortion Budget}
\label{sec:budget}

Section~\ref{sec:decoup} leaves the residual energy $R_i$ to control, and by \hyperref[res:prop1]{Proposition~1} a block's residual does not depend on which other blocks are exact (Appendix~\ref{app:prop1}).
Distortion Budget measures each residual once on the calibration set and solves Eq.~\eqref{eq:problem} as a rate--distortion bit allocation~\cite{shoham1988bitallocation}: units act as quantizers, options as their resolutions, density as the rate, and final-latent error as the distortion (Figure~\ref{fig:method}b).

\noindent\textbf{Options.}
Each unit $u=(t,l,h)$ (\hyperref[res:def2]{Definition~2}) picks one of three kinds of option, which differ in how many key blocks it computes exactly.
\textsc{Dense} computes every block, as dense attention does.
\textsc{Static}$_\theta$ computes the fraction $\theta$ of key blocks that rank highest for the query tile and compensates the rest.
\textsc{CompOnly} computes only the diagonal block and compensates everything else, the analog of the skip mode in rate--distortion mode decision~\cite{sullivan1998rdo}.
The option set and the exact blocks of \textsc{Static}$_\theta$ are
\begin{equation}
\begin{gathered}
\mathcal O=\{\textsc{Dense},\textsc{CompOnly}\}\cup\{\textsc{Static}_\theta\}_{\theta\in\Theta},\\
\mathcal S_g(\textsc{Static}_\theta)=\{b_g\}\cup\mathrm{top}_{K_\theta-1}\big(\mathbf D_u[g,\cdot]\big),
\end{gathered}
\label{eq:options}
\end{equation}
where $\Theta$ is the set of density levels (Appendix~\ref{app:config}), $K_\theta=\max(1,\lfloor\theta N_B\rceil)$ for $N_B$ key blocks, $\mathbf D_u[g,\cdot]$ the ranking of the other key blocks for query tile $g$ (Eq.~\eqref{eq:dtable}), and $b_g$ its diagonal block, which every option computes.
The ranking orders these blocks by how badly compensation handles them, measured by their residual in Eq.~\eqref{eq:residual} on the calibration set with the model run densely:
\begin{equation}
\mathbf D_u[g,b]=\mathbb E_x\,\mathbb E_{i\in g}\,\frac{\|r_{ib}\|_2^2}{Z_i^2\,\|y_i\|_2^2},
\label{eq:dtable}
\end{equation}
where $r_{ib}$ is taken under full compensation, and dividing by $Z_i^2\|y_i\|_2^2$ puts large and small outputs on one scale.

\noindent\textbf{Distortion of an option.}
Distortion Budget measures, unit by unit on dense calibration runs, the projected error of the head under every option $o$ and estimator $\eta$:
\begin{equation}
\begin{gathered}
\varepsilon_u(o,\eta)=\mathbb E_x\big\langle M_h,\,(\hat O_u-O_u)^\top(\hat O_u-O_u)\big\rangle,\\
M_h=W_{o,h}^\top W_{o,h},
\end{gathered}
\label{eq:distortion}
\end{equation}
where $O_u$ and $\hat O_u$ are the head's dense output and its output under $o$ and $\eta$ (Eq.~\eqref{eq:decoup}), the expectation runs over calibration inputs $x$, the inner product is the squared error after projection, $\|(\hat O_u-O_u)W_{o,h}^\top\|_F^2$, and $W_{o,h}$ is the head's slice of the output projection into the residual stream.

\begin{table*}[t]
\renewcommand{\vsd}[2]{\vtop{\baselineskip=4.6pt\lineskiplimit=-\maxdimen\lineskip=0pt\halign{\hfil##\hfil\cr#1\cr{\tiny\textcolor{floorgray}{$\pm$#2}}\cr}\kern0.3pt}}%
\centering
\footnotesize
\renewcommand{\arraystretch}{0.68}
\begin{tabular}{@{}>{\hspace{\tabcolsep}}l@{\hspace{3pt}}C{0.68cm}@{\hspace{2pt}}C{0.62cm}@{\hspace{5.5pt}}C{0.68cm}@{\hspace{2pt}}C{0.62cm}@{\hspace{5.5pt}}C{0.68cm}@{\hspace{2pt}}C{0.62cm}@{\hspace{5.5pt}}C{0.68cm}@{\hspace{2pt}}C{0.62cm}@{\hspace{5.5pt}}C{0.68cm}@{\hspace{2pt}}C{\dimexpr0.62cm+1.7pt\relax}@{\hspace{5.5pt}}C{0.68cm}@{\hspace{2pt}}C{0.62cm}@{\hspace{5.5pt}}C{0.68cm}@{\hspace{2pt}}C{0.62cm}@{\hspace{5.5pt}}C{0.68cm}@{\hspace{2pt}}C{0.62cm}@{\hspace{5.5pt}}>{\columncolor{rowblue}[2.75pt][1pt]}C{0.68cm}@{\hspace{2pt}}>{\columncolor{rowblue}[1pt][2.75pt]}C{0.62cm}@{\hspace{5.5pt}}C{0.87cm}@{\hspace{2pt}}C{0.87cm}@{\hspace{6pt}}}
\toprule
Model & \multicolumn{2}{c}{\textcolor{floorgray}{Floor}} & \multicolumn{2}{c}{CalibAtt} & \multicolumn{2}{c}{SVOO} & \multicolumn{2}{c}{SVG2} & \multicolumn{2}{c}{SVG-EAR} & \multicolumn{2}{c}{Sol-Attn} & \multicolumn{2}{c}{PISA} & \multicolumn{2}{c}{SparsePR} & \multicolumn{2}{c}{\textbf{Ours}} & \multicolumn{2}{c}{$\Delta$}\\
\cmidrule(lr){2-3}\cmidrule(lr){4-5}\cmidrule(lr){6-7}\cmidrule(lr){8-9}\cmidrule(lr){10-11}\cmidrule(lr){12-13}\cmidrule(lr){14-15}\cmidrule(lr){16-17}\cmidrule(lr){18-19}\cmidrule(lr){20-21}
\kk{(steps)} & \kkc{PSNR} & \kkc{$\times$} & \kkc{PSNR} & \kkc{$\times$} & \kkc{PSNR} & \kkc{$\times$} & \kkc{PSNR} & \kkc{$\times$} & \kkc{PSNR} & \kkc{$\times$} & \kkc{PSNR} & \kkc{$\times$} & \kkc{PSNR} & \kkc{$\times$} & \kkc{PSNR} & \kkc{$\times$} & \kkc{PSNR} & \kkc{$\times$} & \kkc{PSNR} & \kkc{$\times$}\\[-1.5pt]
 & \kkc{SSIM} & \kkc{LPIPS} & \kkc{SSIM} & \kkc{LPIPS} & \kkc{SSIM} & \kkc{LPIPS} & \kkc{SSIM} & \kkc{LPIPS} & \kkc{SSIM} & \kkc{LPIPS} & \kkc{SSIM} & \kkc{LPIPS} & \kkc{SSIM} & \kkc{LPIPS} & \kkc{SSIM} & \kkc{LPIPS} & \kkc{SSIM} & \kkc{LPIPS} & \kkc{SSIM} & \kkc{LPIPS}\\
\midrule
CP1\rule{0pt}{7pt} & \vsd{\textcolor{floorgray}{32.66}}{7.17} & \spd{1.00} & \vsd{19.35}{2.46} & \spd{1.73} & \vsd{21.27}{3.70} & \spd{1.35} & \vsd{22.03}{3.51} & \spd{1.34} & \vsd{24.32}{4.27} & \spd{1.18} & \vsd{21.87}{3.63} & \spd{1.47} & \vsd{22.27}{4.05} & \spd{1.43} & \vsd{24.51}{4.31} & \spd{1.06} & \vsd{\textbf{27.77}}{4.60} & \spd{1.48} & \pos{+5.54} & \dspd{+0.11}\\[4pt]
\kk{(35)} & \vsd{\textcolor{floorgray}{0.897}}{0.100} & \vsd{\textcolor{floorgray}{0.072}}{0.071} & \vsd{0.668}{0.127} & \vsd{0.397}{0.074} & \vsd{0.717}{0.136} & \vsd{0.297}{0.097} & \vsd{0.732}{0.136} & \vsd{0.266}{0.099} & \vsd{0.765}{0.130} & \vsd{0.181}{0.080} & \vsd{0.732}{0.133} & \vsd{0.252}{0.092} & \vsd{0.724}{0.143} & \vsd{0.260}{0.099} & \vsd{0.766}{0.131} & \vsd{0.183}{0.089} & \vsd{\textbf{0.835}}{0.090} & \vsd{\textbf{0.121}}{0.056} & \pos{+0.106} & \pos{$-$0.141}\\[3.5pt]
CP2.5\rule{0pt}{7pt} & \vsd{\textcolor{floorgray}{27.78}}{3.90} & \spd{1.00} & \vsd{12.35}{1.48} & \spd{2.44} & \vsd{17.78}{2.57} & \spd{1.08} & \vsd{19.74}{2.85} & \spd{0.98} & \vsd{21.85}{3.32} & \spd{0.90} & \vsd{17.95}{2.65} & \spd{1.50} & \vsd{17.68}{2.48} & \spd{1.40} & \vsd{20.97}{3.00} & \spd{0.60} & \vsd{\textbf{24.70}}{3.78} & \spd{1.48} & \pos{+6.37} & \dspd{+0.21}\\[4pt]
\kk{(4)} & \vsd{\textcolor{floorgray}{0.921}}{0.053} & \vsd{\textcolor{floorgray}{0.054}}{0.030} & \vsd{0.400}{0.107} & \vsd{0.584}{0.071} & \vsd{0.594}{0.127} & \vsd{0.381}{0.099} & \vsd{0.693}{0.104} & \vsd{0.225}{0.065} & \vsd{0.781}{0.096} & \vsd{0.157}{0.066} & \vsd{0.600}{0.128} & \vsd{0.314}{0.067} & \vsd{0.615}{0.116} & \vsd{0.313}{0.070} & \vsd{0.744}{0.092} & \vsd{0.186}{0.057} & \vsd{\textbf{0.855}}{0.077} & \vsd{\textbf{0.102}}{0.046} & \pos{+0.223} & \pos{$-$0.207}\\[3.5pt]
Wan2.1\rule{0pt}{7pt} & \vsd{\textcolor{floorgray}{24.93}}{3.03} & \spd{1.00} & \vsd{10.56}{1.31} & \spd{1.54} & \vsd{19.27}{1.56} & \spd{0.81} & \vsd{18.56}{1.56} & \spd{0.72} & \vsd{22.95}{2.16} & \spd{0.67} & \vsd{18.32}{1.63} & \spd{1.31} & \vsd{17.87}{1.50} & \spd{1.16} & \vsd{20.29}{1.81} & \spd{0.40} & \vsd{\textbf{24.70}}{2.29} & \spd{1.06} & \pos{+6.44} & \dspd{+0.12}\\[4pt]
\kk{(4)} & \vsd{\textcolor{floorgray}{0.902}}{0.049} & \vsd{\textcolor{floorgray}{0.050}}{0.028} & \vsd{0.312}{0.074} & \vsd{0.589}{0.067} & \vsd{0.726}{0.069} & \vsd{0.150}{0.036} & \vsd{0.696}{0.073} & \vsd{0.171}{0.043} & \vsd{0.858}{0.054} & \vsd{0.074}{0.027} & \vsd{0.685}{0.070} & \vsd{0.175}{0.038} & \vsd{0.669}{0.071} & \vsd{0.191}{0.037} & \vsd{0.773}{0.066} & \vsd{0.122}{0.034} & \vsd{\textbf{0.889}}{0.044} & \vsd{\textbf{0.057}}{0.022} & \pos{+0.215} & \pos{$-$0.153}\\[3.5pt]
Wan2.2\rule{0pt}{7pt} & \vsd{\textcolor{floorgray}{21.10}}{3.91} & \spd{1.00} & \vsd{11.73}{2.20} & \spd{1.39} & \vsd{16.53}{2.29} & \spd{0.81} & \vsd{16.09}{2.12} & \spd{0.71} & \vsd{18.12}{2.64} & \spd{0.66} & \vsd{15.03}{1.86} & \spd{1.31} & \vsd{15.38}{2.03} & \spd{1.26} & \vsd{17.09}{2.35} & \spd{0.40} & \vsd{\textbf{18.43}}{2.66} & \spd{1.07} & \pos{+2.72} & \dspd{+0.14}\\[4pt]
\kk{(4)} & \vsd{\textcolor{floorgray}{0.749}}{0.135} & \vsd{\textcolor{floorgray}{0.132}}{0.073} & \vsd{0.304}{0.118} & \vsd{0.576}{0.090} & \vsd{0.521}{0.139} & \vsd{0.296}{0.086} & \vsd{0.496}{0.136} & \vsd{0.314}{0.095} & \vsd{0.614}{0.142} & \vsd{0.218}{0.085} & \vsd{0.452}{0.127} & \vsd{0.361}{0.073} & \vsd{0.468}{0.131} & \vsd{0.340}{0.080} & \vsd{0.559}{0.134} & \vsd{0.260}{0.079} & \vsd{\textbf{0.634}}{0.130} & \vsd{\textbf{0.202}}{0.071} & \pos{+0.146} & \pos{$-$0.136}\\[3.5pt]
HY\rule{0pt}{7pt} & \vsd{\textcolor{floorgray}{37.13}}{11.63} & \spd{1.00} & \vsd{17.26}{1.82} & \spd{1.90} & \vsd{20.14}{3.28} & \spd{1.66} & \vsd{19.98}{3.54} & \spd{1.60} & \vsd{22.06}{5.05} & \spd{1.37} & \vsd{17.44}{2.66} & \spd{1.71} & \vsd{17.24}{2.37} & \spd{1.54} & \vsd{23.68}{5.11} & \spd{1.36} & \vsd{\textbf{27.47}}{5.54} & \spd{1.65} & \pos{+7.78} & \dspd{+0.06}\\[4pt]
\kk{(50)} & \vsd{\textcolor{floorgray}{0.929}}{0.142} & \vsd{\textcolor{floorgray}{0.073}}{0.168} & \vsd{0.664}{0.109} & \vsd{0.428}{0.111} & \vsd{0.755}{0.123} & \vsd{0.287}{0.141} & \vsd{0.742}{0.118} & \vsd{0.279}{0.158} & \vsd{0.820}{0.110} & \vsd{0.222}{0.137} & \vsd{0.698}{0.129} & \vsd{0.356}{0.147} & \vsd{0.685}{0.129} & \vsd{0.375}{0.133} & \vsd{0.820}{0.124} & \vsd{0.183}{0.122} & \vsd{\textbf{0.881}}{0.102} & \vsd{\textbf{0.130}}{0.101} & \pos{+0.140} & \pos{$-$0.174}\\[3.5pt]
Nano\rule{0pt}{7pt} & \vsd{\textcolor{floorgray}{25.34}}{6.20} & \spd{1.00} & \vsd{15.46}{2.53} & \spd{1.54} & \vsd{14.58}{1.68} & \spd{1.16} & \vsd{15.51}{2.28} & \spd{1.12} & \vsd{17.20}{3.19} & \spd{0.97} & \vsd{16.61}{2.56} & \spd{1.28} & \vsd{17.42}{4.11} & \spd{1.28} & \vsd{18.21}{3.45} & \spd{0.94} & \vsd{\textbf{21.56}}{5.04} & \spd{1.21} & \pos{+5.13} & \dspd{+0.03}\\[4pt]
\kk{(35)} & \vsd{\textcolor{floorgray}{0.832}}{0.117} & \vsd{\textcolor{floorgray}{0.091}}{0.075} & \vsd{0.494}{0.128} & \vsd{0.398}{0.115} & \vsd{0.423}{0.110} & \vsd{0.498}{0.078} & \vsd{0.473}{0.121} & \vsd{0.427}{0.097} & \vsd{0.535}{0.144} & \vsd{0.356}{0.087} & \vsd{0.534}{0.122} & \vsd{0.334}{0.099} & \vsd{0.590}{0.172} & \vsd{0.291}{0.146} & \vsd{0.609}{0.166} & \vsd{0.271}{0.123} & \vsd{\textbf{0.731}}{0.146} & \vsd{\textbf{0.154}}{0.093} & \pos{+0.208} & \pos{$-$0.214}\\[3.5pt]
\midrule
T2.5\rule{0pt}{7pt} & \vsd{\textcolor{floorgray}{41.93}}{4.94} & \spd{1.00} & \vsd{23.27}{1.29} & \spd{1.72} & \vsd{25.20}{1.83} & \spd{1.08} & \vsd{26.85}{2.05} & \spd{0.99} & \vsd{32.34}{2.46} & \spd{0.92} & \vsd{26.99}{1.93} & \spd{1.49} & \vsd{25.63}{1.58} & \spd{1.40} & \vsd{30.94}{2.37} & \spd{0.60} & \vsd{\textbf{33.81}}{2.87} & \spd{1.46} & \pos{+6.49} & \dspd{+0.29}\\[4pt]
\kk{(4)} & \vsd{\textcolor{floorgray}{0.988}}{0.010} & \vsd{\textcolor{floorgray}{0.014}}{0.013} & \vsd{0.808}{0.048} & \vsd{0.281}{0.057} & \vsd{0.823}{0.049} & \vsd{0.261}{0.077} & \vsd{0.860}{0.042} & \vsd{0.174}{0.058} & \vsd{0.942}{0.025} & \vsd{0.067}{0.034} & \vsd{0.874}{0.037} & \vsd{0.161}{0.052} & \vsd{0.858}{0.042} & \vsd{0.208}{0.056} & \vsd{0.929}{0.028} & \vsd{0.082}{0.039} & \vsd{\textbf{0.958}}{0.020} & \vsd{\textbf{0.050}}{0.025} & \pos{+0.087} & \pos{$-$0.126}\\[3.5pt]
\midrule
DR\rule{0pt}{7pt} & \vsd{\textcolor{floorgray}{54.70}}{1.61} & \spd{1.00} & \vsd{31.95}{7.85} & \spd{1.68} & \vsd{32.91}{3.43} & \spd{0.73} & \vsd{36.23}{3.61} & \spd{0.72} & \vsd{48.79}{2.42} & \spd{0.71} & \vsd{38.58}{2.69} & \spd{1.21} & \vsd{43.11}{3.16} & \spd{1.23} & \vsd{45.40}{2.85} & \spd{0.43} & \vsd{\textbf{50.87}}{1.73} & \spd{1.18} & \pos{+11.30} & \dspd{+0.22}\\[4pt]
\kk{(4)} & \vsd{\textcolor{floorgray}{0.998}}{0.001} & \vsd{\textcolor{floorgray}{0.001}}{0.001} & \vsd{0.968}{0.055} & \vsd{0.033}{0.047} & \vsd{0.973}{0.019} & \vsd{0.038}{0.022} & \vsd{0.984}{0.011} & \vsd{0.029}{0.020} & \vsd{0.995}{0.001} & \vsd{0.003}{0.001} & \vsd{0.989}{0.008} & \vsd{0.011}{0.004} & \vsd{0.994}{0.002} & \vsd{0.005}{0.003} & \vsd{0.995}{0.002} & \vsd{0.004}{0.002} & \vsd{\textbf{0.997}}{0.001} & \vsd{\textbf{0.002}}{0.001} & \pos{+0.012} & \pos{$-$0.016}\\[3.5pt]
Alaya\rule{0pt}{7pt} & \vsd{\textcolor{floorgray}{39.13}}{3.56} & \spd{1.00} & \vsd{18.87}{2.89} & \spd{1.57} & \vsd{19.24}{2.69} & \spd{1.05} & \vsd{20.11}{2.99} & \spd{0.98} & \vsd{26.36}{4.61} & \spd{0.77} & \vsd{18.64}{2.97} & \spd{1.29} & \vsd{18.67}{2.87} & \spd{1.31} & \vsd{20.45}{4.08} & \spd{0.67} & \vsd{\textbf{28.15}}{3.68} & \spd{1.24} & \pos{+7.82} & \dspd{+0.15}\\[4pt]
\kk{(50)} & \vsd{\textcolor{floorgray}{0.969}}{0.014} & \vsd{\textcolor{floorgray}{0.012}}{0.007} & \vsd{0.636}{0.126} & \vsd{0.433}{0.120} & \vsd{0.673}{0.115} & \vsd{0.369}{0.117} & \vsd{0.733}{0.106} & \vsd{0.262}{0.115} & \vsd{\textbf{0.871}}{0.058} & \vsd{0.096}{0.052} & \vsd{0.705}{0.102} & \vsd{0.302}{0.093} & \vsd{0.688}{0.102} & \vsd{0.327}{0.091} & \vsd{0.728}{0.161} & \vsd{0.274}{0.173} & \vsd{0.868}{0.070} & \vsd{\textbf{0.091}}{0.055} & \pos{+0.149} & \pos{$-$0.204}\\[3.5pt]
UniRel.\rule{0pt}{7pt} & \vsd{\textcolor{floorgray}{55.49}}{1.38} & \spd{1.00} & \vsd{30.99}{4.89} & \spd{1.55} & \vsd{28.76}{3.52} & \spd{1.21} & \vsd{30.25}{3.89} & \spd{1.16} & \vsd{\textbf{44.45}}{4.46} & \spd{0.99} & \vsd{33.47}{5.44} & \spd{1.33} & \vsd{33.93}{5.31} & \spd{1.33} & \vsd{40.84}{4.39} & \spd{0.90} & \vsd{42.75}{4.21} & \spd{1.31} & \pos{+8.08} & \dspd{+0.10}\\[4pt]
\kk{(35)} & \vsd{\textcolor{floorgray}{0.998}}{0.001} & \vsd{\textcolor{floorgray}{0.001}}{0.000} & \vsd{0.956}{0.023} & \vsd{0.069}{0.036} & \vsd{0.954}{0.064} & \vsd{0.055}{0.023} & \vsd{0.961}{0.074} & \vsd{0.041}{0.040} & \vsd{0.993}{0.010} & \vsd{0.008}{0.024} & \vsd{0.967}{0.056} & \vsd{0.042}{0.071} & \vsd{0.973}{0.027} & \vsd{0.036}{0.037} & \vsd{0.992}{0.006} & \vsd{0.008}{0.008} & \vsd{\textbf{0.994}}{0.002} & \vsd{\textbf{0.006}}{0.004} & \pos{+0.023} & \pos{$-$0.031}\\[3.5pt]
\midrule
Avg. & \textcolor{floorgray}{36.02} & \spd{1.00} & 19.18 & \spd{1.71} & 21.57 & \spd{1.09} & 22.54 & \spd{1.03} & 27.84 & \spd{0.91} & 22.49 & \spd{1.39} & 22.92 & \spd{1.33} & 26.24 & \spd{0.74} & \textbf{30.02} & \spd{1.31} & \pos{+6.77} & \dspd{+0.14}\\
 & \textcolor{floorgray}{0.918} & \textcolor{floorgray}{0.050} & 0.621 & 0.379 & 0.716 & 0.263 & 0.737 & 0.219 & 0.817 & 0.138 & 0.724 & 0.231 & 0.726 & 0.235 & 0.792 & 0.157 & \textbf{0.864} & \textbf{0.092} & \pos{+0.131} & \pos{$-$0.140}\\
\bottomrule
\end{tabular}
\caption{\textbf{Quality and speed at density 0.20.}
Each model shows PSNR (dB) and speedup over the fastest dense attention backend (top) and SSIM and LPIPS (bottom) against same-seed dense outputs, with test-prompt standard deviations below.
Floor compares two dense attention backends, bold marks the best sparse method, and $\Delta$ is DecoupAttn minus the baseline mean (Section~\ref{sec:setup}).}
\label{tab:main}
\end{table*}

\noindent\textbf{Sensitivity.}
The distortion of each unit is further scaled by a sensitivity $J_u$, the squared gain from the head to the final latent through the remaining layers and sampler steps (Appendix~\ref{app:th-prop}), which plays the role of Hessian sensitivity in mixed-precision quantization~\cite{dong2019hawq}.
To reduce the number of measurements, units are measured in groups $\mathcal G$, each covering a range of steps and of layers (Appendix~\ref{app:config}): the units of $\mathcal G$ run sparse at a reference level $\theta_0$ with full compensation while the rest stays dense, so the injected errors are the real sparse errors, and the extra final-latent error is divided by the group's summed distortion:
\begin{equation}
J_u=\frac{\mathrm{MSE}_{\mathcal G}-\mathrm{MSE}_0}{\sum_{u'\in\mathcal G}w_{u'}\,\varepsilon_{u'}(\textsc{Static}_{\theta_0},\textsc{Full})},
\label{eq:sens}
\end{equation}
where $\mathrm{MSE}_{\mathcal G}$ is the final-latent error of this run against the dense output, and the subtracted $\mathrm{MSE}_0$ is the run-to-run error of the same dense attention backend.
This reference gain prices every option of the unit (\hyperref[res:A17]{Lemma~A.17}).

\noindent\textbf{Allocation.}
With a price $J_u\varepsilon_u$ for every unit and option, choosing the plan becomes a budgeting problem under the density target $\rho$:
\begin{equation}
\begin{gathered}
\min_{\pi}\sum_u w_u J_u\,\varepsilon_u(o_u,\eta_{tl})\quad\text{s.t.}\quad X(\pi)\le\rho,\\
\mathcal L_\mu(\pi)=\sum_u w_u\big(J_u\,\varepsilon_u(o_u,\eta_{tl})+\mu X_u(o_u)\big),
\end{gathered}
\label{eq:alloc}
\end{equation}
where the first line approximates Eq.~\eqref{eq:problem} (\hyperref[res:prop2]{Proposition~2}).
The second line is the Lagrangian relaxation of Everett~\cite{everett1963lagrange}, which splits the search over plans into one choice per call: under each estimator every head takes the option with the lowest $J_u\varepsilon_u+\mu X_u$, and the call keeps the lower-total estimator.
Bisection on $\mu$ finds the densest plan that still fits $\rho$, and a final greedy pass spends the density left over.

\noindent\phantomsection\label{res:prop2}\textbf{Proposition 2.}
\emph{The summed priced distortion approximates the final-latent error, and for every nonnegative multiplier, the plan that minimizes $\mathcal L_\mu$ has the lowest priced distortion among all plans that are no denser, so bisection returns the best plan at its own density.}
(Proof in Appendix~\ref{app:prop2}, evidence in Section~\ref{sec:exp-budget}.)

\subsection{Reassign Cache}
\label{sec:reassign}

\begin{table*}[t]
\centering
\footnotesize
\newcommand{\vt}[1]{: \textcolor{floorgray}{#1}}
\setlength{\tabcolsep}{1.7pt}
\renewcommand{\arraystretch}{0.84}
\begin{tabular}{@{}c l *{10}{C{0.9cm}} C{0.9cm} C{0.72cm}@{}}
\toprule
Component & Variant & CP1 & CP2.5 & Wan2.1 & Wan2.2 & HY & Nano & T2.5 & DR & Alaya & UniRel. & Avg. & Supp.\\
\midrule
 & Full method & 23.42 & 24.07 & 25.31 & 17.21 & 25.84 & 20.92 & 33.34 & 50.20 & 27.97 & 41.54 & 28.98 & \\[-0.9pt]
 & \quad \textit{\scriptsize ms} & \tms{171{,}006} & \tms{8{,}212} & \tms{12{,}940} & \tms{12{,}820} & \tms{654{,}610} & \tms{188{,}278} & \tms{9{,}620} & \tms{3{,}905} & \tms{46{,}057} & \tms{108{,}377} & \tms{121{,}583} & \\
\midrule
\multirow{6}{*}{\hd{Decoupling\\(\S\ref{sec:decoup})}} & Tied\vt{blocks as groups} & $-$1.95 & $-$1.63 & $-$3.72 & $-$1.48 & $-$1.61 & $-$1.71 & $-$2.21 & $-$2.75 & $-$4.07 & $-$7.39 & $-$2.85 & 10/10\\[-0.9pt]
 & \quad \textit{\scriptsize ms} & \tms{$-$10{,}326} & \tms{$-$725} & \tms{$-$1{,}181} & \tms{$-$1{,}147} & \tms{$-$27{,}949} & \tms{$-$10{,}911} & \tms{$-$804} & \tms{$-$396} & \tms{$-$3{,}673} & \tms{$-$6{,}113} & \tms{$-$6{,}323} & \\[1.2pt]
 & Tied$^\dagger$\vt{same, our plan} & $-$2.45 & $-$1.90 & $-$4.19 & $-$1.66 & $-$0.85 & $-$2.17 & $-$2.67 & $-$5.29 & $-$5.43 & $-$8.59 & $-$3.52 & 10/10\\[-0.9pt]
 & \quad \textit{\scriptsize ms} & \tms{$-$9{,}592} & \tms{$-$707} & \tms{$-$1{,}119} & \tms{$-$1{,}146} & \tms{$-$29{,}257} & \tms{$-$10{,}610} & \tms{$-$802} & \tms{$-$423} & \tms{$-$3{,}482} & \tms{$-$6{,}308} & \tms{$-$6{,}345} & \\[1.2pt]
 & Centroid\vt{no corrections} & $-$1.60 & $-$1.03 & $-$1.24 & $-$0.53 & $-$1.06 & $-$0.60 & $-$1.20 & $-$2.48 & $-$1.43 & $-$3.77 & $-$1.49 & 10/10\\[-0.9pt]
 & \quad \textit{\scriptsize ms} & \tms{$-$2{,}621} & \tms{$-$95} & \tms{$-$176} & \tms{$-$148} & \tms{$-$23{,}626} & \tms{$-$6{,}768} & \tms{$-$75} & \tms{+121} & \tms{$-$1{,}154} & \tms{$-$1{,}885} & \tms{$-$3{,}643} & \\
\midrule
\multirow{4}{*}{\hd{Distortion\\Budget\\(\S\ref{sec:budget})}} & U\vt{every $J_u{=}1$} & $-$1.93 & $-$2.30 & $-$2.14 & $-$1.06 & $-$4.10 & $-$2.05 & $-$0.55 & $-$0.77 & $-$1.06 & $-$0.85 & $-$1.68 & 9/10\\[-0.9pt]
 & \quad \textit{\scriptsize ms} & \tms{+1{,}814} & \tms{+71} & \tms{+38} & \tms{$-$10} & \tms{+13{,}275} & \tms{+1{,}473} & \tms{+51} & \tms{$-$15} & \tms{$-$129} & \tms{+1{,}367} & \tms{+1{,}794} & \\[1.2pt]
 & D\vt{uncompensated $\varepsilon_u$} & $-$0.34 & $-$0.96 & $-$1.65 & $-$0.65 & $-$0.87 & $-$1.06 & $-$1.31 & $-$1.71 & $-$0.86 & $-$2.41 & $-$1.18 & 9/10\\[-0.9pt]
 & \quad \textit{\scriptsize ms} & \tms{$-$1{,}738} & \tms{+21} & \tms{+19} & \tms{+24} & \tms{+2{,}311} & \tms{$-$171} & \tms{+3} & \tms{$-$6} & \tms{+37} & \tms{+284} & \tms{+78} & \\
\midrule
\multirow{2}{*}{\hd{Reassign\\Cache (\S\ref{sec:reassign})}} & Frozen\vt{no reassignment} & $-$0.95 & $-$0.38 & $-$1.14 & $-$0.36 & $-$0.38 & $-$0.12 & $-$0.69 & -- & $-$2.59 & $-$4.45 & $-$1.23 & 8/9\\[-0.9pt]
 & \quad \textit{\scriptsize ms} & \tms{$-$7{,}484} & \tms{$-$150} & \tms{$-$218} & \tms{$-$255} & \tms{$-$26{,}463} & \tms{$-$8{,}050} & \tms{$-$167} & -- & \tms{$-$2{,}850} & \tms{$-$4{,}780} & \tms{$-$5{,}602} & \\
\bottomrule
\end{tabular}
\caption{\textbf{Component ablations (RQ3).}
Variant minus the full method (first row) on eight validation clips per model, in PSNR (dB) and latency (ms rows).
Supp.\ counts the models where the variant loses at least 0.3\,dB on average and on at least 6 of 8 clips (Table~\ref{tab:ablfull}).}
\label{tab:abl}
\end{table*}

The plan prices the residuals of clusters built at the first sparse step, and a cluster's residual is of second order in the spread of its members around the centroid (\hyperref[res:A6]{Proposition~A.6}).
As keys drift between denoising steps, the spread grows and the residual outgrows what the plan priced.
Reassign Cache bounds this spread at every step at little cost (Figure~\ref{fig:method}a).

\noindent\textbf{Cached centroids.}
We treat the centroids of the first sparse step as a fixed codebook and, at every later step, run the assignment step of Lloyd's algorithm~\cite{lloyd1982least}: each key moves to the cluster of its nearest cached centroid, and the statistics of Eq.~\eqref{eq:comp} are recomputed from the current members:
\begin{equation}
c_t(j)=\arg\min_c\big\|k_j^{(t)}-m_c\big\|_2^2,\qquad m_c=\bar k_c^{(t_1)},
\label{eq:reassign}
\end{equation}
where $t_1$ is the first sparse step, $k_j^{(t)}$ is key $j$ at step $t$, and $m_c$ is the cached centroid of cluster $c$.
One call costs one key--centroid product and an argmin (Appendix~\ref{app:th-cost}).

\noindent\textbf{Assignment rule.}
One measurement on the calibration set decides whether reassignment pays:
\begin{equation}
\text{reassign}=\mathbf 1\big[\delta_{\text{stale}}>\tau_s\big],
\label{eq:rules}
\end{equation}
where $\mathbf 1[\cdot]$ is the indicator function, $\delta_{\text{stale}}$ is the PSNR lost when the cluster assignments of the first sparse step stay frozen, and $\tau_s$ is a threshold (Appendix~\ref{app:config}).

\noindent\phantomsection\label{res:prop3}\textbf{Proposition 3.}
\emph{Reassigning the keys at a step gives the lowest total squared distance to the cached centroids of any assignment, and it changes only the compensation: the exact blocks and every unit's budget stay as planned.}
(Proof in Appendix~\ref{app:prop3}, evidence in Section~\ref{sec:exp-labels}.)

\section{Experiments}
\label{sec:exp}

We ask whether DecoupAttn is more accurate than prior sparse attention at equal density yet faster than dense attention (RQ1), whether one procedure fits all families and samplers (RQ2), what each component adds (RQ3), the quality--speed trade-off (RQ4), and what the plan costs (RQ5).

\subsection{Experimental Setup}
\label{sec:setup}

\begin{figure*}[t]
\centering
\includegraphics[width=\linewidth]{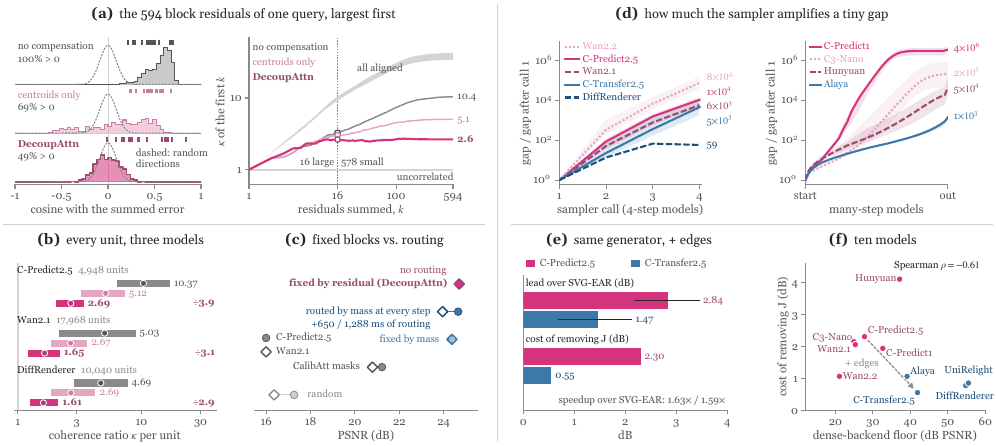}
\captionsetup{skip=2pt}
\caption{\textbf{Compensated block residuals cancel like dithered quantization errors, and conditioning damps error amplification.}
(a) For one Cosmos-Predict2.5 query, the small residuals align with the summed error without compensation but point randomly with DecoupAttn, so the coherence ratio \(\kappa\) of Eq.~\eqref{eq:coherence} stops growing after the 16 largest (ticks: these 16, 90\% of the energy).
(b) Per-unit \(\kappa\) drops 2.9--3.9-fold on three models (bars: quartiles, dots: medians, \(\div\): drop in median), and (c) under one compensator, blocks fixed by their residual match per-step routing by attention mass at no routing cost during inference.
(d) Two dense attention backends diverge along the sampler, less under per-pixel conditions (blue) than under prompts (pink).
(e) Adding edges to Cosmos-Predict2.5 halves our lead over SVG-EAR and cuts the value of the sensitivity fourfold (lines: 95\% bootstrap CI), and (f) on ten models, it falls as the dense-backend floor rises.}
\label{fig:obs}
\end{figure*}

\noindent\textbf{Models.}
We test ten models in three families (Table~\ref{tab:models}).
Video generation covers Cosmos-Predict1 (CP1)~\cite{agarwal2025cosmos}, Cosmos-Predict2.5 (CP2.5)~\cite{nvidia2025cosmos}, Wan2.1-14B at 480p (Wan2.1) and Wan2.2-A14B (Wan2.2)~\cite{wan2025wan}, HunyuanVideo (HY)~\cite{kong2024hunyuanvideo}, and image-to-video Cosmos3-Nano (Nano)~\cite{nvidia2026cosmos3}.
Controlled generation uses edge-conditioned Cosmos-Transfer2.5 (T2.5)~\cite{nvidia2025cosmos}, and neural rendering uses DiffusionRenderer (DR)~\cite{liang2025diffusionrenderer}, AlayaRenderer (Alaya)~\cite{huang2026alaya}, and UniRelight (UniRel.)~\cite{he2025unirelight}.
Five run four-step distilled samplers and five their original many-step samplers.
Each model has disjoint calibration (at most 16 clips), validation (8), and test (8--39) splits, all with seed 1000.

\noindent\textbf{Baselines.}
We compare with seven methods from three lines of work, all run from their official code: offline calibration (CalibAtt~\cite{yehezkel2026calibatt}), online routing by clustering (SVOO~\cite{luo2026svoo}, SVG2~\cite{yang2025svg2}) or by block scoring (Sol-Attn~\cite{li2026solattn}), and compensation-based sparse attention (SVG-EAR~\cite{zhou2026svgear}, PISA~\cite{li2026pisa}, SparsePR~\cite{taghavi2026sparsepr}).
For a fair comparison, every method keeps the first step and layer dense, runs the other calls at density 0.20 or its official minimum (Appendix~\ref{app:exp}, 0.23--0.42 with padding in Table~\ref{tab:density}), and is scored and timed against the same dense outputs in one process.

\noindent\textbf{Metrics and implementation.}
Quality is the PSNR of each video against the dense output with the same seed, together with SSIM~\cite{wang2004ssim} and LPIPS~\cite{zhang2018lpips} and six VBench dimensions~\cite{huang2024vbench} (Appendix~\ref{app:exp}), and speed is the latency of the denoising loop (DiT) and its speedup over dense attention.
DecoupAttn sparsifies the spatiotemporal self-attention of every model and keeps cross-attention dense.
Exact blocks run on FlashInfer's variable-block FlashAttention-3 kernel~\cite{ye2025flashinfer,shah2024flashattention3}, and one kernel merges the cluster terms (Appendix~\ref{app:config}).

\subsection{Main Results}
\label{sec:main}

\begin{figure*}[t]
\centering
\newcommand{\ptitle}[2]{{\centering\fontsize{6.3}{7.5}\selectfont\textbf{#1}\hspace{2.6pt}#2\par}\vspace{1.5pt}}
\begin{minipage}[t]{0.372\linewidth}
\vspace{0pt}
\ptitle{(a)}{Time of one video (s) and the plan's cost}
\scriptsize
\setlength{\tabcolsep}{1.9pt}
\renewcommand{\arraystretch}{0.98}
\begin{tabular*}{\linewidth}{@{\extracolsep{\fill}}l *{5}{r}@{}}
\toprule
Method & k-means & per step & Attn. & DiT & E2E\\
\midrule
\multicolumn{6}{@{}l}{\textit{CP2.5}\hfill\textcolor{floorgray}{plan: 5.9 GPU-h, repaid after 5.3k videos}}\\
Dense & -- & -- & 9.35 & 12.15 & 19.68\\
SVG-EAR & 4.73 & 0.92 & 10.69 & 13.35 & 20.96\\
Ours & \textbf{0.62} & \textbf{0.57} & \textbf{5.48} & \textbf{8.20} & \textbf{15.71}\\
\multicolumn{6}{@{}l}{\textit{Wan2.1}\hfill\textcolor{floorgray}{plan: 5.3 GPU-h, repaid after 23.7k videos}}\\
Dense & -- & -- & 4.84 & 13.65 & 16.40\\
SVG-EAR & 7.14 & 1.33 & 11.91 & 20.52 & 23.27\\
Ours & \textbf{0.92} & \textbf{0.91} & \textbf{4.27} & \textbf{12.91} & \textbf{15.66}\\
\multicolumn{6}{@{}l}{\textit{UniRel.}\hfill\textcolor{floorgray}{plan: 19.6 GPU-h, repaid after 2.1k videos}}\\
Dense & -- & -- & 68.42 & 141.10 & 144.40\\
SVG-EAR & 18.44 & 20.88 & 77.65 & 144.76 & 147.92\\
Ours & \textbf{1.22} & \textbf{15.27} & \textbf{39.69} & \textbf{108.43} & \textbf{111.72}\\
\bottomrule
\end{tabular*}

\end{minipage}\hfill
\begin{minipage}[t]{0.615\linewidth}
\vspace{0pt}
\includegraphics[width=\linewidth]{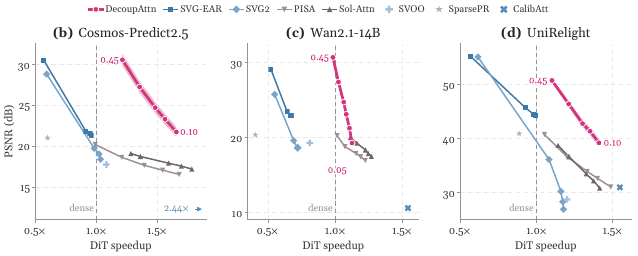}
\end{minipage}

\vspace{3.6pt}
\begin{minipage}[t]{0.745\linewidth}
\vspace{0pt}
\ptitle{(e)}{Skipped keys and exact blocks, changed one at a time}
\scriptsize
\setlength{\tabcolsep}{1.5pt}
\renewcommand{\arraystretch}{0.9}
\begin{tabular*}{\linewidth}{@{\extracolsep{\fill}}l l *{10}{c}@{}}
\toprule
 & & \multicolumn{5}{c}{Cosmos-Predict2.5} & \multicolumn{5}{c}{Wan2.1-14B}\\
\cmidrule(lr){3-7}\cmidrule(l){8-12}
Grouping & Corrections & Attn. & Latent & PSNR & LPIPS & DiT (s) & Attn. & Latent & PSNR & LPIPS & DiT (s)\\
\midrule
-- & none & 415 & 214 & 17.57\sd{2.33} & 0.385 & 7.18 & 577 & 186 & 17.08\sd{1.10} & 0.300 & 11.29\\
position & centroid & 38.6 & 174 & 18.36\sd{2.72} & 0.297 & 7.39 & 109 & 88.6 & 19.69\sd{1.46} & 0.155 & 11.44\\
position & both & 23.1 & 69.0 & 22.33\sd{3.26} & 0.152 & 7.47 & 73.2 & 68.5 & 20.66\sd{1.62} & 0.123 & 11.55\\
semantic & centroid & 10.8 & 54.1 & 23.47\sd{3.68} & 0.124 & 8.13 & 14.7 & 36.4 & 23.41\sd{2.05} & 0.078 & 12.76\\
semantic & both (ours) & 6.23 & 41.2 & 24.70\sd{3.78} & 0.102 & 8.21 & 8.80 & 26.0 & 24.70\sd{2.29} & 0.057 & 12.90\\
\midrule
Selection & Ranking & \multicolumn{2}{c}{Routing (ms)} & PSNR & LPIPS & DiT (s) & \multicolumn{2}{c}{Routing (ms)} & PSNR & LPIPS & DiT (s)\\
\midrule
offline & attention mass & \multicolumn{2}{c}{0} & 24.38\sd{3.66} & 0.107 & 8.22 & \multicolumn{2}{c}{0} & 24.37\sd{2.24} & 0.061 & 12.91\\
offline & residual (ours) & \multicolumn{2}{c}{0} & 24.70\sd{3.78} & 0.102 & 8.21 & \multicolumn{2}{c}{0} & 24.70\sd{2.29} & 0.057 & 12.90\\
online & attention mass & \multicolumn{2}{c}{650} & 24.65\sd{3.46} & 0.098 & 8.86 & \multicolumn{2}{c}{1{,}288} & 23.95\sd{2.08} & 0.066 & 14.19\\
offline & random & \multicolumn{2}{c}{0} & 17.26\sd{2.61} & 0.415 & 8.22 & \multicolumn{2}{c}{0} & 16.36\sd{1.42} & 0.289 & 12.87\\
\multicolumn{2}{@{}l}{CalibAtt masks} & \multicolumn{2}{c}{0} & 21.22\sd{3.39} & 0.183 & 8.21 & \multicolumn{2}{c}{0} & 20.79\sd{1.73} & 0.116 & 12.88\\
\bottomrule
\end{tabular*}
\end{minipage}\hfill
\begin{minipage}[t]{0.243\linewidth}
\vspace{0pt}
\includegraphics[width=\linewidth]{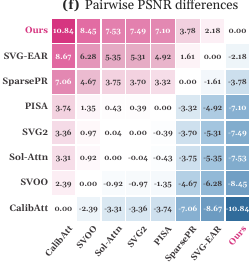}
\end{minipage}
\captionsetup{skip=3pt}
\caption{\textbf{Where the speed and accuracy come from, and how they hold.}
(a) Medians of interleaved runs; per step covers assignment, statistics, counts, and merging.
(b--d) DecoupAttn sweeps its density (labels; shaded: $\pm$1 standard error).
(e) Top: our exact mask, bottom: our compensator (test prompts; MSE $\times10^{3}$; small $\pm$: standard deviation).
(f) Row minus column PSNR (dB), mean over ten models.}
\label{fig:tradeoff}
\end{figure*}

\noindent\textbf{DecoupAttn is accurate and fast at once (RQ1).}
As shown in Table~\ref{tab:main}, DecoupAttn has the highest PSNR on nine of the ten models, 2.18\,dB above the strongest baseline on average, while running 1.31$\times$ faster than dense attention and up to 1.65$\times$.
Among the methods faster than dense attention it is the most accurate on all ten and leads every baseline in pairwise PSNR (Figure~\ref{fig:tradeoff}f), with the lowest LPIPS on all ten, VBench within 0.007 of dense attention (Tables~\ref{tab:full} and~\ref{tab:vbench}), and dense-level error to ground truth (Table~\ref{tab:cond}, cases in Appendix~\ref{app:cases}).
The three lines of prior work show where this comes from: dropping the skipped keys is fastest but least accurate, compensating with cheap position blocks lands in between, and compensation that re-clusters at every step falls below dense speed on all four-step models (Figure~\ref{fig:tradeoff}a, Table~\ref{tab:timing}).
Decoupling removes this trade-off: blocks and clusters meet only through the counts $n_{gc}$, so it fixes the blocks offline, clusters once, and compensates accurately at fixed-mask speed.

\noindent\textbf{One procedure serves every family and sampler (RQ2).}
With the same procedure and constants, DecoupAttn leads the best baseline in all three families and both sampler types, by up to 3.79\,dB, with each model's sampler setting its own $J_u$ (Eq.~\eqref{eq:sens}).

\subsection{Component Analysis (RQ3)}
\label{sec:exp-components}

We validate each component by removing it from the full method and re-solving the plan at the same density (Table~\ref{tab:abl}).
Every removal lowers PSNR in all 59 cells.

\noindent\textbf{Decoupling matters most.}
\label{sec:exp-decoup}
Grouping compensation by the selection blocks (Tied), as in \hyperref[res:def1]{Definition~1}, costs 2.85\,dB on average at almost unchanged speed, and 3.52\,dB under our plan (Tied$^\dagger$).
Position blocks group keys by place, not content, so their members spread far from the centroid and leave large residuals: under one exact mask, semantic clusters raise PSNR from 18.36 to 23.47\,dB (Figure~\ref{fig:tradeoff}e, Figure~\ref{fig:groups}).
A centroid underestimates the mass of a spread-out cluster, so dropping both corrections (Centroid) costs 1.49\,dB (Appendix~\ref{app:supp-corr}).

\noindent\textbf{With compensation on, fixed blocks are as good as routing.}
\label{sec:exp-dither}
Under the same compensator, routing by attention mass at every call matches our fixed blocks on Cosmos-Predict2.5 at 650\,ms of routing and falls behind on Wan2.1-14B (Figure~\ref{fig:tradeoff}e, Table~\ref{tab:selection}).
The cancellation of Section~\ref{sec:decoup} explains this: compensated residuals cancel each other, cutting the median coherence ratio on Cosmos-Predict2.5 from 10.37 to 2.69 (Figure~\ref{fig:obs}a--c, Table~\ref{tab:kappa}).
Ranking blocks offline by this residual also beats ranking by mass or CalibAtt's masks (Figure~\ref{fig:tradeoff}e).

\noindent\textbf{Pricing the residual by its effect on the output matters next.}
\label{sec:exp-budget}
Setting every sensitivity to one (U) costs 1.68\,dB on average, and pricing options by their error before compensation (D), which buys exact blocks where compensation already works, costs 1.18\,dB, both at unchanged speed (Table~\ref{tab:ablspeed}).
The price orders 21 candidate plans by their held-out error as \hyperref[res:prop2]{Proposition~2} expects, and the deployed plan is the second best (Figure~\ref{fig:proxy}, Table~\ref{tab:proxy}).
The sensitivity pays most where the sampler amplifies errors: removing it costs 2.30\,dB on Cosmos-Predict2.5 but 0.55\,dB with edges (Figure~\ref{fig:obs}e,f, Appendix~\ref{app:supp-budget}).

\noindent\textbf{Reassignment keeps the clusters current at little cost.}
\label{sec:exp-labels}
Drifting keys spread frozen clusters beyond what the plan priced, so freezing them (Frozen) costs 1.23\,dB on average over the nine models that reassign (Table~\ref{tab:rule}).
Reassignment never increases this spread (\hyperref[res:prop3]{Proposition~3}, Figure~\ref{fig:drift}), so per-step k-means (Rebuild) changes PSNR by at most 0.95\,dB for 11--43\% more latency (Tables~\ref{tab:rebuild},~\ref{tab:ablspeed}, and~\ref{tab:cache}).

\subsection{Trade-off and Offline Plan (RQ4, RQ5)}
\label{sec:exp-tradeoff}

\noindent\textbf{The lead holds across densities (RQ4).}
Sweeping its density from 0.10 (0.05 on Wan2.1-14B) to 0.45, every baseline lies below its curve within its speed range (Figure~\ref{fig:tradeoff}b--d), since each point is its density's lowest-priced plan (\hyperref[res:prop2]{Proposition~2}), and at the latency of the strongest baseline it is 5.74--8.40\,dB more accurate (Figure~\ref{fig:pareto}, Table~\ref{tab:iso}).

\noindent\textbf{The offline plan is cheap and holds on new content (RQ5).}
\label{sec:exp-practice}
Calibration takes 5.3--19.6 GPU-hours per model, which the faster inference repays after 2.1k--23.7k videos (Figure~\ref{fig:tradeoff}a, Table~\ref{tab:offline}).
As the frozen plan fixes only where exact attention runs and each video clusters its own keys, it beats SVG-EAR on all held-out content (Table~\ref{tab:shift}), and Appendix~\ref{app:supp-gen} varies calibration, resolution, and cluster count.

\section{Conclusion}
\label{sec:conclusion}

We introduced DecoupAttn, a decoupled sparse attention that keeps compensation when the selection is fixed offline: semantic clusters compensate every key, exact attention corrects their residual on a few fixed blocks, Distortion Budget prices that residual, and Reassign Cache keeps the clusters current.
On ten video generation, controlled generation, and neural rendering models, DecoupAttn is the most accurate method faster than dense attention, at up to 1.65$\times$ its speed.
Our results argue for organizing exact computation and approximation separately: position blocks for the kernel, semantic clusters for the approximation error.

\newpage
\section*{LLM Usage Statement}
We used LLMs (Claude) to polish the grammar and word choice of the paper and to refine the plotting scripts of the figures.
The authors designed the method and the experiments, ran all experiments, and checked every statement, number, and reference in the paper.

{
    \small
    \bibliographystyle{ieeenat_fullname}
    \bibliography{main}
}

\clearpage
\setcounter{page}{1}
\maketitlesupplementary
\appendix
\setcounter{topnumber}{5}\setcounter{bottomnumber}{3}\setcounter{totalnumber}{8}\setcounter{dbltopnumber}{3}
\renewcommand{\topfraction}{0.9}\renewcommand{\bottomfraction}{0.8}\renewcommand{\textfraction}{0.07}
\renewcommand{\floatpagefraction}{0.75}\renewcommand{\dbltopfraction}{0.9}\renewcommand{\dblfloatpagefraction}{0.75}

\section{Theory of Decoupled Sparse Attention}
\label{app:proofs}

This appendix develops the theory behind the three components of Section~\ref{sec:method} and proves its three propositions.
Section~\ref{app:th-setting} fixes the notation and the assumptions.
Section~\ref{app:prop1} shows that the decoupled estimator is an exact rearrangement of attention (\hyperref[res:prop1]{Proposition~1}) whose error is additive over blocks.
Section~\ref{app:th-comp} expands the centroid compensation to second order and identifies what each correction removes.
Section~\ref{app:th-coh} explains why fixed blocks suffice: compensation subtracts exactly the part of the block residuals that aligns, and for blocks fixed independently of the cluster membership the remaining deviation part is no more coherent than random errors.
Section~\ref{app:th-sel} turns this into a rule for choosing the exact blocks, and Section~\ref{app:th-diff} relates the estimator to difference estimators.
Section~\ref{app:prop2} analyzes the plan objective, the sensitivity estimator, and the Lagrangian relaxation (\hyperref[res:prop2]{Proposition~2}).
Section~\ref{app:prop3} proves \hyperref[res:prop3]{Proposition~3} and the monotone chain behind Reassign Cache, Section~\ref{app:th-cost} counts the cost of one call, and Section~\ref{app:th-prop} traces how errors travel through the sampler.
Algebraic identities hold on every attention call, the expansions hold under the bounded-deviation assumption (A3), their orders in the deviations under its joint form (A3$'$), and the statements about expectations under the exchangeability model (A4).

\subsection{Setting and Notation}
\label{app:th-setting}

Fix one self-attention call and one query $i$, and drop the indices of step, layer, and head.
The logits are $s_{ij}=q_i^\top k_j/\sqrt d$, and the dense quantities $N_{ib}$, $Z_{ib}$, $N_i$, $Z_i$, and $y_i=N_i/Z_i$ are those of Eq.~\eqref{eq:dense}.
The key blocks $b$ and the clusters $c$ are two partitions of the same $L$ keys, and $n_{bc}=|b\cap c|$ counts the keys they share, so that
\begin{equation*}
\textstyle\sum_b n_{bc}=|c|,\qquad \sum_c n_{bc}=|b|.
\end{equation*}
The query tile $g$ of query $i$ has exact blocks $\mathcal S_g$ and skipped blocks $\mathcal K_g$, the complement of $\mathcal S_g$.
Every member $j$ of cluster $c$ is written as a deviation from the cluster means,
\begin{equation*}
\begin{gathered}
k_j=\bar k_c+\delta_j,\qquad v_j=\bar v_c+\nu_j,\\
\textstyle\sum_{j\in c}\delta_j=0,\qquad \sum_{j\in c}\nu_j=0,
\end{gathered}
\end{equation*}
and $x_{ij}=q_i^\top\delta_j/\sqrt d$ is the logit deviation, so that $e^{s_{ij}}=\phi_{ic}\,e^{x_{ij}}$ with $\phi_{ic}=e^{q_i^\top\bar k_c/\sqrt d}$.
The key covariance and the value--key cross-covariance of a cluster are
\begin{equation*}
C_c=\frac{1}{|c|}\sum_{j\in c}\delta_j\delta_j^\top,\qquad B_c=\frac{1}{|c|}\sum_{j\in c}\nu_j\delta_j^\top,
\end{equation*}
so the per-dimension variance $\sigma_c^2$ of Eq.~\eqref{eq:comp} is the diagonal of $C_c$ and the variance correction reads
\begin{equation*}
\gamma_{ic}=\frac{1}{2d}\,q_i^\top\operatorname{diag}(C_c)\,q_i .
\end{equation*}
The per-member predictions of Eq.~\eqref{eq:comp} are $\omega_{ic}=\phi_{ic}(1+\gamma_{ic})$ and $a_{ic}=\omega_{ic}\bar v_c+\phi_{ic}Hq_i/\sqrt d$.
We use four assumptions, the third in a weak and a joint form.
Table~\ref{tab:notation} collects the symbols used in the proofs.

\begin{table}[!htbp]
\centering
\footnotesize
\setlength{\tabcolsep}{3pt}
\renewcommand{\arraystretch}{1.08}
\begin{tabular}{@{}l p{6.1cm}@{}}
\toprule
Symbol & Meaning\\
\midrule
$s_{ij}$, $N_{ib}$, $Z_{ib}$, $y_i$ & logit, block numerator and mass, dense output (Eq.~\eqref{eq:dense})\\
$b$, $c$, $n_{bc}$ & key block, cluster, keys they share\\
$\mathcal S_g$, $\mathcal K_g$, $n_{gc}$ & exact and skipped blocks of query tile $g$, members of $c$ outside $\mathcal S_g$\\
$\bar k_c$, $\bar v_c$, $\delta_j$, $\nu_j$ & cluster means and member deviations\\
$x_{ij}$, $\phi_{ic}$ & logit deviation $q_i^\top\delta_j/\sqrt d$, centroid weight\\
$C_c$, $B_c$, $H$ & key covariance, value--key cross-covariance, pooled $B_c$\\
$\gamma_{ic}$, $\omega_{ic}$, $a_{ic}$ & variance correction, per-member mass and numerator prediction\\
$u_{ij}$, $r_{ib}$, $U_{ic}$ & error of one compensated key, of a block, of a whole cluster\\
$\bar z_{ic}$, $w_{ij}$ & cluster-mean contribution, member deviation\\
$D_b$, $\beta_{ib}$ & deviation term and cluster bias of a block\\
$\eta$, $V$ & deviation bound, largest value norm $\max_j\|v_j\|_2$\\
$\kappa_i$, $R_i$, $\zeta_i$ & coherence ratio, residual energy, mass ratio error\\
$\varepsilon_u$, $J_u$, $w_u$, $X_u$ & distortion, sensitivity, step weight, density of a unit\\
$\mu$, $\rho$, $F$, $\mathcal L_\mu$ & multiplier, density target, priced distortion, Lagrangian\\
\bottomrule
\end{tabular}
\caption{Notation of Appendix~\ref{app:proofs}.}
\label{tab:notation}
\end{table}

\noindent\textbf{(A1) Count-based compensation.}
Every member of cluster $c$ that lies outside the exact blocks contributes $a_{ic}$ to the numerator and $\omega_{ic}$ to the denominator, and these terms depend only on the cluster and the query.

\noindent\textbf{(A2) Current statistics.}
$\bar k_c$, $\bar v_c$, $\sigma_c^2$, and $H$ are computed from the current members of each cluster, at every call.

\noindent\textbf{(A3) Bounded deviations.}
$|x_{ij}|\le\eta$ for every query and every member of every cluster.

\noindent\textbf{(A3$'$) Small joint deviations.}
$\|q_i\|_2\|\delta_j\|_2/\sqrt d\le\eta$ and $\|\nu_j\|_2\le\eta V$, with $V=\max_j\|v_j\|_2$, for every query and every member of every cluster.

\noindent\textbf{(A4) Exchangeable membership.}
Within each cluster, the members are distributed over the blocks by a uniformly random permutation that keeps the counts $n_{bc}$, independently across clusters and of the members' keys and values.

(A1) and (A2) describe the implementation and hold exactly.
(A3) is the regime in which the expansions of Section~\ref{app:th-comp} are accurate, and clustering makes $\eta$ small because k-means groups nearby keys.
(A3$'$) implies (A3) and adds that the values of a cluster vary on the same small scale, as when values move with their keys.
Under it every correction term is quadratic and every remainder cubic in $\eta$.
(A4) is a statistical model of the membership under which Section~\ref{app:th-coh} computes expectations: it states that which block a member falls into carries no information about the direction of its deviation once its cluster is known.
Blocks fixed offline never read the membership of a new input, so the skipped set is independent of the membership, and Figure~\ref{fig:cosdist} tests the exchangeability itself on saved attention inputs.

\subsection{The Decoupled Estimator (Proposition 1)}
\label{app:prop1}

\noindent\textbf{Proof of Proposition 1.}
The key blocks partition the keys, so $|c|=\sum_b n_{bc}$ and $n_{gc}=\sum_{b\notin\mathcal S_g}n_{bc}$.
Substituting into Eq.~\eqref{eq:decoup} gives
\begin{equation*}
\begin{aligned}
\hat N_i&=\sum_{b\in\mathcal S_g}N_{ib}+\sum_{b\notin\mathcal S_g}\sum_c n_{bc}\,a_{ic}\\
&=\sum_c|c|\,a_{ic}+\sum_{b\in\mathcal S_g}\Big(N_{ib}-\sum_c n_{bc}\,a_{ic}\Big),
\end{aligned}
\end{equation*}
which is the first identity of Eq.~\eqref{eq:residual}, and the denominator follows in the same way with $Z_{ib}$ and $\omega_{ic}$.
Since $N_i=\sum_b N_{ib}$ and $Z_i=\sum_b Z_{ib}$ (Eq.~\eqref{eq:dense}), the same rearrangement gives $N_i-\hat N_i=\sum_{b\notin\mathcal S_g}\Delta N_{ib}$ and $Z_i-\hat Z_i=\sum_{b\notin\mathcal S_g}\Delta Z_{ib}$.
Using $N_i=y_iZ_i$,
\begin{equation*}
\hat y_i-y_i=\frac{\hat N_i-y_i\hat Z_i}{\hat Z_i}=-\frac{1}{\hat Z_i}\sum_{b\notin\mathcal S_g}\big(\Delta N_{ib}-y_i\Delta Z_{ib}\big).
\end{equation*}
The identity holds for any fixed blocks and any clusters. $\square$

The residual has a per-member form that the rest of this appendix uses.
For a member $j$ of cluster $c$, let
\begin{equation*}
u_{ij}=e^{s_{ij}}\big(v_j-y_i\big)-\big(a_{ic}-y_i\,\omega_{ic}\big)
\end{equation*}
be the error that compensating this one key leaves.
Expanding $\Delta N_{ib}$ and $\Delta Z_{ib}$ over the members of block $b$ shows that the residual of a block is the sum of the errors of its keys:
\begin{equation*}
r_{ib}=\sum_{j\in b}u_{ij}.
\end{equation*}

\noindent\phantomsection\label{res:A1}\textbf{Lemma A.1 (Additive block removal).}
\emph{Let $E_i(\mathcal S)=\hat N_i-y_i\hat Z_i$ be the unnormalized error of the estimator with exact blocks $\mathcal S$.
Then $E_i(\mathcal S)=-\sum_{b\notin\mathcal S}r_{ib}$, and for every block $b\notin\mathcal S$,}
\begin{equation*}
E_i(\mathcal S\cup\{b\})=E_i(\mathcal S)+r_{ib},\qquad \hat Z_i(\mathcal S\cup\{b\})=\hat Z_i(\mathcal S)+\Delta Z_{ib}.
\end{equation*}
\emph{Proof.}
The first claim is the last display of the proof of \hyperref[res:prop1]{Proposition~1} multiplied by $\hat Z_i$.
Making $b$ exact removes its term from the sum over $b\notin\mathcal S$, and replaces its compensated mass $\sum_c n_{bc}\omega_{ic}$ by $Z_{ib}$ in the denominator. $\square$

\hyperref[res:A1]{Lemma~A.1} says that every block owns a fixed share $r_{ib}$ of the unnormalized error, whatever the other blocks do.
This is what makes a residual a property of a block that can be measured once and priced offline, while the coherence ratio and the mass $\hat Z_i$ decide how the shares combine into the output error (\hyperref[res:A2]{Corollary~A.2}).

\noindent\phantomsection\label{res:A2}\textbf{Corollary A.2 (Three factors of the error).}
\emph{With $R_i=\sum_{b\notin\mathcal S_g}\|r_{ib}\|_2^2$ and the coherence ratio $\kappa_i$ of Eq.~\eqref{eq:coherence}, set to one when $R_i=0$ and between $0$ and $|\mathcal K_g|$ by Cauchy--Schwarz,}
\begin{equation*}
\|\hat y_i-y_i\|_2^2=\frac{\kappa_i\,R_i}{\hat Z_i^{\,2}},\qquad \frac{\|\hat y_i-y_i\|_2}{\|y_i\|_2}\le\frac{1}{1-\zeta_i}\,\frac{\sqrt{\kappa_iR_i}}{Z_i\|y_i\|_2}
\end{equation*}
\emph{whenever $\zeta_i=|\hat Z_i/Z_i-1|<1$ and $y_i\ne0$.}
\emph{Proof.}
The equality combines \hyperref[res:prop1]{Proposition~1} with the definition of $\kappa_i$.
For the bound, $\hat Z_i\ge(1-\zeta_i)Z_i$. $\square$

The three factors are the residual energy $R_i$, its coherence $\kappa_i$, and the mass ratio $\hat Z_i/Z_i$, which are the columns of Table~\ref{tab:resid}.
Only the variance correction acts on $\zeta_i$, because the value correction enters $\hat N_i$ but not $\hat Z_i$: on DiffusionRenderer the median $|\hat Z_i/Z_i-1|$ is 0.27 without compensation, 0.100 with plain centroids or the value correction alone, and 0.035 with the variance correction (Table~\ref{tab:resid}).

\subsection{Second-Order Analysis of the Compensation}
\label{app:th-comp}

\noindent\phantomsection\label{res:A3}\textbf{Lemma A.3 (A centroid underestimates the mass).}
\emph{For every cluster and every query, $\sum_{j\in c}e^{s_{ij}}\ge|c|\,\phi_{ic}$, with equality only if all logit deviations $x_{ij}$ of the cluster are zero.}

\emph{Proof.}
The exponential is strictly convex and the members' keys average to $\bar k_c$, so Jensen's inequality gives
\begin{equation*}
\frac{1}{|c|}\sum_{j\in c}e^{s_{ij}}\ge\exp\Big(\frac{1}{|c|}\sum_{j\in c}s_{ij}\Big)=\exp\big(q_i^\top\bar k_c/\sqrt d\big)=\phi_{ic},
\end{equation*}
with equality exactly when the $s_{ij}$ are equal, that is, when the $x_{ij}$ vanish. $\square$

\hyperref[res:A3]{Lemma~A.3} needs no expansion.
It shows that a centroid underestimates the mass of a spread-out cluster and, under (A4), in expectation that of the members the exact blocks leave, since each member is skipped with equal probability.

\noindent\phantomsection\label{res:A4}\textbf{Lemma A.4 (The variance correction is the diagonal second-order term).}
\emph{Under (A3),}
\begin{equation*}
\sum_{j\in c}e^{s_{ij}}=|c|\,\phi_{ic}\Big(1+\frac{1}{2d}\,q_i^\top C_c\,q_i\Big)+\phi_{ic}\,\rho_{ic},\qquad |\rho_{ic}|\le\frac{|c|\,\eta^3e^{\eta}}{6},
\end{equation*}
\emph{and therefore the corrected prediction misses the mass of the cluster by}
\begin{equation*}
\sum_{j\in c}e^{s_{ij}}-|c|\,\omega_{ic}=\frac{|c|\,\phi_{ic}}{2d}\,q_i^\top\big(C_c-\operatorname{diag}(C_c)\big)q_i+\phi_{ic}\,\rho_{ic}.
\end{equation*}

\emph{Proof.}
Taylor's theorem gives $e^x=1+x+x^2/2+\varrho(x)$ with $|\varrho(x)|\le|x|^3e^{|x|}/6$.
Summing $e^{s_{ij}}=\phi_{ic}e^{x_{ij}}$ over the cluster, the linear term vanishes because $\sum_{j\in c}x_{ij}=q_i^\top\sum_{j\in c}\delta_j/\sqrt d=0$, and the quadratic term is
\begin{equation*}
\frac12\sum_{j\in c}x_{ij}^2=\frac{1}{2d}\,q_i^\top\Big(\sum_{j\in c}\delta_j\delta_j^\top\Big)q_i=\frac{|c|}{2d}\,q_i^\top C_c\,q_i .
\end{equation*}
The remainders add up to $\rho_{ic}=\sum_{j\in c}\varrho(x_{ij})$, which (A3) bounds term by term.
Subtracting $|c|\,\omega_{ic}=|c|\,\phi_{ic}(1+\gamma_{ic})$ with $\gamma_{ic}=q_i^\top\operatorname{diag}(C_c)\,q_i/(2d)$ gives the second display. $\square$

The \textsc{Centroid} estimator therefore leaves the whole second-order term $|c|\,\phi_{ic}\,q_i^\top C_cq_i/(2d)\ge0$, in agreement with \hyperref[res:A3]{Lemma~A.3}.
The variance correction removes its diagonal part and leaves only the covariance between different key dimensions within a cluster, of order $\eta^2$ under (A3$'$), plus a remainder of order $\eta^3$.

\noindent\phantomsection\label{res:A5}\textbf{Lemma A.5 (The value correction is the pooled cross-covariance).}
\emph{(i) The matrix of Eq.~\eqref{eq:comp} is the size-weighted average of the clusters' cross-covariances,}
\begin{equation*}
H=\sum_c\frac{|c|}{L}\,B_c .
\end{equation*}
\emph{(ii) Under (A3), with $V=\max_j\|v_j\|_2$ and $V_\nu=\max_j\|\nu_j\|_2$,}
\begin{equation*}
\sum_{j\in c}e^{s_{ij}}v_j=|c|\,\phi_{ic}\Big[\bar v_c\Big(1+\frac{q_i^\top C_cq_i}{2d}\Big)+\frac{B_c\,q_i}{\sqrt d}\Big]+\phi_{ic}\,\rho'_{ic},
\end{equation*}
\emph{with $\|\rho'_{ic}\|_2\le|c|\big(\eta^2V_\nu/2+\eta^3e^{\eta}V/6\big)$.}

\emph{Proof.}
(i) Within one cluster the cross terms vanish because the deviations have zero mean:
\begin{equation*}
\sum_{j\in c}v_jk_j^\top=|c|\,\bar v_c\bar k_c^\top+\sum_{j\in c}\nu_j\delta_j^\top=|c|\,\bar v_c\bar k_c^\top+|c|\,B_c .
\end{equation*}
Summing over clusters and dividing by $L$ gives $H=\frac1L\big(\sum_jv_jk_j^\top-\sum_c|c|\,\bar v_c\bar k_c^\top\big)=\sum_c\frac{|c|}{L}B_c$.
(ii) Multiply the expansion of $e^{x_{ij}}$ by $v_j=\bar v_c+\nu_j$ and sum over the cluster.
The terms $\sum_j\nu_j$ and $\bar v_c\sum_jx_{ij}$ vanish, $\sum_jx_{ij}\nu_j=\big(\sum_j\nu_j\delta_j^\top\big)q_i/\sqrt d=|c|\,B_cq_i/\sqrt d$, and $\frac12\bar v_c\sum_jx_{ij}^2$ is the variance term of \hyperref[res:A4]{Lemma~A.4}.
What remains is $\frac12\sum_jx_{ij}^2\nu_j+\sum_j\varrho(x_{ij})\,v_j$, which (A3) bounds by the stated amount. $\square$

\noindent\phantomsection\label{res:A6}\textbf{Proposition A.6 (The residual of a whole cluster is second order).}
\emph{Let $U_{ic}=\sum_{j\in c}u_{ij}$ be the error that compensation leaves on all members of cluster $c$ together.
Under (A3),}
\begin{equation*}
\begin{aligned}
U_{ic}={}&|c|\,\phi_{ic}\Big[\big(\bar v_c-y_i\big)\frac{q_i^\top\big(C_c-\operatorname{diag}(C_c)\big)q_i}{2d}\\
&+\frac{(B_c-H)\,q_i}{\sqrt d}\Big]+\phi_{ic}\big(\rho'_{ic}-y_i\,\rho_{ic}\big).
\end{aligned}
\end{equation*}
\emph{Under (A3$'$), moreover, $\|U_{ic}\|_2\le|c|\,\phi_{ic}\,V\big(3\eta^2+\eta^3e^{\eta}\big)$.}

\emph{Proof.}
$U_{ic}=\big(\sum_je^{s_{ij}}v_j-|c|\,a_{ic}\big)-y_i\big(\sum_je^{s_{ij}}-|c|\,\omega_{ic}\big)$.
Insert \hyperref[res:A5]{Lemma~A.5}(ii) and $|c|\,a_{ic}=|c|\,\phi_{ic}\big[\bar v_c(1+\gamma_{ic})+Hq_i/\sqrt d\big]$ in the first bracket, and \hyperref[res:A4]{Lemma~A.4} in the second.
For the bound, (A3$'$) gives $|x_{ij}|\le\eta$ and $\sum_pq_{ip}^2\delta_{jp}^2\le\|q_i\|_2^2\|\delta_j\|_2^2\le d\eta^2$, so $q_i^\top C_cq_i/(2d)$ and $q_i^\top\operatorname{diag}(C_c)q_i/(2d)$ both lie in $[0,\eta^2/2]$.
It also gives $\|B_cq_i\|_2/\sqrt d\le\eta^2V$ and $\|Hq_i\|_2/\sqrt d\le\eta^2V$, because $B_cq_i/\sqrt d$ averages $x_{ij}\nu_j$ and $H$ averages the $B_c$.
With $\|\bar v_c-y_i\|_2\le2V$ and $\|y_i\|_2\le V$, the bracket is at most $3\eta^2V$, and the remainders of Lemmas~\hyperref[res:A4]{A.4} and~\hyperref[res:A5]{A.5} add at most $|c|\,\eta^3V(1/2+e^{\eta}/3)\le|c|\,\eta^3e^{\eta}V$. $\square$

\hyperref[res:A6]{Proposition~A.6} says what the two corrections achieve.
On every cluster, compensation is exact up to three terms: the off-diagonal key covariance, the gap between the cluster's own cross-covariance $B_c$ and the head average $H$, and a remainder, of orders $\eta^2$, $\eta^2$, and $\eta^3$ under (A3$'$).
The two corrections of Eq.~\eqref{eq:comp} are therefore the second-order terms that a diagonal, head-shared model can capture.
A full covariance per cluster and a per-cluster $H_c$ remove the two second-order terms, and Table~\ref{tab:corrdiag} measures what these additions gain on saved attention inputs.

\subsection{Why Fixed Blocks Suffice}
\label{app:th-coh}

\noindent\phantomsection\label{res:A7}\textbf{Lemma A.7 (First-order cancellation).}
\emph{Under (A3$'$), the per-member error splits as $u_{ij}=u^{(1)}_{ij}+u^{(2)}_{ij}$ with $\|u^{(2)}_{ij}\|_2\le(3+e^{\eta})\,\phi_{ic}V\eta^2$ and the first-order part}
\begin{equation*}
u^{(1)}_{ij}=\phi_{ic}\big(\nu_j+(\bar v_c-y_i)\,x_{ij}\big),
\end{equation*}
\emph{which sums to zero over every cluster.
Consequently the first-order parts $r^{(1)}_{ib}=\sum_{j\in b}u^{(1)}_{ij}$ of the block residuals sum to zero over all blocks, and}
\begin{equation*}
\sum_{b\notin\mathcal S_g}r^{(1)}_{ib}=-\sum_{b\in\mathcal S_g}r^{(1)}_{ib}.
\end{equation*}

\emph{Proof.}
With $e^{s_{ij}}=\phi_{ic}e^{x_{ij}}$, $v_j=\bar v_c+\nu_j$, and $a_{ic}-y_i\omega_{ic}=\phi_{ic}(1+\gamma_{ic})(\bar v_c-y_i)+\phi_{ic}Hq_i/\sqrt d$, the zeroth-order terms $\phi_{ic}(\bar v_c-y_i)$ cancel and
\begin{equation*}
\begin{aligned}
u^{(2)}_{ij}=\phi_{ic}\Big[&x_{ij}\nu_j+\big(e^{x_{ij}}-1-x_{ij}\big)(v_j-y_i)\\
&-\gamma_{ic}(\bar v_c-y_i)-Hq_i/\sqrt d\,\Big].
\end{aligned}
\end{equation*}
Under (A3$'$) the four terms are at most $\eta^2V$, $\eta^2e^{\eta}V$, $\eta^2V$, and $\eta^2V$, by $|e^x-1-x|\le x^2e^{|x|}/2$, $\|v_j-y_i\|_2\le2V$, and the bounds in the proof of \hyperref[res:A6]{Proposition~A.6}.
The first-order part sums to zero because $\sum_{j\in c}\nu_j=0$ and $\sum_{j\in c}x_{ij}=0$.
Since $r_{ib}=\sum_{j\in b}u_{ij}$ and the blocks partition the keys, $\sum_b r^{(1)}_{ib}=\sum_c\sum_{j\in c}u^{(1)}_{ij}=0$. $\square$

Dropping a block has an analogous identity: the dropped residuals $r^{\mathrm{drop}}_{ib}=\sum_{j\in b}e^{s_{ij}}(v_j-y_i)$ also sum to zero over all blocks, because $\sum_je^{s_{ij}}(v_j-y_i)=N_i-y_iZ_i=0$.
What separates the two is not the sum over all blocks but how the residuals of the skipped blocks align with each other.
The next proposition makes this precise without any expansion.

\noindent\phantomsection\label{res:A8}\textbf{Proposition A.8 (Compensation subtracts the aligned part).}
\emph{Let $\bar z_{ic}=\frac{1}{|c|}\sum_{j\in c}e^{s_{ij}}(v_j-y_i)$ be the mean contribution of a member of cluster $c$, and $w_{ij}=e^{s_{ij}}(v_j-y_i)-\bar z_{ic}$ its deviation, which sums to zero over the cluster.
Then for every block,}
\begin{equation*}
r^{\mathrm{drop}}_{ib}=\underbrace{\sum_c n_{bc}\,\bar z_{ic}}_{\text{cluster means}}+\underbrace{\sum_{j\in b}w_{ij}}_{\text{deviations}},\qquad
r_{ib}=\underbrace{\sum_{j\in b}w_{ij}}_{\text{deviations}}+\underbrace{\sum_c n_{bc}\,\frac{U_{ic}}{|c|}}_{\text{cluster bias }\beta_{ib}} .
\end{equation*}

\emph{Proof.}
The first identity splits each member's contribution into $\bar z_{ic}$ and $w_{ij}$ and groups the means by cluster.
For the second, $r_{ib}=r^{\mathrm{drop}}_{ib}-\sum_cn_{bc}(a_{ic}-y_i\omega_{ic})$, and averaging $u_{ij}$ over the cluster gives $a_{ic}-y_i\omega_{ic}=\bar z_{ic}-U_{ic}/|c|$. $\square$

The cluster-mean term of a dropped block is a sum of the same vectors $\bar z_{ic}$ for every block that contains members of $c$.
Blocks with similar cluster composition therefore push the output in the same direction.
Compensation removes this term and replaces it by the cluster bias $\beta_{ib}$, built from the $U_{ic}/|c|$ that \hyperref[res:A6]{Proposition~A.6} shows to be of second order under (A3$'$) and that the variance correction shrinks in practice (Appendix~\ref{app:supp-resid}).
What remains is the deviation term, whose behavior the next lemma computes.

\noindent\phantomsection\label{res:A9}\textbf{Lemma A.9 (Sampling identities).}
\emph{Let a finite population $w_1,\dots,w_n$ of vectors, $n\ge2$, have $\sum_jw_j=0$ and $\tau^2=\frac1n\sum_j\|w_j\|_2^2$, and let disjoint subsets $A$ and $A'$ of sizes $a$ and $a'$ be drawn by a uniformly random permutation.
Then}
\begin{equation*}
\begin{gathered}
\mathbb E\sum_{j\in A}w_j=0,\qquad \mathbb E\Big\|\sum_{j\in A}w_j\Big\|_2^2=\frac{a(n-a)}{n-1}\,\tau^2,\\
\mathbb E\Big\langle\sum_{j\in A}w_j,\sum_{j\in A'}w_j\Big\rangle=-\frac{a\,a'}{n-1}\,\tau^2 .
\end{gathered}
\end{equation*}

\emph{Proof.}
Each $w_j$ belongs to $A$ with probability $a/n$, so the mean is $\frac an\sum_jw_j=0$.
For the second moment, $\mathbb E\|\sum_Aw_j\|^2=\sum_j\Pr[j\in A]\,\|w_j\|^2+\sum_{j\ne j'}\Pr[j,j'\in A]\,\langle w_j,w_{j'}\rangle$ with $\Pr[j\in A]=a/n$ and $\Pr[j,j'\in A]=\frac{a(a-1)}{n(n-1)}$.
Because $\sum_{j\ne j'}\langle w_j,w_{j'}\rangle=\|\sum_jw_j\|^2-\sum_j\|w_j\|^2=-n\tau^2$, this equals $a\tau^2-\frac{a(a-1)}{n-1}\tau^2=\frac{a(n-a)}{n-1}\tau^2$.
Applying the second identity to $A$, $A'$, and $A\cup A'$ and expanding $\|\sum_{A\cup A'}w_j\|^2$ gives the cross term:
\begin{equation*}
\begin{aligned}
2\,\mathbb E\big\langle\textstyle\sum_Aw_j,\sum_{A'}w_j\big\rangle
&=\frac{\tau^2}{n-1}\big[(a+a')(n-a-a')\\
&\qquad-a(n-a)-a'(n-a')\big]\\
&=-\frac{2aa'}{n-1}\,\tau^2. \qquad\square
\end{aligned}
\end{equation*}

\noindent\phantomsection\label{res:A10}\textbf{Theorem A.10 (Deviation terms add like random errors).}
\emph{Under (A4), let $\mathcal K$ be a set of skipped blocks fixed independently of the permutation, let $\tau_{ic}^2=\frac{1}{|c|}\sum_{j\in c}\|w_{ij}\|_2^2$ for clusters with $|c|\ge2$ (a singleton has $w_{ij}=0$ and drops out), let $m_c=\sum_{b\in\mathcal K}n_{bc}$ count the members of $c$ in $\mathcal K$, and let $D_b=\sum_{j\in b}w_{ij}$ be the deviation term of \hyperref[res:A8]{Proposition~A.8}.
Then $\mathbb E\,D_b=0$ and}
\begin{equation*}
\begin{aligned}
\mathbb E\Big\|\sum_{b\in\mathcal K}D_b\Big\|_2^2&=\sum_c\frac{m_c(|c|-m_c)}{|c|-1}\,\tau_{ic}^2\\
&\le\sum_c\sum_{b\in\mathcal K}\frac{n_{bc}(|c|-n_{bc})}{|c|-1}\,\tau_{ic}^2\\
&=\sum_{b\in\mathcal K}\mathbb E\,\|D_b\|_2^2 .
\end{aligned}
\end{equation*}
\emph{The expected coherence ratio $\bar\kappa_D=\mathbb E\|\sum_{b\in\mathcal K}D_b\|_2^2/\sum_{b\in\mathcal K}\mathbb E\|D_b\|_2^2$ is therefore at most one, and distinct blocks are negatively correlated:
$\mathbb E\langle D_b,D_{b'}\rangle=-\sum_c n_{bc}n_{b'c}\tau_{ic}^2/(|c|-1)\le0$.
With the cluster bias $\beta_{ib}$ of \hyperref[res:A8]{Proposition~A.8}, the full residual satisfies}
\begin{equation*}
\mathbb E\Big\|\sum_{b\in\mathcal K}r_{ib}\Big\|_2^2=\mathbb E\Big\|\sum_{b\in\mathcal K}D_b\Big\|_2^2+\Big\|\sum_c\frac{m_c}{|c|}\,U_{ic}\Big\|_2^2 .
\end{equation*}

\emph{Proof.}
By (A4), the members of each cluster that fall into each block form disjoint random subsets of sizes $n_{bc}$, and different clusters are independent.
Writing $D_b=\sum_cD_{bc}$ with $D_{bc}=\sum_{j\in b\cap c}w_{ij}$, every $D_{bc}$ has zero mean (\hyperref[res:A9]{Lemma~A.9}), so cross terms between different clusters vanish.
Within a cluster, $\sum_{b\in\mathcal K}D_{bc}$ is the sum over a random subset of size $m_c$, and \hyperref[res:A9]{Lemma~A.9} gives the left side.
Each block alone is a subset of size $n_{bc}$, which gives the right side and the cross term.
The inequality holds cluster by cluster because $m_c\ge n_{bc}$ for every $b\in\mathcal K$:
\begin{equation*}
m_c(|c|-m_c)=\sum_{b\in\mathcal K}n_{bc}(|c|-m_c)\le\sum_{b\in\mathcal K}n_{bc}(|c|-n_{bc}).
\end{equation*}
For the full residual, $r_{ib}=D_b+\beta_{ib}$, and $\sum_{b\in\mathcal K}\beta_{ib}=\sum_cm_cU_{ic}/|c|$ depends only on the counts because $U_{ic}$ sums over the whole cluster, so the cross term vanishes with $\mathbb E\,D_b=0$. $\square$

\noindent\phantomsection\label{res:A11}\textbf{Corollary A.11 (The selection enters only through counts).}
\emph{Under (A4), for exact sets fixed independently of the permutation, the expected deviation error $\mathbb E\|\sum_{b\notin\mathcal S_g}D_b\|_2^2$ depends on the exact set $\mathcal S_g$ only through the numbers $m_c=\sum_{b\notin\mathcal S_g}n_{bc}$ of skipped members of each cluster.
Two exact sets that skip the same number of members of every cluster have the same expected deviation error, whichever blocks they contain.}

\emph{Proof.}
The left side of \hyperref[res:A10]{Theorem~A.10} is $\sum_c m_c(|c|-m_c)\tau_{ic}^2/(|c|-1)$, which contains $\mathcal S_g$ only through the $m_c$. $\square$

Blocks fixed offline never read the keys of a new input, so they meet this independence by construction.
Among such sets, a change of blocks under the same budget changes the deviation error only through the $m_c$, and with hundreds of clusters spread over many blocks each $m_c$ moves by little.
A selection made anew at every call reads the current keys and can correlate with the deviations, which moves the error at equal counts in either direction, and Table~\ref{tab:selection} measures the net effect: under one compensator, routing by attention mass at every call reaches 24.65\,dB against 24.70\,dB for fixed blocks on Cosmos-Predict2.5.

\noindent\phantomsection\label{res:A12}\textbf{Corollary A.12 (Dropped residuals align).}
\emph{Under (A4), the dropped residuals have mean $\mu_b=\sum_cn_{bc}\bar z_{ic}$, and for every set $\mathcal K$ fixed independently of the permutation}
\begin{equation*}
\mathbb E\Big\|\sum_{b\in\mathcal K}r^{\mathrm{drop}}_{ib}\Big\|_2^2=\Big\|\sum_{b\in\mathcal K}\mu_b\Big\|_2^2+\mathbb E\Big\|\sum_{b\in\mathcal K}D_b\Big\|_2^2 .
\end{equation*}
\emph{If the skipped blocks share one cluster composition, $n_{bc}=\alpha_bp_c$ with $\sum_cp_c\bar z_{ic}\ne0$, the cluster-mean terms are parallel and their coherence ratio $\|\sum_b\mu_b\|^2/\sum_b\|\mu_b\|^2=(\sum_b\alpha_b)^2/\sum_b\alpha_b^2$ reaches $|\mathcal K|$ for blocks of equal size.}

\emph{Proof.}
By \hyperref[res:A8]{Proposition~A.8}, $r^{\mathrm{drop}}_{ib}=\mu_b+D_b$ with $\mathbb E\,D_b=0$, so the cross term between the two parts vanishes in expectation.
Under a shared composition, $\mu_b=\alpha_b\sum_cp_c\bar z_{ic}$ for every $b$. $\square$

\hyperref[res:A10]{Theorem~A.10} and \hyperref[res:A12]{Corollary~A.12} give the two regimes of the coherence ratio.
Dropped blocks carry a cluster-mean term that adds coherently over the blocks that share clusters, so its coherence ratio grows with the number of such blocks.
Compensated blocks carry only the deviation term, whose expected coherence ratio $\bar\kappa_D$ is at most one, plus the cluster bias, of second order under (A3$'$) and the coherent part of the residual in practice (Appendix~\ref{app:supp-resid}).
This is the mechanism of subtractive dither in quantization~\cite{schuchman1964dither}: subtracting a known prediction leaves zero-mean errors that add like random ones.
The measured medians follow these regimes, from 4.69, 5.03, and 10.37 without compensation to 1.61, 1.65, and 2.69 with both corrections on DiffusionRenderer, Wan2.1-14B, and Cosmos-Predict2.5 (Table~\ref{tab:kappa}).
\hyperref[res:A10]{Theorem~A.10} holds for every set of blocks fixed independently of the membership, so blocks fixed offline by position keep $\bar\kappa_D$ at most one, as observed in Section~\ref{sec:decoup}.

\subsection{Choosing the Exact Blocks}
\label{app:th-sel}

\noindent\phantomsection\label{res:A13}\textbf{Proposition A.13 (Residual ranking).}
\emph{Fix a query tile and a budget of $K$ exact blocks besides its diagonal block $b_g$, and take expectations over calibration inputs.
If the residuals of distinct blocks are orthogonal in expectation, $\mathbb E\langle r_{ib},r_{ib'}\rangle=0$ for $b\ne b'$, the set of $K$ blocks that minimizes the expected unnormalized error $\mathbb E\|\sum_{b\notin\mathcal S}r_{ib}\|_2^2$, summed over the queries of the tile, consists of the $K$ blocks with the largest $\sum_{i\in g}\mathbb E\|r_{ib}\|_2^2$.
If instead $\mathbb E\langle r_{ib},r_{ib'}\rangle\le0$, as for the deviation terms of \hyperref[res:A10]{Theorem~A.10}, the same set minimizes an upper bound on this error.}

\emph{Proof.}
With orthogonal residuals, $\mathbb E\|\sum_{b\notin\mathcal S}r_{ib}\|^2=\sum_b\mathbb E\|r_{ib}\|^2-\sum_{b\in\mathcal S}\mathbb E\|r_{ib}\|^2$ for every query, and summing over the query tile leaves a total that the largest $K$ terms minimize.
With nonpositive cross moments, $\mathbb E\|\sum_{b\notin\mathcal S}r_{ib}\|^2\le\sum_{b\notin\mathcal S}\mathbb E\|r_{ib}\|^2$, and the same choice minimizes the right side. $\square$

Equation~\eqref{eq:dtable} implements this rule with each query weighted by $1/(Z_i^2\|y_i\|_2^2)$, for queries with $y_i\ne0$.
The weight is the scale of the relative error of \hyperref[res:A2]{Corollary~A.2} with the dense mass $Z_i$ in place of $\hat Z_i$, which depends on the exact set, so the ranking minimizes a residual-energy proxy of the relative error.
The rule ranks blocks by how badly compensation handles them, not by the attention they hold.
A block with large attention mass that compensation already predicts well has a small residual and stays skipped.

\subsection{Relation to Difference Estimators}
\label{app:th-diff}

The decoupled estimator has the form of a difference estimator from survey sampling~\cite{sarndal1992model}: a model predicts every term cheaply, and the exact values correct the prediction on a chosen subset.
Scatterbrain~\cite{chen2021scatterbrain} and control-variate attention~\cite{zheng2023eva} use the same structure with a support chosen for each input, whereas DecoupAttn keeps the clusters and the exact blocks as two groupings joined by counts and fixes the blocks offline.
This view explains why compensation changes what the selection has to capture.

\noindent\phantomsection\label{res:A14}\textbf{Proposition A.14 (Random exact blocks).}
\emph{Let every block be made exact independently with probability $\pi_b\in(0,1]$, let $I_b$ indicate the event, and weight each correction of the numerator and the mass by $1/\pi_b$:}
\begin{equation*}
(\hat N_i^{\mathrm{HT}},\hat Z_i^{\mathrm{HT}})=\sum_c|c|\,(a_{ic},\omega_{ic})+\sum_b\frac{I_b}{\pi_b}\,(\Delta N_{ib},\Delta Z_{ib}).
\end{equation*}
\emph{Then $\mathbb E\,\hat N_i^{\mathrm{HT}}=N_i$, $\mathbb E\,\hat Z_i^{\mathrm{HT}}=Z_i$, the output $\hat y_i^{\mathrm{HT}}=\hat N_i^{\mathrm{HT}}/\hat Z_i^{\mathrm{HT}}$ has the error $\hat y_i^{\mathrm{HT}}-y_i=\sum_b(I_b/\pi_b-1)\,r_{ib}/\hat Z_i^{\mathrm{HT}}$, and with the mass at its mean $Z_i$}
\begin{equation*}
\mathbb E\Big\|\frac{1}{Z_i}\sum_b\Big(\frac{I_b}{\pi_b}-1\Big)r_{ib}\Big\|_2^2=\frac{1}{Z_i^2}\sum_b\frac{1-\pi_b}{\pi_b}\,\|r_{ib}\|_2^2 .
\end{equation*}

\emph{Proof.}
By \hyperref[res:prop1]{Proposition~1} with every block exact, $N_i=\sum_c|c|\,a_{ic}+\sum_b\Delta N_{ib}$ and $Z_i=\sum_c|c|\,\omega_{ic}+\sum_b\Delta Z_{ib}$.
Hence $\hat N_i^{\mathrm{HT}}-N_i=\sum_b(I_b/\pi_b-1)\Delta N_{ib}$ and $\hat Z_i^{\mathrm{HT}}-Z_i=\sum_b(I_b/\pi_b-1)\Delta Z_{ib}$, and with $N_i=y_iZ_i$ the error $(\hat N_i^{\mathrm{HT}}-y_i\hat Z_i^{\mathrm{HT}})/\hat Z_i^{\mathrm{HT}}$ collects $\Delta N_{ib}-y_i\Delta Z_{ib}=r_{ib}$.
The weights $I_b/\pi_b-1$ have zero mean, are independent, and have $\mathbb E(I_b/\pi_b-1)^2=(1-\pi_b)/\pi_b$. $\square$

Without compensation the same estimator has $a_{ic}=\omega_{ic}=0$, so $r_{ib}$ becomes the dropped residual $N_{ib}-y_iZ_{ib}$ and the error is driven by the attention a block holds.
With compensation it is driven by the residual $r_{ib}$ that the clusters leave, which is the quantity that Eq.~\eqref{eq:dtable} ranks and Distortion Budget prices.

\noindent\phantomsection\label{res:A15}\textbf{Proposition A.15 (Optimal inclusion follows the residual).}
\emph{Let every $\|r_{ib}\|_2$ be positive.
Among inclusion probabilities with $0<\pi_b\le1$ and $\sum_b\pi_b=K$ exact blocks in expectation, the error of \hyperref[res:A14]{Proposition~A.14} is minimized by $\pi_b=\min\big(1,\lambda\|r_{ib}\|_2\big)$, with $\lambda$ chosen to meet the budget.}

\emph{Proof.}
For the blocks with $\pi_b<1$, Cauchy--Schwarz gives $\big(\sum_b\|r_{ib}\|\big)^2\le\sum_b\pi_b\cdot\sum_b\|r_{ib}\|^2/\pi_b$, with equality exactly when $\pi_b\propto\|r_{ib}\|$.
The error is $Z_i^{-2}\big(\sum_b\|r_{ib}\|^2/\pi_b-\sum_b\|r_{ib}\|^2\big)$, so it is minimized by the proportional rule.
When the rule asks for $\pi_b>1$, the constraint is active, those blocks become certain, and the rule applies to the remaining budget, which is the stated form. $\square$

A block with $r_{ib}=0$ adds no error at any $\pi_b>0$, so with such blocks the optimum is approached as their $\pi_b\downarrow0$ and the rule applies to the others.

The blocks with the largest residuals become certain, which is the residual ranking of \hyperref[res:A13]{Proposition~A.13}.
DecoupAttn makes this rule deterministic, $\pi_b\in\{0,1\}$ without the weights $1/\pi_b$: the blocks with the largest residuals are exact, and the others are left to compensation.
Its error is then the deterministic sum of the remaining residuals (\hyperref[res:prop1]{Proposition~1}), and \hyperref[res:A10]{Theorem~A.10} shows that the deviation part of this sum adds like random errors.

\subsection{The Plan Objective (Proposition 2)}
\label{app:prop2}

\noindent\phantomsection\label{res:A16}\textbf{Lemma A.16 (Projected distortion).}
\emph{For a head output error $A=\hat O_u-O_u$ and the head's slice $W_{o,h}$ of the output projection,}
\begin{equation*}
\big\|A\,W_{o,h}^\top\big\|_F^2=\big\langle W_{o,h}^\top W_{o,h},\,A^\top A\big\rangle,
\end{equation*}
\emph{so $\varepsilon_u$ of Eq.~\eqref{eq:distortion} is the expected squared error that the head adds to the residual stream, and it vanishes exactly when every row of the error lies in the null space of $W_{o,h}$ almost surely.}

\emph{Proof.}
$\|AW^\top\|_F^2=\operatorname{tr}(WA^\top AW^\top)=\operatorname{tr}(A^\top A\,W^\top W)$ by the cyclic property of the trace.
The value is zero exactly when $AW^\top=0$. $\square$

The projected errors of one call's heads add in the residual stream, with cross terms as in \hyperref[res:A17]{Lemma~A.17}.

\noindent\phantomsection\label{res:A17}\textbf{Lemma A.17 (First-order additivity).}
\emph{Let the final latent be $\hat x_0=\Phi(e)$, where $e=(e_u)_u$ collects the projected head errors injected at the units and $\Phi(0)$ is the dense output.
If $\Phi$ is twice continuously differentiable with $\|\nabla^2\Phi\|\le\beta$ on the segment from $0$ to $e$, then with $G_u$ the Jacobian of $\Phi$ with respect to $e_u$ at zero,}
\begin{equation*}
\hat x_0-x_0=\sum_uG_ue_u+R,\qquad \|R\|_2\le\frac\beta2\|e\|_2^2 .
\end{equation*}
\emph{If the propagated errors of distinct units, and of the sampler steps that one unit stands for, are orthogonal in expectation, $\mathbb E\langle G_ue_u,G_{u'}e_{u'}\rangle=0$ for $u\ne u'$, and the joint Jacobian $G=(G_u)_u$ has $\|G\|\le M$, then}
\begin{equation*}
\begin{aligned}
&\Big|\mathbb E\|\hat x_0-x_0\|_2^2-\sum_u\mathbb E\|G_ue_u\|_2^2\Big|\\
&\qquad\le M\beta\,\mathbb E\|e\|_2^3+\tfrac{\beta^2}{4}\,\mathbb E\|e\|_2^4 .
\end{aligned}
\end{equation*}

\emph{Proof.}
The first display is Taylor's theorem with the integral remainder $R=\int_0^1(1-s)\,\nabla^2\Phi(se)[e,e]\,ds$.
Squaring it gives $\|Ge\|^2+2\langle Ge,R\rangle+\|R\|^2$, the cross terms between units in $\|Ge\|^2$ vanish in expectation by assumption, and $2|\langle Ge,R\rangle|\le M\beta\|e\|^3$ and $\|R\|^2\le\beta^2\|e\|^4/4$. $\square$

Writing $\mathbb E\|G_ue_u\|^2=J_u(o,\eta)\,w_u\,\varepsilon_u(o,\eta)$ defines the sensitivity of a unit under each option as its gain from projected error to final-latent error, with $w_u$ counting the sampler steps the unit stands for.
The gain depends on the option only through the direction of the error, and Distortion Budget prices all options of a unit with the one gain $J_u=J_u(\textsc{Static}_{\theta_0},\textsc{Full})$ measured at the reference option of Eq.~\eqref{eq:sens}, so the gaps $J_u(o,\eta)-J_u$ enter the modeling error $\delta_{\mathrm m}$ below.
\hyperref[res:A17]{Lemma~A.17} then turns the final-latent error of Eq.~\eqref{eq:problem} into the first line of Eq.~\eqref{eq:alloc}.
Table~\ref{tab:additive} measures directly the cross terms that the lemma sets to zero.

\noindent\phantomsection\label{res:A18}\textbf{Lemma A.18 (The sensitivity estimator).}
\emph{Suppose the units of a group $\mathcal G$ share one gain $J_{\mathcal G}$, their propagated errors are orthogonal as in \hyperref[res:A17]{Lemma~A.17}, the run and the dense output differ, besides the injected errors, by a zero-mean numerical error $\xi$, the run-to-run difference of the same dense attention backend, independent of the injected errors and with $\mathbb E\|\xi\|_2^2=\mathrm{MSE}_0$, and the group's summed distortion is positive.
Then, to first order, Eq.~\eqref{eq:sens} returns $J_{\mathcal G}$.}

\emph{Proof.}
The run of the group gives $\hat x_0-x_0=\sum_{u\in\mathcal G}G_ue_u+\xi$ up to second order.
Independence and zero mean remove the cross terms with $\xi$, so
\begin{equation*}
\mathrm{MSE}_{\mathcal G}=\sum_{u\in\mathcal G}\mathbb E\|G_ue_u\|_2^2+\mathrm{MSE}_0=J_{\mathcal G}\sum_{u\in\mathcal G}w_u\,\varepsilon_u+\mathrm{MSE}_0 .
\end{equation*}
Subtracting $\mathrm{MSE}_0$ and dividing by the summed distortion leaves $J_{\mathcal G}$. $\square$

Without the subtraction, the estimate would carry the bias $\mathrm{MSE}_0/\sum_uw_u\varepsilon_u$, which is largest exactly where two dense attention backends already disagree, on the prompt-driven few-step models.
In this model $\mathrm{MSE}_{\mathcal G}\ge\mathrm{MSE}_0$ and the estimate is nonnegative, so a negative cell of Table~\ref{tab:refstab} marks a group whose injected error lies below the run-to-run variation of the two measured errors.

\noindent\phantomsection\label{res:A19}\textbf{Lemma A.19 (Measuring one unit at a time).}
\emph{Let the projected error of a unit be $e_u=E_u(z)$, a function of the unit's input $z$, and let the distortion be measured at the dense input $z_{\mathrm d}$, as in Section~\ref{sec:budget}.
If $E_u$ is $\ell_u$-Lipschitz on the inputs that occur, and a sparse run perturbs the input to $z=z_{\mathrm d}+\Delta z$, then}
\begin{equation*}
\Big|\big(\mathbb E\|E_u(z)\|_2^2\big)^{1/2}-\varepsilon_u^{1/2}\Big|\le\ell_u\Delta,\qquad \Delta=\big(\mathbb E\|\Delta z\|_2^2\big)^{1/2},
\end{equation*}
\emph{and therefore $\big|\mathbb E\|E_u(z)\|_2^2-\varepsilon_u\big|\le2\,\varepsilon_u^{1/2}\ell_u\Delta+\ell_u^2\Delta^2$.}

\emph{Proof.}
$\|E_u(z)-E_u(z_{\mathrm d})\|\le\ell_u\|\Delta z\|$ pointwise, and the triangle inequality in $L^2$ gives the first bound.
The second follows from $a^2-b^2=(a-b)(a+b)$ with $a+b\le2b+\ell_u\Delta$. $\square$

The deviation of the measured distortion is thus led by the product of the unit's own error $\varepsilon_u^{1/2}$ and the upstream error $\ell_u\Delta$, where $\Delta z$ is the accumulated error of the units upstream and $\ell_u$ bounds how fast the unit's error changes with its input.
Measuring every unit at the dense input therefore captures its distortion up to a product of two errors whenever $\ell_u$ is bounded.
It also keeps the measurement of a unit independent of the plan, so the distortion table is built once and every candidate plan is priced from it.

\noindent\textbf{Cost of solving.}
For one multiplier, the per-call minimization visits every unit and option once under each of the two estimators, which costs $O(|\mathcal U|\,|\mathcal O|)$ operations on a table that is already measured.
Bisection repeats this $B_\mu$ times and the greedy pass sorts the candidate changes once, for $O\big(|\mathcal U|\,|\mathcal O|\,B_\mu+|\mathcal U|\log|\mathcal U|\big)$ in total.
This is why plan solving takes under a minute in Table~\ref{tab:offline}, while measuring $\varepsilon_u$ and $J_u$ takes the GPU hours.

\noindent\textbf{Proof of Proposition 2.}
Plans $\pi$ (\hyperref[res:def2]{Definition~2}) range over a finite set, $F(\pi)=\sum_uw_uJ_u\varepsilon_u(o_u,\eta_{tl})$ is the priced distortion, and $X(\pi)=\sum_uw_uX_u(o_u)/w_\Sigma$ the density of Eq.~\eqref{eq:config}, with $w_\Sigma=\sum_uw_u$.
For a multiplier $\mu\ge0$, let $\pi_\mu$ be a global minimizer of the Lagrangian of Eq.~\eqref{eq:alloc}, $\mathcal L_\mu(\pi)=F(\pi)+\mu\,w_\Sigma X(\pi)$.

\emph{Exact minimization.}
$\mathcal L_\mu$ is a sum over calls $(t,l)$ of $\sum_hw_u\big(J_u\varepsilon_u(o_u,\eta_{tl})+\mu X_u(o_u)\big)$, and the estimator couples only the heads of one call.
Minimizing each call over $\eta_{tl}$ of $\sum_h\min_{o_u}(\cdot)$ therefore minimizes $\mathcal L_\mu$ exactly, which is the procedure of Section~\ref{sec:budget}.
Adding the two minimality inequalities shows that any minimizers at $\mu_1<\mu_2$ have $X(\pi_{\mu_2})\le X(\pi_{\mu_1})$, whichever ties are taken, so bisection on $\mu$ is well defined.

\emph{Optimality at its own density.}
Let $\pi$ be any plan with $X(\pi)\le X(\pi_\mu)$.
Because $\pi_\mu$ minimizes $\mathcal L_\mu$,
\begin{equation*}
F(\pi_\mu)\le F(\pi)+\mu\,w_\Sigma\big(X(\pi)-X(\pi_\mu)\big)\le F(\pi),
\end{equation*}
so $\pi_\mu$ has the lowest priced distortion among all plans at most as dense as itself~\cite{everett1963lagrange}.
Bisection returns the feasible $\pi_\mu$ of largest density, and the greedy pass then lowers $F$ within the remaining slack $\rho-X(\pi_\mu)$, which \hyperref[res:A20]{Lemma~A.20}(i) covers. $\square$

\noindent\phantomsection\label{res:A20}\textbf{Lemma A.20 (Gap of the relaxation).}
\emph{Let $F^\star(\rho)$ be the lowest priced distortion over plans with $X(\pi)\le\rho$.
(i) If bisection returns a feasible exact minimizer $\pi_{\mathrm b}$ of $\mathcal L_\mu$ for some $\mu\ge0$, then}
\begin{equation*}
F(\pi_{\mathrm b})-F^\star(\rho)\le\mu\,w_\Sigma\big(\rho-X(\pi_{\mathrm b})\big),
\end{equation*}
\emph{and every plan that the greedy pass reaches from $\pi_{\mathrm b}$ by feasible changes that do not raise $F$ keeps this bound.
(ii) Let $W_{\max}$ be the largest total weight $\sum_hw_u$ of one call and $\mu^\star$ the multiplier at which the minimizers of $\mathcal L_\mu$ cross the budget.
If, at $\mu^\star$, the minimizers are switched one call at a time, in any order, from an infeasible one toward a feasible one and the first feasible plan is kept, the bound of (i) is below $\mu^\star W_{\max}$.}

\emph{Proof.}
(i) For every feasible $\pi$, minimality of $\pi_{\mathrm b}$ gives $F(\pi)\ge F(\pi_{\mathrm b})+\mu\,w_\Sigma\big(X(\pi_{\mathrm b})-X(\pi)\big)\ge F(\pi_{\mathrm b})-\mu\,w_\Sigma\big(\rho-X(\pi_{\mathrm b})\big)$, since $X(\pi)\le\rho$.
Taking the minimum over feasible $\pi$ gives the bound, and a later plan with no larger $F$ inherits it.
(ii) If a minimizer of $\mathcal L_0=F$ is feasible, it attains $F^\star(\rho)$ and the gap is zero.
Otherwise, as a function of $\mu$, the set of minimizers of $\mathcal L_\mu$ changes only at finitely many breakpoints, and at the breakpoint $\mu^\star$ it contains a plan $\pi_-$ with $X(\pi_-)>\rho$ and a plan $\pi_+$ with $X(\pi_+)\le\rho$.
Because $\mathcal L_{\mu^\star}$ separates over calls, every plan that takes a minimizing choice in each call is again a minimizer, so $\pi_-$ turns into $\pi_+$ one call at a time through minimizers.
The first feasible plan of this chain follows an infeasible one that differs from it in one call, so its slack $\rho-X$ is below $W_{\max}/w_\Sigma$, and (i) bounds its gap by less than $\mu^\star W_{\max}$. $\square$

The bound of (i) is known once the plan is returned, and (ii) caps it at the price of a single call's density for a solver that switches tied minimizers one call at a time.
Our solver keeps the whole minimizer at the feasible end of bisection, and on the deployed plans of the eight models it solves the bound of (i) is at most 0.54\% of the priced distortion (Cosmos-Predict2.5) and at most 0.094\% on the others, with at most 0.026\% left after the greedy pass.
With every step, layer, and head of a video in the plan, one call holds a small share of $w_\Sigma$, so the relaxation is nearly tight.

The gap concerns the priced distortion $F$, and it carries over to the final-latent error $D(\pi)=\mathbb E_x\|\hat x_0(\pi)-x_0\|_2^2$ of Eq.~\eqref{eq:problem} through the modeling error.
If $|D(\pi)-F(\pi)|\le\delta_{\mathrm m}$ on every feasible plan and $F(\hat\pi)\le F^\star(\rho)+\epsilon_{\mathrm o}$, then for every feasible $\pi$
\begin{equation*}
D(\hat\pi)\le F(\hat\pi)+\delta_{\mathrm m}\le F(\pi)+\epsilon_{\mathrm o}+\delta_{\mathrm m}\le D(\pi)+\epsilon_{\mathrm o}+2\delta_{\mathrm m},
\end{equation*}
so $\hat\pi$ is within $\epsilon_{\mathrm o}+2\delta_{\mathrm m}$ of the best feasible plan in final-latent error.
Lemmas~\hyperref[res:A17]{A.17}--\hyperref[res:A19]{A.19} trace where $\delta_{\mathrm m}$ comes from: the cross terms between units, the shared gain of a group and of the options of a unit, and the measurement at the dense input.

\subsection{Reassign Cache (Proposition 3)}
\label{app:prop3}

\noindent\textbf{Proof of Proposition 3.}
With the centroids $m_c$ fixed, the total squared distance $\sum_j\|k_j^{(t)}-m_{c(j)}\|_2^2$ is a sum of one term per key, and Eq.~\eqref{eq:reassign} minimizes each term separately.
Hence, for any other assignment $c'$, including the frozen assignment of the first sparse step,
\begin{equation*}
\sum_j\big\|k_j^{(t)}-m_{c_t(j)}\big\|_2^2\le\sum_j\big\|k_j^{(t)}-m_{c'(j)}\big\|_2^2,
\end{equation*}
which is the assignment step of Lloyd's algorithm~\cite{lloyd1982least}.
The exact blocks $\mathcal S_g$ come from the offline ranking of Eq.~\eqref{eq:dtable}, and the options and estimators from the offline solution of Eq.~\eqref{eq:alloc}.
Neither reads the assignments at inference, so reassignment leaves both unchanged. $\square$

\noindent\phantomsection\label{res:A21}\textbf{Lemma A.21 (Monotone chain with recomputed statistics).}
\emph{Let $c_t$ be the assignment after reassignment at step $t$, $\bar k^{(t)}_c$ the means of the reassigned clusters computed from their current keys, and $c_f$ the frozen assignment.
Then}
\begin{equation*}
\begin{aligned}
\sum_j\big\|k_j^{(t)}-\bar k^{(t)}_{c_t(j)}\big\|_2^2&\le\sum_j\big\|k_j^{(t)}-m_{c_t(j)}\big\|_2^2\\
&\le\sum_j\big\|k_j^{(t)}-m_{c_f(j)}\big\|_2^2 .
\end{aligned}
\end{equation*}

\emph{Proof.}
Within every cluster the mean minimizes the sum of squared distances, which gives the first inequality, and \hyperref[res:prop3]{Proposition~3} gives the second. $\square$

The left side is the within-cluster spread $\sum_c|c|\operatorname{tr}(C_c)$ that enters the compensation error.
By \hyperref[res:A4]{Lemma~A.4}, the second-order mass term of a cluster is $|c|\,\phi_{ic}\,q_i^\top C_cq_i/(2d)$, and for Gaussian queries $q_i\sim\mathcal N(0,I)$ its mean is $|c|\,e^{\|a_c\|^2/2}\big(\operatorname{tr}C_c+a_c^\top C_ca_c\big)/(2d)$ with $a_c=\bar k_c/\sqrt d$, which grows with the spread $\operatorname{tr}C_c$ and with its part along the centroid.
Reassignment thus lowers an upper bound on the spread, the distance to the cached centroids, at the cost of one product between the keys and the cached centroids, $LCd$ multiply--adds, followed by an argmin.
A cluster that loses all of its members has $|c|=n_{gc}=0$ and drops out of the estimator and of $H$, and its cached centroid stays available to later assignments.
Reassignment with recomputed statistics is one step of $k$-means whose centroids for the next assignment stay at the cached $m_c$, so the assignments cannot drift away from the clusters that the plan was priced on.
\hyperref[res:prop3]{Proposition~3} concerns the distance to the cached centroids, and the Frozen ablation, which keeps the assignments and recomputes the cluster statistics, measures the quality gain of reassignment directly (Tables~\ref{tab:abl} and~\ref{tab:cache}).

\subsection{Cost of One Call}
\label{app:th-cost}

Count multiply--adds per head for $L$ tokens, head dimension $d$, $C$ clusters, and density $X$ (block padding included).
Dense attention costs $2L^2d$, for $QK^\top$ and for the weighted sum of values.
The exact part costs $2XL^2d$.
The compensation computes the centroid logits $q_i^\top\bar k_c$ for every query and cluster ($LCd$), the variance corrections $\sum_pq_{ip}^2\sigma_{cp}^2$ ($LCd$), the numerator $\sum_cn_{gc}a_{ic}$ ($LCd$), and the products $Hq_i$ ($Ld^2$).
The statistics $\bar k_c$, $\bar v_c$, $\sigma_c^2$, and $H$ cost $O(Ld+Ld^2)$, reassignment costs $LCd$, and the counts $n_{gc}$ are bookkeeping over the $N_B\times C$ table of $n_{bc}$.
Relative to dense attention, one call therefore costs
\begin{equation*}
r=X+\big(2+\mathbf 1_{\mathrm F}+\mathbf 1_{\mathrm R}\big)\frac{C}{2L}+\mathbf 1_{\mathrm F}\,\frac{d}{L},
\end{equation*}
where $\mathbf 1_{\mathrm F}=1$ for the \textsc{Full} estimator, which adds the variance corrections and $H$, and $\mathbf 1_{\mathrm R}=1$ when Reassign Cache reassigns.
On Cosmos-Predict2.5, the latent has $L=24\cdot44\cdot80=84{,}480$ tokens, with $C=880$ and $d=128$, so for the full method the two overhead terms add about $0.021$ and $0.0015$ to $X$.
In multiply--adds, compensation, counts, and reassignment thus cost about one tenth of the exact part at $X=0.20$, and the $k$-means of the first sparse step adds $I\cdot LCd$ once per video for $I$ rounds.
If attention takes a share $f$ of the latency and runs $r_{\mathrm{time}}$ times as long as dense attention, the speedup is
\begin{equation*}
S=\frac{1}{(1-f)+f\,r_{\mathrm{time}}},
\end{equation*}
where the measured time ratio $r_{\mathrm{time}}$ adds to $r$ the memory traffic and kernel launches that multiply--adds leave out.
The share $f$ differs across models, which is why one density gives different speedups in Table~\ref{tab:main}.
The measured components of Table~\ref{tab:timing} map onto these terms: \emph{Exact} is $2XL^2d$, \emph{Statistics}, \emph{Counts}, and \emph{Merge} are the compensation, \emph{Assign} is reassignment, and \emph{$k$-means} is the one-time clustering.

\subsection{Error Propagation Through the Sampler}
\label{app:th-prop}

The sensitivity $J_u$ measures how much of a unit's error reaches the final latent.
This section shows where its spread across steps and models comes from, and why the disagreement between two dense attention backends probes it.

\noindent\phantomsection\label{res:A22}\textbf{Lemma A.22 (Amplification along the sampler).}
\emph{Let the sampler update the latent by $x_{s-1}=\Psi_s(x_s)$ for $s=T,\dots,1$, where $\Psi_s$ contains the network evaluation of step $s$ and is $L_s$-Lipschitz on the latents that occur.
An error $e$ added to the latent after step $t$ changes the final latent by at most}
\begin{equation*}
\big\|\hat x_0-x_0\big\|_2\le\Big(\prod_{s=1}^{t-1}L_s\Big)\|e\|_2 .
\end{equation*}
\emph{If the $\Psi_s$ are also differentiable along the dense trajectory, the change is, to first order, $\big(\prod_{s=1}^{t-1}\partial\Psi_s\big)e$, the product of the remaining step Jacobians applied to $e$.}

\emph{Proof.}
Let $\hat x_{t-1}=x_{t-1}+e$.
For every later step, $\|\hat x_{s-1}-x_{s-1}\|=\|\Psi_s(\hat x_s)-\Psi_s(x_s)\|\le L_s\|\hat x_s-x_s\|$, and induction down to $s=1$ gives the bound.
The first-order form follows from the chain rule. $\square$

An attention error $e_u$ at step $t$ enters the latent through the network evaluation inside $\Psi_t$, to first order as $B_ue_u$ with $B_u$ the Jacobian of $\Psi_t$ with respect to the unit's projected output, and then follows the same products, so $G_u=\big(\prod_{s=1}^{t-1}\partial\Psi_s\big)B_u$ and $J_u$ is the squared gain of $G_u$ in the direction of the unit's projected error, averaged over inputs and normalized as in \hyperref[res:A17]{Lemma~A.17}.
Few-step samplers take large steps, and each Jacobian of a large step can stretch some directions strongly, so the products, and with them the sensitivities, differ by orders of magnitude across steps and models.
A per-pixel condition anchors the output to the input, so the step maps contract the directions the condition determines and the products stay small.

\noindent\phantomsection\label{res:A23}\textbf{Corollary A.23 (The backend floor probes the same products).}
\emph{Let two dense attention backends differ at step $s$ by a numerical error $\xi_s$, the difference of their updates of $x_{s-1}$, which given $x_s$ has zero mean and covariance $\Sigma_s$ and is uncorrelated across steps.
Then, to first order and with $P_s=\prod_{s'=1}^{s-1}\partial\Psi_{s'}$ the Jacobian product from $x_{s-1}$ to the output along the dense trajectory,}
\begin{equation*}
\begin{aligned}
\mathbb E\big\|x_0^{(1)}-x_0^{(2)}\big\|_2^2&=\sum_s\operatorname{tr}\big(P_s\Sigma_sP_s^\top\big)\\
&\le\sum_s\Big(\prod_{s'=1}^{s-1}L_{s'}\Big)^2\operatorname{tr}\Sigma_s .
\end{aligned}
\end{equation*}

\emph{Proof.}
By \hyperref[res:A22]{Lemma~A.22}, each $\xi_s$ reaches the output as $P_s\xi_s$ to first order, and $\mathbb E\|P_s\xi_s\|^2=\operatorname{tr}(P_s\Sigma_sP_s^\top)$.
The terms between steps vanish, and $\|P_s\|\le\prod_{s'<s}L_{s'}$ bounds each term. $\square$

The floor weights the Jacobian products by the covariance of the numerical noise, and the sensitivities weight the same products by the directions of the sparse errors, so the floor probes the sensitivities.
The Floor of Table~\ref{tab:main} reads this disagreement on the decoded frames, as a PSNR averaged over clips.
Figure~\ref{fig:obs}d--f and Table~\ref{tab:cv} measure this: the cost of removing the sensitivity falls as the floor rises, with rank correlation $-0.61$ on ten models.

\section{Constants and Implementation}
\label{app:config}

\begin{table}[!htbp]
\centering
\footnotesize
\setlength{\tabcolsep}{5pt}
\begin{tabular}{@{}>{\raggedright\arraybackslash}p{1.535in}>{\raggedright\arraybackslash}p{1.56in}@{}}
\toprule
Quantity & Value\\
\midrule
Clusters per head, k-means rounds & 880, 10 (first sparse step)\\
Density levels $\Theta$ & 0.05, 0.10, 0.15, 0.20, 0.30, 0.45\\
Target density $\rho$ & 0.200 (reported band 0.195--0.205)\\
Kernel block size for density accounting & $128\times96$, dense units count one\\
Always dense & first step, first layer\\
Reassignment threshold $\tau_s$ & 1.0\,dB\\
Speedup threshold $\tau_g$ & 1.10\\
Reference level $\theta_0$ for $J$ & 0.20\\
Block ranking $\mathbf D_u$ & 16 calibration clips\\
Distortion $\varepsilon_u$, sensitivity $J_u$ & 4 calibration clips\\
Block shapes tried (frames$\times$rows$\times$cols) & $2{\times}4{\times}16$, $1{\times}8{\times}16$, $2{\times}8{\times}8$, $4{\times}4{\times}8$, $1{\times}16{\times}8$\\
Seed & 1000\\
\bottomrule
\end{tabular}
\caption{Constants of DecoupAttn, shared by all models.}
\label{tab:config}
\end{table}

\noindent\textbf{Shared constants.}
All ten models use the constants of Table~\ref{tab:config}, and what differs between them is measured on their own clips: the block shape, the block ranking $\mathbf D_u$, the distortion table $\varepsilon_u$, the sensitivity $J_u$, and the assignment rule of Eq.~\eqref{eq:rules}.
The per-model plans come from Distortion Budget's measurements, not tuning, so one procedure serves every family and sampler (Section~\ref{sec:main}).

\noindent\textbf{Block shape.}
Five block shapes are scored by the attention a density-0.20 static mask keeps on eight calibration clips; the default and the best challenger are compared on eight validation clips, a gap of at least 0.1\,dB decides, and otherwise the faster shape is kept.

\noindent\textbf{Sensitivity groups.}
$J$ is measured on groups of sampler steps (one step for four-step models) and one third of the layers, run at level $\theta_0$ with full compensation while the rest stays dense.
Every unit in a group shares its $J_u$, and $J$ is normalized to mean one, so $J_u>1$ marks a unit whose errors reach the final latent more than the average unit's.
Cells that the floor subtraction leaves negative are set to zero before this normalization, and for the models measured by injection a group without a positive excess stops the plan build.
No cell of the deployed plans is affected under the reference backend.
DiffusionRenderer uses one weight per step from an earlier sensitivity map of the same quantity.

\noindent\textbf{Speed check.}
The implementation also measures $S_{\max}$, the dense latency divided by the lowest latency any sparse plan reaches, and runs dense attention when $S_{\max}<\tau_g$.
The experiments in this paper do not use this check.

\noindent\textbf{Kernels.}
The exact blocks run on FlashInfer's variable-block FlashAttention-3 kernel.
Each call computes the cluster means, the per-dimension key variances, and the matrix $H$ (two GEMMs) from the current keys and values, and a fused Triton kernel adds the cluster terms of Eq.~\eqref{eq:decoup} and merges them with the exact part through the log-sum-exp.
Centroid-only calls skip the variance pass and $H$.

\section{Experimental Details}
\label{app:exp}

\begin{table}[!htbp]
\centering
\footnotesize
\setlength{\tabcolsep}{3pt}
\begin{tabular}{lccc}
\toprule
Model & Steps & Video & Test clips\\
\midrule
Cosmos-Predict1 & 35 + CFG & $1280{\times}704$, 121 fr. & 39\\
Cosmos-Predict2.5 & 4 & 720p, 93 fr. & 32\\
Wan2.1-14B & 4 & $832{\times}480$, 81 fr. & 32\\
Wan2.2-A14B & 4 & 480p, 81 fr. & 32\\
HunyuanVideo & 50 & $1280{\times}720$, 129 fr. & 8\\
Cosmos3-Nano (I2W) & 35 + CFG & 720p, 189 fr. & 32\\
Cosmos-Transfer2.5 & 4 & edge control, 93 fr. & 39\\
DiffusionRenderer & 4 & $1280{\times}704$, 57 fr. & 39\\
AlayaRenderer & 50 + CFG & latent $21{\times}30{\times}52$ & 39\\
UniRelight & 35 + CFG & latent $24{\times}30{\times}53$ & 39\\
\bottomrule
\end{tabular}
\caption{Models, samplers, and test splits. Calibration, validation, and test clips are pairwise disjoint.}
\label{tab:models}
\end{table}

\paragraph{Models.}
Table~\ref{tab:models} covers five four-step distilled samplers and five samplers with 35 or 50 steps, four of them with classifier-free guidance, with 8 to 39 test clips per model.
Both sampler types run with the constants of Table~\ref{tab:config}, so what adapts to each model is its calibrated plan.

\paragraph{Density accounting and baseline settings.}
Density counts every computed block after padding to the kernel block size, with dense calls counted as one.
Baselines run from their official code with their density knob set to 0.20; where the official configuration sets a higher per-head minimum, the baseline runs at that minimum (SVG2 and SVG-EAR at 0.225 and 0.224 on Wan2.2-A14B).
On Cosmos-Predict1 and UniRelight, which have no official CalibAtt schedule, its energy schedule follows the official 50-step Wan schedule, scaled to density 0.20 on the calibration set.

\paragraph{Timing.}
Each timing process runs dense attention, the method, and dense attention again, with one warm-up and one timed generation each, measured with CUDA events; latency is the wall-clock time of the denoising loop, and the speedup is the mean dense latency divided by that of the method.
Plan building, clustering, and reassignment are all counted inside the measured latency.

\paragraph{Ablation variants.}
Tied groups compensation by the selection blocks; Centroid drops both corrections; U sets every $J_u$ to one; D replaces $\varepsilon_u$ by the error of the same option without compensation; Frozen keeps the assignments of the first sparse step on the nine models where Eq.~\eqref{eq:rules} reassigns them; Rebuild runs k-means at every sparse step (Table~\ref{tab:rebuild}).
Tied$^\dagger$ applies the same grouping swap under our plan without re-solving, and every other variant is re-solved at density 0.20.
DiffusionRenderer already keeps its assignments under the rule of Eq.~\eqref{eq:rules}, where reassigning them changes PSNR by $+$0.05\,dB.
In Table~\ref{tab:ablfull} the Full column gives the full method's PSNR with its standard deviation across clips.

\paragraph{Ablation results.}
Table~\ref{tab:ablfull} adds the spread and the clip count behind each cell of Table~\ref{tab:abl}.
The decoupling variants lose most consistently: Tied and Tied$^\dagger$ lose on at least six of the eight clips of every model and Centroid on at least seven, since all three enlarge the residual that every query carries (\hyperref[res:prop1]{Proposition~1}).
The cost of U depends more on the model, because it follows how unevenly the sampler amplifies the errors of different units (Table~\ref{tab:cv}): U costs 4.10\,dB on HunyuanVideo but 0.55\,dB on Cosmos-Transfer2.5, whose edges damp the error growth (Figure~\ref{fig:obs}d,e).
Frozen costs at most 1.14\,dB on the video models and most on AlayaRenderer and UniRelight (2.59 and 4.45\,dB).

\paragraph{Speed of the variants.}
Table~\ref{tab:ablspeed} shows what each component costs in time.
U and D change only the prices, so their speedup stays within 0.03 of the full method's, and Distortion Budget gains its 1.18--1.68\,dB at no cost in speed.
Tied and Tied$^\dagger$ group by the selection blocks and skip the clustering, which makes them slightly faster (0.04--0.15 in speedup) but costs 0.85--8.59\,dB (Table~\ref{tab:ablfull}), so the semantic clusters of Section~\ref{sec:decoup} are worth their time.
Frozen saves the per-step assignment, at most 0.08 in speedup, and Rebuild, which runs k-means at every sparse step, falls below dense speed on five of the ten models.

\paragraph{Re-clustering and reassignment.}
Rebuild changes PSNR by at most 0.95\,dB (UniRelight) and by less than 0.4\,dB on the other nine models, while adding 0.8--96.1\,s per video (Table~\ref{tab:rebuild}).
Reassignment already moves every key to its nearest cached centroid (Eq.~\eqref{eq:reassign}), which never increases the spread (\hyperref[res:prop3]{Proposition~3}), so fresh centroids gain little, and the full method wins on five of the eight clips of HunyuanVideo and Cosmos3-Nano.
In the timing session of Table~\ref{tab:labels}, reassignment adds 108--227\,ms (2--3\%) over frozen assignments, whereas Rebuild adds 0.8--1.8\,s (14--21\%) and makes DiffusionRenderer and Wan2.1-14B slower than dense attention, which is why Reassign Cache keeps the cached centroids instead.

\paragraph{VBench.}
Table~\ref{tab:full} scores each video on its own, without the dense output.
DecoupAttn stays within 0.007 of dense attention on every model and is the best or tied-best sparse method on four of them, and on Cosmos3-Nano, where SVOO and SVG-EAR fall to 0.7988 and 0.8100, it keeps the dense score (0.8616 against 0.8612).

\begin{table*}[!tp]
\centering
\footnotesize
\setlength{\tabcolsep}{2pt}
\renewcommand{\arraystretch}{1.05}
\begin{tabular}{l C{1.2cm} *{6}{C{1.8cm}}}
\toprule
 & & \multicolumn{3}{c}{\hd{Decoupling\\(\S\ref{sec:decoup})}} & \multicolumn{2}{c}{\hd{Distortion Budget\\(\S\ref{sec:budget})}} & \hd{Reassign Cache\\(\S\ref{sec:reassign})}\\
\cmidrule(lr){3-5}\cmidrule(lr){6-7}\cmidrule(lr){8-8}
Model (steps) & Full & Tied & Tied$^\dagger$ & Centroid & U & D & Frozen\\
\midrule
\rowcolor{groupgray}\multicolumn{8}{l}{\textit{Video generation}}\\
Cosmos-Predict1 (35) & 23.42\sd{3.91} & $-$1.95\sd{2.14}\,{\scriptsize(6)} & $-$2.45\sd{1.60}\,{\scriptsize(8)} & $-$1.60\sd{1.64}\,{\scriptsize(8)} & $-$1.93\sd{2.97}\,{\scriptsize(7)} & $-$0.34\sd{2.21}\,{\scriptsize(5)} & $-$0.95\sd{0.90}\,{\scriptsize(7)}\\
Cosmos-Predict2.5 (4) & 24.07\sd{3.97} & $-$1.63\sd{0.71}\,{\scriptsize(8)} & $-$1.90\sd{0.84}\,{\scriptsize(8)} & $-$1.03\sd{0.67}\,{\scriptsize(7)} & $-$2.30\sd{0.94}\,{\scriptsize(8)} & $-$0.96\sd{0.39}\,{\scriptsize(8)} & $-$0.38\sd{0.18}\,{\scriptsize(8)}\\
Wan2.1-14B, 480p (4) & 25.31\sd{2.80} & $-$3.72\sd{0.75}\,{\scriptsize(8)} & $-$4.19\sd{1.04}\,{\scriptsize(8)} & $-$1.24\sd{0.50}\,{\scriptsize(8)} & $-$2.14\sd{0.79}\,{\scriptsize(8)} & $-$1.65\sd{0.41}\,{\scriptsize(8)} & $-$1.14\sd{0.31}\,{\scriptsize(8)}\\
Wan2.2-A14B (4) & 17.21\sd{2.87} & $-$1.48\sd{0.53}\,{\scriptsize(8)} & $-$1.66\sd{0.36}\,{\scriptsize(8)} & $-$0.53\sd{0.24}\,{\scriptsize(8)} & $-$1.06\sd{0.56}\,{\scriptsize(8)} & $-$0.65\sd{0.51}\,{\scriptsize(8)} & $-$0.36\sd{0.18}\,{\scriptsize(8)}\\
HunyuanVideo (50) & 25.84\sd{4.57} & $-$1.61\sd{1.53}\,{\scriptsize(6)} & $-$0.85\sd{1.05}\,{\scriptsize(7)} & $-$1.06\sd{1.31}\,{\scriptsize(7)} & $-$4.10\sd{2.43}\,{\scriptsize(8)} & $-$0.87\sd{0.81}\,{\scriptsize(8)} & $-$0.38\sd{0.68}\,{\scriptsize(6)}\\
Cosmos3-Nano I2W (35) & 20.92\sd{2.94} & $-$1.71\sd{2.28}\,{\scriptsize(7)} & $-$2.17\sd{1.73}\,{\scriptsize(8)} & $-$0.60\sd{1.93}\,{\scriptsize(7)} & $-$2.05\sd{1.22}\,{\scriptsize(8)} & $-$1.06\sd{2.25}\,{\scriptsize(7)} & $-$0.12\sd{1.37}\,{\scriptsize(5)}\\
\rowcolor{groupgray}\multicolumn{8}{l}{\textit{Controlled generation}}\\
Cosmos-Transfer2.5 (4) & 33.34\sd{2.01} & $-$2.21\sd{0.78}\,{\scriptsize(8)} & $-$2.67\sd{0.90}\,{\scriptsize(8)} & $-$1.20\sd{0.62}\,{\scriptsize(8)} & $-$0.55\sd{0.51}\,{\scriptsize(6)} & $-$1.31\sd{0.68}\,{\scriptsize(8)} & $-$0.69\sd{0.65}\,{\scriptsize(8)}\\
\rowcolor{groupgray}\multicolumn{8}{l}{\textit{Neural rendering}}\\
DiffusionRenderer (4) & 50.20\sd{1.44} & $-$2.75\sd{0.64}\,{\scriptsize(8)} & $-$5.29\sd{2.58}\,{\scriptsize(8)} & $-$2.48\sd{0.86}\,{\scriptsize(8)} & $-$0.77\sd{0.47}\,{\scriptsize(7)} & $-$1.71\sd{0.91}\,{\scriptsize(8)} & --\\
AlayaRenderer (50) & 27.97\sd{2.81} & $-$4.07\sd{1.51}\,{\scriptsize(8)} & $-$5.43\sd{1.81}\,{\scriptsize(8)} & $-$1.43\sd{0.90}\,{\scriptsize(7)} & $-$1.06\sd{1.37}\,{\scriptsize(7)} & $-$0.86\sd{0.95}\,{\scriptsize(6)} & $-$2.59\sd{1.07}\,{\scriptsize(8)}\\
UniRelight (35) & 41.54\sd{4.39} & $-$7.39\sd{3.89}\,{\scriptsize(8)} & $-$8.59\sd{4.42}\,{\scriptsize(8)} & $-$3.77\sd{1.79}\,{\scriptsize(8)} & $-$0.85\sd{3.89}\,{\scriptsize(3)} & $-$2.41\sd{0.69}\,{\scriptsize(8)} & $-$4.45\sd{2.09}\,{\scriptsize(8)}\\
\bottomrule
\end{tabular}
\caption{\textbf{Component ablations with standard deviations and clip counts} (full version of Table~\ref{tab:abl}).
Each cell is the PSNR of the variant minus that of the full method (dB) on eight validation clips, with the standard deviation of this difference (small $\pm$) and the number of clips on which the full method is better.
Variants are defined in Appendix~\ref{app:exp}, their latencies are in Table~\ref{tab:abl}, and their speedups are in Table~\ref{tab:ablspeed}.}
\label{tab:ablfull}
\end{table*}

\begin{table*}[!tp]
\centering
\footnotesize
\setlength{\tabcolsep}{2pt}
\begin{tabular}{l*{9}{C{1.36cm}}}
\toprule
Model (steps) & \textcolor{floorgray}{Floor} & CalibAtt & SVOO & SVG2 & SVG-EAR & Sol-Attn & PISA & SparsePR & \textbf{Ours}\\
\midrule
\rowcolor{groupgray}\multicolumn{10}{l}{\textit{Video generation}}\\
Cosmos-Predict1 (35) & \textcolor{floorgray}{0.8375} & 0.8177 & 0.8273 & 0.8331 & 0.8360 & 0.8342 & 0.8339 & \textbf{0.8364} & 0.8313\\
Cosmos-Predict2.5 (4) & \textcolor{floorgray}{0.8623} & 0.8407 & 0.8633 & 0.8600 & 0.8621 & \textbf{0.8646} & 0.8597 & 0.8624 & \textbf{0.8646}\\
Wan2.1-14B, 480p (4) & \textcolor{floorgray}{0.8668} & 0.8589 & 0.8664 & 0.8655 & 0.8675 & \textbf{0.8689} & 0.8677 & 0.8665 & 0.8673\\
Wan2.2-A14B (4) & \textcolor{floorgray}{0.8636} & 0.8638 & 0.8642 & 0.8622 & 0.8634 & \textbf{0.8669} & 0.8662 & 0.8628 & 0.8639\\
HunyuanVideo (50) & \textcolor{floorgray}{0.8500} & 0.8470 & 0.8469 & 0.8445 & 0.8499 & \textbf{0.8548} & 0.8544 & 0.8517 & 0.8468\\
Cosmos3-Nano I2W (35) & \textcolor{floorgray}{0.8612} & 0.8457 & 0.7988 & 0.8273 & 0.8100 & 0.8456 & 0.8585 & 0.8499 & \textbf{0.8616}\\
\rowcolor{groupgray}\multicolumn{10}{l}{\textit{Controlled generation}}\\
Cosmos-Transfer2.5 (4) & \textcolor{floorgray}{0.8320} & 0.8107 & 0.8173 & 0.8208 & 0.8292 & 0.8264 & 0.8179 & 0.8310 & \textbf{0.8335}\\
\rowcolor{groupgray}\multicolumn{10}{l}{\textit{Neural rendering}}\\
DiffusionRenderer (4) & \textcolor{floorgray}{0.8342} & 0.8315 & 0.8279 & 0.8306 & 0.8341 & \textbf{0.8347} & 0.8334 & 0.8344 & 0.8343\\
AlayaRenderer (50) & \textcolor{floorgray}{0.8269} & 0.8082 & 0.8136 & 0.8161 & 0.8244 & 0.8213 & 0.8232 & 0.8172 & \textbf{0.8268}\\
UniRelight (35) & \textcolor{floorgray}{0.8330} & 0.8297 & 0.8284 & 0.8309 & 0.8331 & \textbf{0.8336} & 0.8330 & 0.8325 & 0.8333\\
\bottomrule
\end{tabular}
\caption{\textbf{VBench} for the models of Table~\ref{tab:main}: mean of subject consistency, background consistency, temporal flickering, motion smoothness, imaging quality, and aesthetic quality. SSIM and LPIPS with their standard deviations are in Table~\ref{tab:main}. Bold: best sparse method.}
\label{tab:full}
\end{table*}

\begin{table*}[!tp]
\centering
\footnotesize
\setlength{\tabcolsep}{3.4pt}
\begin{tabular}{l*{8}{C{1.0cm}}}
\toprule
Model (steps) & Full & Tied & Tied$^\dagger$ & Centroid & U & D & Frozen & Rebuild\\
\midrule
\rowcolor{groupgray}\multicolumn{9}{l}{\textit{Video generation}}\\
Cosmos-Predict1 (35) & 1.47 & 1.56 & 1.55 & 1.49 & 1.45 & 1.48 & 1.53 & 1.18\\
Cosmos-Predict2.5 (4) & 1.48 & 1.63 & 1.62 & 1.50 & 1.47 & 1.48 & 1.51 & 1.33\\
Wan2.1-14B, 480p (4) & 1.06 & 1.17 & 1.16 & 1.08 & 1.06 & 1.06 & 1.08 & 0.95\\
Wan2.2-A14B (4) & 1.07 & 1.17 & 1.17 & 1.08 & 1.07 & 1.07 & 1.09 & 0.95\\
HunyuanVideo (50) & 1.65 & 1.71 & 1.69 & 1.71 & 1.62 & 1.62 & 1.67 & 1.42\\
Cosmos3-Nano I2W (35) & 1.21 & 1.28 & 1.28 & 1.25 & 1.20 & 1.21 & 1.26 & 0.97\\
\rowcolor{groupgray}\multicolumn{9}{l}{\textit{Controlled generation}}\\
Cosmos-Transfer2.5 (4) & 1.47 & 1.60 & 1.60 & 1.48 & 1.46 & 1.47 & 1.49 & 1.32\\
\rowcolor{groupgray}\multicolumn{9}{l}{\textit{Neural rendering}}\\
DiffusionRenderer (4) & 1.11 & 1.23 & 1.24 & 1.07 & 1.11 & 1.11 & -- & 0.92\\
AlayaRenderer (50) & 1.24 & 1.35 & 1.34 & 1.27 & 1.24 & 1.24 & 1.32 & 0.87\\
UniRelight (35) & 1.30 & 1.39 & 1.38 & 1.33 & 1.29 & 1.30 & 1.36 & 1.01\\
\bottomrule
\end{tabular}
\caption{\textbf{Speedup over dense attention of every ablation variant} (validation clips, same process as its dense anchor). DiffusionRenderer already freezes its cluster assignments.}
\label{tab:ablspeed}
\end{table*}

\begin{table}[!htbp]
\centering
\footnotesize
\setlength{\tabcolsep}{3pt}
\begin{tabular}{lccc}
\toprule
Model (steps) & $\Delta$PSNR (dB) & Clips & $\Delta$ms\\
\midrule
Cosmos-Predict1 (35) & +0.35\sd{0.30} & 2 & \tms{+42{,}257}\\
Cosmos-Predict2.5 (4) & +0.07\sd{0.05} & 1 & \tms{+944}\\
Wan2.1-14B, 480p (4) & +0.20\sd{0.10} & 0 & \tms{+1{,}543}\\
Wan2.2-A14B (4) & +0.07\sd{0.09} & 1 & \tms{+1{,}526}\\
HunyuanVideo (50) & +0.31\sd{0.86} & 5 & \tms{+96{,}081}\\
Cosmos3-Nano I2W (35) & $-$0.36\sd{0.61} & 5 & \tms{+47{,}008}\\
Cosmos-Transfer2.5 (4) & 0.00\sd{0.24} & 2 & \tms{+1{,}089}\\
DiffusionRenderer (4) & +0.09\sd{0.21} & 3 & \tms{+825}\\
AlayaRenderer (50) & +0.03\sd{1.05} & 4 & \tms{+19{,}443}\\
UniRelight (35) & +0.95\sd{0.89} & 2 & \tms{+31{,}747}\\
\bottomrule
\end{tabular}
\caption{\textbf{Re-clustering at every step} (Rebuild) on the validation clips of Table~\ref{tab:abl}.
$\Delta$ is Rebuild minus the full method with its standard deviation over the clips (small $\pm$), and Clips counts the clips (of 8) on which the full method is better.}
\label{tab:rebuild}
\end{table}

\begin{table}[!htbp]
\centering
\footnotesize
\setlength{\tabcolsep}{3pt}
\begin{tabular}{lcccc}
\toprule
Model & Frozen & Reassign & Rebuild & Rule\\
\midrule
DiffusionRenderer & 3,904.0 & 4,012.2 & 4,729.7 & frozen\\
Cosmos-Predict2.5 & 8,053.8 & 8,219.5 & 9,165.9 & reassign\\
Wan2.1-14B & 12,710.6 & 12,937.7 & 14,508.7 & reassign\\
\bottomrule
\end{tabular}
\caption{Latency (ms) of the three assignment schemes in one timing session; dense is 4,329, 12,170, and 13,747\,ms.}
\label{tab:labels}
\end{table}

\paragraph{Coherence measurement.}
Table~\ref{tab:kappa} takes the (clip, step, layer, head) units of the deployed plans on four calibration clips with seed 1000 (conditional branch, sparse steps, layers after the first): 10,040 units on DiffusionRenderer, 17,968 on Wan2.1-14B, and 4,948 on Cosmos-Predict2.5.
Each unit pools the sums over its queries, $\kappa=\sum_i\|\sum_b r_{ib}\|_2^2/\sum_i\sum_b\|r_{ib}\|_2^2$, in value space with dense probabilities ($Z_i=1$), so high-energy queries weigh most.
The normalized ratio $\zeta=(\kappa-1)/(K_g-1)$ divides out the number $K_g$ of compensated key blocks $b\notin\mathcal S_g$ per query (medians 198, 319, and 594) and equals the mean pairwise cosine of the residuals when their norms are equal.
Compensation lowers $\kappa$ on all three models, and the corrections lower it further: the median falls 2.9--3.9-fold, from 4.69--10.37 without compensation to 2.67--5.12 with plain centroids and 1.61--2.69 with DecoupAttn, and the median $\zeta$ falls to 0.002--0.003, so the residuals of a few hundred compensated blocks are nearly orthogonal.
This is the cancellation of \hyperref[res:prop1]{Proposition~1}: a compensated residual comes from member deviations that sum to zero over each cluster, so the skipped blocks share almost no common direction, and blocks fixed offline suffice (Section~\ref{sec:exp-dither}).

\paragraph{Sensitivity spread.}
Table~\ref{tab:cv} relates the cost of U to how unevenly $J$ spreads over the cells a plan uses.
On the four-step video models the cost tracks the spread (coefficient of variation 1.11--2.23, cost 1.06--2.30\,dB).
A flat $J$ misprices units when their true sensitivities differ.
Cosmos-Transfer2.5 and UniRelight spread $J$ widely yet lose less than 0.9\,dB without it, and both have high dense-backend floors (41.93 and 55.49\,dB): their conditions damp the error growth along the sampler, which leaves less for the sensitivity to steer (Figure~\ref{fig:obs}d--f).

\begin{table}[!htbp]
\centering
\footnotesize
\setlength{\tabcolsep}{3pt}
\begin{tabular}{llcc}
\toprule
Model & Compensation & $\kappa$ (25/50/75\%) & median $\zeta$\\
\midrule
DiffusionRenderer & none & 2.83 / 4.69 / 7.71 & 0.021\\
 & centroid only & 1.87 / 2.69 / 4.24 & 0.010\\
 & ours & 1.26 / 1.61 / 2.12 & 0.003\\
\midrule
Wan2.1-14B & none & 2.15 / 5.03 / 8.96 & 0.014\\
 & centroid only & 1.86 / 2.67 / 3.66 & 0.006\\
 & ours & 1.20 / 1.65 / 2.18 & 0.002\\
\midrule
Cosmos-Predict2.5 & none & 6.29 / 10.37 / 16.96 & 0.017\\
 & centroid only & 3.22 / 5.12 / 7.36 & 0.008\\
 & ours & 2.02 / 2.69 / 3.42 & 0.003\\
\bottomrule
\end{tabular}
\caption{\textbf{Coherence ratio} $\kappa$ of Eq.~\eqref{eq:coherence} as quartiles over the units of the deployed plans, with the median of the normalized ratio $\zeta$.}
\label{tab:kappa}
\end{table}

\begin{table}[!htbp]
\centering
\footnotesize
\setlength{\tabcolsep}{3pt}
\begin{tabular}{lccc}
\toprule
Model & CV of $J$ & Cost of U (dB) & Floor (dB)\\
\midrule
Cosmos-Predict2.5 & 2.23 & 2.30 & 27.78\\
Wan2.1-14B & 1.99 & 2.14 & 24.93\\
Wan2.2-A14B & 1.11 & 1.06 & 21.10\\
Cosmos-Transfer2.5 & 1.80 & 0.55 & 41.93\\
DiffusionRenderer & 0.62 & 0.77 & 54.70\\
AlayaRenderer & 1.02 & 1.06 & 39.13\\
UniRelight & 1.76 & 0.85 & 55.49\\
\bottomrule
\end{tabular}
\caption{Spread of the sensitivity (coefficient of variation over the cells the plan uses), the cost of removing it (Table~\ref{tab:abl}), and the dense-backend Floor column (Table~\ref{tab:main}).}
\label{tab:cv}
\end{table}

\section{Supplementary Analyses}
\label{app:supp}

This appendix isolates each mechanism of Section~\ref{sec:method} and tests whether its gain carries over to matched latency, unseen inputs, and task metrics.
Every comparison shares prompts, seeds, conditioning inputs, and the dense reference, and runs on clips that took no part in calibration or validation, after every variant is fixed; Table~\ref{tab:suppmodels} lists the models that each analysis uses.
Cosmos-Predict2.5 is in every analysis, and the other models add another video family (Wan2.1-14B), many- and four-step rendering (UniRelight, DiffusionRenderer), and edge control (Cosmos-Transfer2.5), and the task metrics cover nine of the ten models, all but UniRelight.

\noindent\textbf{Statistics.}
Mechanism diagnostics record the queries, keys, and values of selected (step, layer, head) units on eight held-out clips without storing attention maps, plan screening uses eight held-out clips, and the final comparisons use 8 to 39 prompts or conditions per model with one seed each.
We report paired differences with 95\% bootstrap confidence intervals, resampling prompts, together with the win rate and the worst-decile difference; frames, heads, and layers are never counted as independent samples.

\begin{table}[!htbp]
\centering
\footnotesize
\setlength{\tabcolsep}{3pt}
\begin{tabular}{ll}
\toprule
Model & Analyses (Appendix)\\
\midrule
All ten models & \ref{app:supp-density}, \ref{app:supp-cache} (rule)\\
Cosmos-Predict2.5 & \ref{app:supp-decoup}--\ref{app:supp-task}\\
Wan2.1-14B & \ref{app:supp-decoup}, \ref{app:supp-corr}, \ref{app:supp-resid}, \ref{app:supp-eff}--\ref{app:supp-task}\\
UniRelight & \ref{app:supp-cache}, \ref{app:supp-eff}\\
DiffusionRenderer & \ref{app:supp-resid}, \ref{app:supp-cache}, \ref{app:supp-task}\\
Cosmos-Transfer2.5 & \ref{app:supp-budget}, \ref{app:supp-task}\\
Other five models & \ref{app:supp-task}\\
\bottomrule
\end{tabular}
\caption{\textbf{Models of each supplementary analysis}, by the subsections of Appendix~\ref{app:supp} that use them.}
\label{tab:suppmodels}
\end{table}

\subsection{Density Accounting}
\label{app:supp-density}
Table~\ref{tab:density} separates the always-dense first step and first layer from the units the plan sparsifies: $f_{\text{dense}}$ is their share of the attention work, $X_{\text{sp}}$ is the density of the remaining units, and the whole run computes
\begin{equation*}
\rho_{\text{global}}=f_{\text{dense}}+(1-f_{\text{dense}})\,X_{\text{sp}}.
\end{equation*}
Both guidance branches of the CFG models, the first step, the first layer, and every diagonal block count as computed, and every block is padded to the kernel block size.
All ten plans spend density 0.200 on the units they sparsify, and the global density differs only through the dense share: the first step and the first layer take 27--28\% of the attention work on the four-step models but 3.6--6.3\% on the many-step models, giving global densities of 0.415--0.421 and 0.229--0.251.
All methods keep these parts dense (Section~\ref{sec:setup}), so this share bounds every method's speedup.

\begin{table*}[!tp]
\centering
\footnotesize
\setlength{\tabcolsep}{3pt}
\begin{tabular}{l c *{4}{C{1.9cm}}}
\toprule
Model & Steps & $f_{\text{dense}}$ & $X_{\text{sp}}$ & $\rho_{\text{global}}$ & Computed blocks (M)\\
\midrule
Cosmos-Predict1 & 35 + CFG & 0.0633 & 0.200 & 0.251 & 4{,}057\\
Cosmos-Predict2.5 & 4 & 0.2768 & 0.200 & 0.421 & 439\\
Wan2.1-14B & 4 & 0.2687 & 0.200 & 0.415 & 232\\
Wan2.2-A14B & 4 & 0.2687 & 0.200 & 0.415 & 232\\
HunyuanVideo & 50 & 0.0363 & 0.200 & 0.229 & 18{,}943\\
Cosmos3-Nano (I2W) & 35 + CFG & 0.0556 & 0.200 & 0.244 & 3{,}128\\
Cosmos-Transfer2.5 & 4 & 0.2734 & 0.200 & 0.419 & 498\\
DiffusionRenderer & 4 & 0.2768 & 0.200 & 0.421 & 97\\
AlayaRenderer & 50 + CFG & 0.0527 & 0.200 & 0.242 & 761\\
UniRelight & 35 + CFG & 0.0633 & 0.200 & 0.251 & 1{,}863\\
\bottomrule
\end{tabular}
\caption{\textbf{Density accounting of the deployed plans.} Shares of attention work are weighted by sampler steps; computed blocks count $128{\times}96$ kernel blocks per video, in millions.}
\label{tab:density}
\end{table*}

\subsection{Decoupling}
\label{app:supp-decoup}
\begin{figure}[!htbp]
\centering
\includegraphics[width=\linewidth]{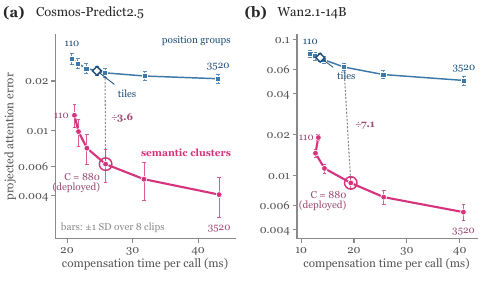}
\caption{\textbf{Group count.} Projected attention error against compensation time as the number of position or semantic groups per head varies, with both corrections on (Cosmos-Predict2.5, Wan2.1-14B). Time: CUDA time per attention call of the diagnostic implementation, exact part and compensation, without the assignment refresh. Bars: $\pm1$ standard deviation over eight clips.}
\label{fig:groups}
\end{figure}

\noindent\textbf{Grouping under a fixed mask.}
Table~\ref{tab:grouping} holds the exact mask, the per-unit exact budget, the block size, the assignment scheme, and the precision fixed, and changes only how the skipped keys are summarized: by nothing, by position (the selection blocks), or by semantic clusters, each with plain centroids or with both corrections of Eq.~\eqref{eq:comp}.
Figure~\ref{fig:groups} varies the number of groups of both schemes and plots the projected attention error against the compensation time.
With the mask fixed and plain centroids, semantic clusters raise PSNR over position groups by 5.11\,dB on Cosmos-Predict2.5 and 3.72\,dB on Wan2.1-14B (Table~\ref{tab:grouping}), and with both corrections they lower the projected error at every group count in the diagnostic implementation of Figure~\ref{fig:groups}, 3.6-fold and 7.1-fold at the deployed 880 clusters.
The corrections help both groupings, yet position groups with both corrections (22.33 and 20.66\,dB) remain below semantic clusters without them (23.47 and 23.41\,dB): the corrections repair the mass and the first-order value of each group but cannot pull far-apart members together, and the spread of the members is what sets the residual (\hyperref[res:A6]{Proposition~A.6}).
The semantic clusters cost 0.7--1.4\,s more DiT time than position groups, which buys 2.37 and 4.04\,dB with both corrections on, so decoupling the grouping from the selection pays on both models.

\begin{table*}[!tp]
\centering
\footnotesize
\setlength{\tabcolsep}{3pt}
\begin{tabular}{l l l *{10}{C{1.14cm}}}
\toprule
 & & & \multicolumn{5}{c}{Cosmos-Predict2.5} & \multicolumn{5}{c}{Wan2.1-14B}\\
\cmidrule(lr){4-8}\cmidrule(lr){9-13}
Variant & Grouping & Corrections & Attn.\ MSE & Latent MSE & PSNR & LPIPS & DiT ms & Attn.\ MSE & Latent MSE & PSNR & LPIPS & DiT ms\\
\midrule
A & -- & none & 0.415 & 0.214 & 17.57\sd{2.33} & 0.385 & 7{,}176 & 0.577 & 0.186 & 17.08\sd{1.10} & 0.300 & 11{,}291\\
B & Position & Centroid & 0.0386 & 0.174 & 18.36\sd{2.72} & 0.297 & 7{,}390 & 0.109 & 0.0886 & 19.69\sd{1.46} & 0.155 & 11{,}444\\
C & Semantic & Centroid & 0.0108 & 0.0541 & 23.47\sd{3.68} & 0.124 & 8{,}127 & 0.0147 & 0.0364 & 23.41\sd{2.05} & 0.078 & 12{,}756\\
D & Position & Both & 0.0231 & 0.069 & 22.33\sd{3.26} & 0.152 & 7{,}471 & 0.0732 & 0.0685 & 20.66\sd{1.62} & 0.123 & 11{,}552\\
E (ours) & Semantic & Both & 0.00623 & 0.0412 & 24.70\sd{3.78} & 0.102 & 8{,}212 & 0.00880 & 0.0260 & 24.70\sd{2.29} & 0.057 & 12{,}902\\
\bottomrule
\end{tabular}
\caption{\textbf{Compensation grouping under a fixed exact mask.}
Only the summary of the skipped keys changes, A computes the exact blocks alone, and Both adds the variance and value corrections of Eq.~\eqref{eq:comp}.
Small $\pm$: standard deviation of PSNR over the test prompts.}
\label{tab:grouping}
\end{table*}

\noindent\textbf{Selection under a fixed compensator.}
Table~\ref{tab:selection} keeps the semantic compensator, the cluster assignments, the per-unit exact budget, and the kernels fixed and changes only how the exact blocks are chosen.
Online residual selection ranks the blocks of each call by their residual under that call's dense attention, information that no deployed method reads.
The last row attaches the same compensator to the masks of CalibAtt~\cite{yehezkel2026calibatt}.
Under one compensator, ranking blocks by their residual beats ranking them by attention mass by 0.32 and 0.33\,dB at the same cost (Table~\ref{tab:selection}), because the plan keeps exact the blocks that compensation handles worst, while a block with high attention mass may be one the centroids already predict well.
Routing by mass at every call gains 0.27\,dB over offline mass ranking on Cosmos-Predict2.5 for 650\,ms of routing, loses 0.42\,dB on Wan2.1-14B for 1{,}288\,ms, and stays below our offline blocks on both, since with every key compensated the residuals cancel, and the remaining error depends on the blocks that compensation handles worst, which calibration measures once (\hyperref[res:prop1]{Proposition~1}).
The blocks themselves matter: random blocks lose 7.44 and 8.34\,dB, and the masks of CalibAtt reach 21.22 and 20.79\,dB under our compensator, 3.48 and 3.91\,dB below our blocks.
The diagnostic online residual gains 2.33--3.27\,dB at 86--143 times the latency, while our fixed blocks add no routing time.

\begin{table*}[!tp]
\centering
\footnotesize
\setlength{\tabcolsep}{3pt}
\begin{tabular}{l l *{8}{C{1.2cm}}}
\toprule
 & & \multicolumn{4}{c}{Cosmos-Predict2.5} & \multicolumn{4}{c}{Wan2.1-14B}\\
\cmidrule(lr){3-6}\cmidrule(lr){7-10}
Selection & Ranking & PSNR & LPIPS & Routing ms & DiT ms & PSNR & LPIPS & Routing ms & DiT ms\\
\midrule
Offline & Attention mass & 24.38\sd{3.66} & 0.107 & 0 & 8{,}220 & 24.37\sd{2.24} & 0.061 & 0 & 12{,}905\\
Offline & Residual (ours) & 24.70\sd{3.78} & 0.102 & 0 & 8{,}212 & 24.70\sd{2.29} & 0.057 & 0 & 12{,}902\\
Online & Attention mass & 24.65\sd{3.46} & 0.098 & 650 & 8{,}862 & 23.95\sd{2.08} & 0.066 & 1{,}288 & 14{,}192\\
Online & Residual$^\ast$ & 27.97\sd{3.96} & 0.058 & 1{,}162{,}863 & 1{,}171{,}074 & 27.03\sd{2.48} & 0.036 & 1{,}092{,}072 & 1{,}104{,}968\\
Offline & Random & 17.26\sd{2.61} & 0.415 & 0 & 8{,}219 & 16.36\sd{1.42} & 0.289 & 0 & 12{,}868\\
\multicolumn{2}{l}{CalibAtt masks~\cite{yehezkel2026calibatt}} & 21.22\sd{3.39} & 0.183 & 0 & 8{,}206 & 20.79\sd{1.73} & 0.116 & 0 & 12{,}875\\
\bottomrule
\end{tabular}
\caption{\textbf{Block selection under a fixed semantic compensator.}
The starred row reads the dense attention of the current call (diagnostic), and Random keeps every diagonal block.
Small $\pm$: standard deviation of PSNR over the test prompts of each model.}
\label{tab:selection}
\end{table*}

\subsection{Compensation Corrections}
\label{app:supp-corr}
\noindent\textbf{Factorial ablation.}
Table~\ref{tab:corr} switches the variance correction $\gamma_{ic}$ and the first-order value correction $H$ of Eq.~\eqref{eq:comp} on and off.
Under a fixed plan, the exact mask, the per-unit options, the assignments, and the set of \textsc{Full} calls are held, so the solver cannot route around a removed term; under a re-solved plan, every variant is re-solved at the same density.
The mass error $|\hat Z_i/Z_i-1|$ isolates what a centroid misses in the denominator, and the projected error of Eq.~\eqref{eq:distortion} measures the resulting output error.
Under a fixed plan, each correction alone moves PSNR by $-$0.16 to $+$0.28\,dB, but together they add 1.23 and 1.29\,dB (Table~\ref{tab:corr}).
The variance correction restores the mass that a centroid misses and cuts the median mass error about threefold (0.0512 to 0.0162 and 0.0641 to 0.0228), and the value correction fixes the numerator; alone, each changes the latent error little (0.0541 to 0.0560 and 0.0546 on Cosmos-Predict2.5, 0.0364 to 0.0362 and 0.0325 on Wan2.1-14B), while together they lower it to 0.0412 and 0.0260.
Under re-solved plans each correction alone already recovers 0.76--1.16\,dB, because the solver moves exact blocks to the units that the missing term hurts most, so the fixed-plan rows measure the terms themselves, and both terms of Eq.~\eqref{eq:comp} are needed.

\begin{table*}[!tp]
\centering
\footnotesize
\setlength{\tabcolsep}{3pt}
\begin{tabular}{l c c *{7}{C{1.25cm}}}
\toprule
Plan & $\gamma_{ic}$ & $H$ & $|\hat Z/Z-1|$ (50\%) & $|\hat Z/Z-1|$ (99\%) & Attn.\ MSE & Latent MSE & PSNR & LPIPS & DiT ms\\
\midrule
\rowcolor{groupgray}\multicolumn{10}{l}{\textit{Cosmos-Predict2.5}}\\
Fixed & off & off & 0.0512 & 0.479 & 0.0108 & 0.0541 & 23.47\sd{3.68} & 0.124 & 8{,}127\\
 & on & off & 0.0162 & 0.33 & 0.00832 & 0.056 & 23.44\sd{3.71} & 0.129 & 8{,}206\\
 & off & on & 0.0512 & 0.479 & 0.00916 & 0.0546 & 23.31\sd{3.52} & 0.128 & 8{,}171\\
 & on & on & 0.0162 & 0.33 & 0.00623 & 0.0412 & 24.70\sd{3.78} & 0.102 & 8{,}212\\
Re-solved & off & off & -- & -- & -- & 0.0539 & 23.51\sd{3.67} & 0.124 & 8{,}138\\
 & on & off & -- & -- & -- & 0.0439 & 24.44\sd{3.77} & 0.106 & 8{,}222\\
 & off & on & -- & -- & -- & 0.0423 & 24.55\sd{3.65} & 0.105 & 8{,}225\\
 & on & on & 0.0162 & 0.33 & 0.00623 & 0.0412 & 24.70\sd{3.78} & 0.102 & 8{,}212\\
\rowcolor{groupgray}\multicolumn{10}{l}{\textit{Wan2.1-14B}}\\
Fixed & off & off & 0.0641 & 0.433 & 0.0147 & 0.0364 & 23.41\sd{2.05} & 0.078 & 12{,}756\\
 & on & off & 0.0228 & 0.311 & 0.0123 & 0.0362 & 23.47\sd{2.09} & 0.079 & 12{,}836\\
 & off & on & 0.0641 & 0.433 & 0.0117 & 0.0325 & 23.69\sd{2.13} & 0.069 & 12{,}808\\
 & on & on & 0.0228 & 0.311 & 0.00880 & 0.0260 & 24.70\sd{2.29} & 0.057 & 12{,}902\\
Re-solved & off & off & -- & -- & -- & 0.0365 & 23.41\sd{2.03} & 0.078 & 12{,}733\\
 & on & off & -- & -- & -- & 0.0293 & 24.17\sd{2.24} & 0.063 & 12{,}907\\
 & off & on & -- & -- & -- & 0.0262 & 24.57\sd{2.15} & 0.059 & 12{,}887\\
 & on & on & 0.0228 & 0.311 & 0.00880 & 0.0260 & 24.70\sd{2.29} & 0.057 & 12{,}902\\
\bottomrule
\end{tabular}
\caption{\textbf{Variance and value corrections, $2{\times}2$.}
Fixed holds the exact mask, options, assignments, and \textsc{Full} calls, re-solved re-solves each variant at equal density, and mass errors are query medians and 99th percentiles.
Small $\pm$: standard deviation over test prompts.}
\label{tab:corr}
\end{table*}

\begin{figure}[!htbp]
\centering
\includegraphics[width=\linewidth]{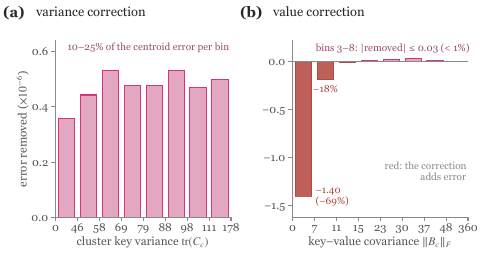}
\caption{\textbf{Where the corrections act} (Cosmos-Predict2.5). Per-cluster residual energy removed by the variance correction, binned by cluster key variance (a), and by the value correction, binned by key--value covariance (b). Each bin holds one eighth of the clusters, and a negative bar means the correction adds error.}
\label{fig:corrbins}
\end{figure}

\noindent\textbf{Where each correction acts.}
Figure~\ref{fig:corrbins} bins clusters by key variance and by key--value covariance and plots the residual energy each correction removes from the clusters of each bin.
The variance correction removes 10--25\% of the centroid error in every bin.
The value correction applies the head average $H$ to every cluster, so on the clusters whose own $B_c$ is smallest the gap $B_c-H$ of \hyperref[res:A6]{Proposition~A.6} exceeds $B_c$ and the correction adds error, and elsewhere it changes the per-cluster error by less than 1\%.
Over the whole output it lowers the projected attention error from 0.0108 to 0.00916 under a fixed plan (Table~\ref{tab:corr}), and a per-cluster $H_c$ closes the gap (Table~\ref{tab:corrdiag}).
Table~\ref{tab:corrdiag} replaces the diagonal variance by the full covariance and the shared $H$ by one $H_c$ per cluster on saved attention inputs.
Full covariance lowers the attention error by only 5\% (0.0059 to 0.00559), so the diagonal variance already captures nearly all the mass a centroid misses, whereas one $H_c$ per cluster lowers it by 27\% (to 0.00428), in line with Figure~\ref{fig:corrbins}b.
The deployed compensator keeps the diagonal variance and the shared $H$ of Eq.~\eqref{eq:comp} at two GEMMs per call, and $H_c$ extends the same correction.

\begin{table}[!htbp]
\centering
\footnotesize
\setlength{\tabcolsep}{3pt}
\begin{tabular}{llcc}
\toprule
Variance & Value & Attn.\ MSE & Extra ms\\
\midrule
Diagonal & Shared $H$ (ours) & 0.0059 & 0\\
Full covariance & Shared $H$ & 0.00559 & --\\
Diagonal & Per-cluster $H_c$ & 0.00428 & --\\
\bottomrule
\end{tabular}
\caption{\textbf{Richer corrections on saved attention inputs} (Cosmos-Predict2.5). Extra ms: added time per call (--: offline implementation).}
\label{tab:corrdiag}
\end{table}

\subsection{Distortion Budget}
\label{app:supp-budget}
\noindent\textbf{Pricing and sensitivity.}
Table~\ref{tab:budget} crosses the error that prices a unit, measured before or after compensation, with the sensitivity, set to one or measured, at one compensator, block candidate set, assignment scheme, and density.
Shuffling $J$ across units while keeping its values tests whether the gain comes from locating sensitive units, and pricing the head output without the $W_o$ projection of Eq.~\eqref{eq:distortion} tests the projection.
The sensitivity also pays at the density 0.10 of Figure~\ref{fig:tradeoff}b--d, where setting it to one costs 1.54\,dB on Cosmos-Predict2.5.
Two time-only rows compare the curvature schedule of PASA~\cite{zhang2026pasa}, ported into DecoupAttn, with Distortion Budget restricted to one weight per step, and the density of every row is re-measured after padding.
Both factors of the price $J_u\varepsilon_u$ matter, and they pay most together (Table~\ref{tab:budget}): on Cosmos-Predict2.5, pricing the compensated error gains 0.41\,dB over the uncompensated error with $J=1$, the measured $J$ gains 1.94\,dB, and both together gain 2.99\,dB (24.70 against 21.71\,dB), with the same ordering on Cosmos-Transfer2.5.
Shuffling $J$ keeps its values but not their places and falls to 20.05\,dB, below $J=1$, so the gain comes from locating the units whose errors reach the final latent.
Pricing the head output without $W_o$ changes PSNR by at most 0.14\,dB, and the two time-only schedules reach 21.79 and 22.01\,dB, near $J=1$, so most of the gain comes from pricing layers and heads, not steps alone.

\begin{table*}[!tp]
\centering
\footnotesize
\setlength{\tabcolsep}{3pt}
\begin{tabular}{l l *{8}{C{1.15cm}}}
\toprule
 & & \multicolumn{4}{c}{Cosmos-Predict2.5} & \multicolumn{4}{c}{Cosmos-Transfer2.5}\\
\cmidrule(lr){3-6}\cmidrule(lr){7-10}
Error priced & Sensitivity & Latent MSE & PSNR & LPIPS & Density & Latent MSE & PSNR & LPIPS & Density\\
\midrule
Before compensation & $J=1$ & 0.0813 & 21.71 & 0.163 & 0.200 & 0.0180 & 31.62 & 0.074 & 0.200\\
After compensation & $J=1$ (U) & 0.0743 & 22.12 & 0.151 & 0.200 & 0.0146 & 32.70 & 0.060 & 0.200\\
Before compensation & Measured $J$ (D) & 0.052 & 23.65 & 0.122 & 0.200 & 0.0124 & 33.16 & 0.059 & 0.200\\
After compensation & Measured $J$ (ours) & 0.0412 & 24.70 & 0.102 & 0.200 & 0.0111 & 33.90 & 0.050 & 0.200\\
After compensation & Shuffled $J$ & 0.112 & 20.05 & 0.206 & 0.200 & 0.0180 & 31.68 & 0.072 & 0.200\\
Head output (no $W_o$) & Measured $J$ & 0.041 & 24.77 & 0.100 & 0.200 & 0.0114 & 33.76 & 0.051 & 0.200\\
\multicolumn{2}{l}{Time only, PASA schedule~\cite{zhang2026pasa}} & 0.0784 & 21.79 & 0.160 & 0.200 & 0.0197 & 30.95 & 0.082 & 0.200\\
\multicolumn{2}{l}{Time only, Distortion Budget} & 0.0744 & 22.01 & 0.154 & 0.200 & 0.0179 & 31.35 & 0.079 & 0.200\\
\bottomrule
\end{tabular}
\caption{\textbf{Pricing error and sensitivity.}
U and D are the ablations of Table~\ref{tab:abl} under this protocol, and density is measured after padding.
Cosmos-Transfer2.5 comes from one rerun with saved latents, in which the full method reaches 33.90\,dB (33.81\,dB in Table~\ref{tab:main}).}
\label{tab:budget}
\end{table*}

\begin{figure}[!htbp]
\centering
\includegraphics[width=\linewidth]{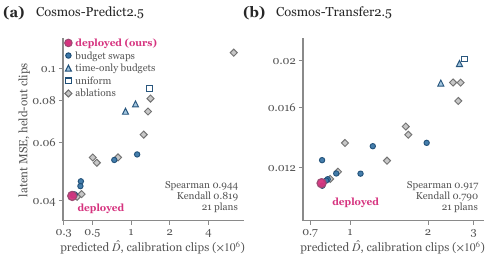}
\caption{\textbf{Predicted versus actual distortion} of candidate plans at one density; predicted on calibration clips, measured on held-out clips (Cosmos-Predict2.5, Cosmos-Transfer2.5).}
\label{fig:proxy}
\end{figure}

\noindent\textbf{Predicted versus actual distortion.}
For each model, 21 candidate plans at one density (the deployed plan, uniform allocation, compensation error alone, a time-only schedule, small budget swaps of the deployed plan, and the ablation plans) are priced on the calibration set by $\hat D(\pi)=\sum_u w_uJ_u\varepsilon_u(o_u,\eta_{tl})$, and their actual distortion $D_{\text{actual}}(\pi)=\mathbb E_x\|\hat x_0(\pi)-x_0\|_2^2$ is measured on held-out clips (Figure~\ref{fig:proxy}).
Table~\ref{tab:proxy} reports the rank correlations within the budget, the actual rank of the plan the objective selects, and its regret against the best candidate.
On both models the price ranks the 21 plans nearly as the held-out error does (Spearman 0.944 and 0.917), and the plan the objective selects is the second best, 0.6\% and 1.2\% above the best candidate, as \hyperref[res:prop2]{Proposition~2} expects: the solver minimizes the price, and the price follows the actual distortion.

\begin{table}[!htbp]
\centering
\footnotesize
\setlength{\tabcolsep}{3pt}
\begin{tabular}{lccccc}
\toprule
Model & Plans & Spearman & Kendall & Rank & Regret (\%)\\
\midrule
CP2.5 & 21 & 0.944 & 0.819 & 2 & 0.6\\
T2.5 & 21 & 0.917 & 0.790 & 2 & 1.2\\
\bottomrule
\end{tabular}
\caption{\textbf{How well the priced distortion ranks plans.} CP2.5: Cosmos-Predict2.5; T2.5: Cosmos-Transfer2.5. Rank: actual rank of the selected plan among the candidates; regret: its excess actual distortion over the best candidate.}
\label{tab:proxy}
\end{table}

\noindent\textbf{Additivity.}
Eq.~\eqref{eq:alloc} sums the priced distortion over units.
For two groups of units $A$ and $B$, sparse error is injected into $A$ alone, $B$ alone, and both, giving final-latent errors $e_A$, $e_B$, and $e_{AB}$; Table~\ref{tab:additive} reports the nonlinearity $\|e_{AB}-e_A-e_B\|_2/(\|e_A\|_2+\|e_B\|_2)$ and the cross term $2e_A^\top e_B/(\|e_A\|_2^2+\|e_B\|_2^2)$ that summed squared errors omit.
Errors of different unit groups interact moderately: the cross term is 0.12--0.45 of the summed energies and smallest between the first and the last sparse step (Table~\ref{tab:additive}).
The summed price ranks the candidate plans as their held-out error does (Table~\ref{tab:proxy}), all the plan choice of Eq.~\eqref{eq:alloc} needs.

\begin{table}[!htbp]
\centering
\footnotesize
\setlength{\tabcolsep}{3pt}
\begin{tabular}{llcc}
\toprule
Model & Groups $A$, $B$ & Nonlinearity & Cross term\\
\midrule
CP2.5 & Adjacent steps & 0.436\sd{0.050} & 0.252\sd{0.055}\\
 & Early vs.\ late layers & 0.400\sd{0.057} & 0.165\sd{0.062}\\
 & First vs.\ last sparse step & 0.326\sd{0.044} & 0.122\sd{0.038}\\
T2.5 & Adjacent steps & 0.507\sd{0.087} & 0.394\sd{0.124}\\
 & Early vs.\ late layers & 0.607\sd{0.104} & 0.448\sd{0.215}\\
 & First vs.\ last sparse step & 0.496\sd{0.103} & 0.294\sd{0.135}\\
\bottomrule
\end{tabular}
\caption{\textbf{Interaction between unit groups} in the final latent (CP2.5: Cosmos-Predict2.5; T2.5: Cosmos-Transfer2.5). Mean over the test clips (32 and 39), small $\pm$: standard deviation.}
\label{tab:additive}
\end{table}

\noindent\textbf{Reference backend.}
The sensitivity of Eq.~\eqref{eq:sens} subtracts the run-to-run floor of the same dense attention backend; Table~\ref{tab:refstab} recomputes $J$ and the method ranking against a second dense attention backend as reference and counts the cells where the subtraction turns negative.
$J$ keeps its order under the second reference (Spearman 0.884 and 0.967), the method ranking stays the same (Kendall 1.000), and at most 6 of 81 cells turn negative (Table~\ref{tab:refstab}), so Distortion Budget measures the sensitivity of the model itself.

\begin{table}[!htbp]
\centering
\footnotesize
\setlength{\tabcolsep}{3pt}
\begin{tabular}{lccc}
\toprule
Model & Spearman of $J$ & Kendall of ranking & Negative cells\\
\midrule
CP2.5 & 0.884 & 1.000 & 6/81\\
T2.5 & 0.967 & 1.000 & 0/9\\
\bottomrule
\end{tabular}
\caption{\textbf{Stability under the dense reference} (CP2.5: Cosmos-Predict2.5; T2.5: Cosmos-Transfer2.5). Correlations between the two references; negative cells: $J$ cells whose floor-subtracted error is below zero under the second reference.}
\label{tab:refstab}
\end{table}

\subsection{Residual Structure}
\label{app:supp-resid}
\begin{figure}[!htbp]
\centering
\includegraphics[width=\linewidth]{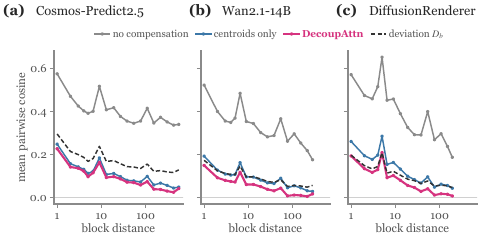}
\caption{\textbf{Residual correlation against block distance.} Mean pairwise cosine of the residuals of two skipped blocks, binned by the distance between the blocks, without compensation, with plain centroids, and with full DecoupAttn, and of the deviation terms $D_b$ of \hyperref[res:A8]{Proposition~A.8} (dashed), which are the same under every compensator.}
\label{fig:cosdist}
\end{figure}

\noindent\textbf{Exchangeability of the cluster members.}
Assumption (A4) of Appendix~\ref{app:th-setting} treats the block a cluster member falls into as independent of its deviation.
Figure~\ref{fig:cosdist} tests it by the mean pairwise cosine of two skipped blocks' residuals, and of their deviation terms $D_b$ of \hyperref[res:A8]{Proposition~A.8}, against their distance.
\hyperref[res:A10]{Theorem~A.10} bounds the coherence of the deviation terms by one under (A4), and the measured deviation terms add like random errors: their median coherence ratio is 0.97--1.72, against 4.86--8.25 for the residuals without compensation (Table~\ref{tab:beta}), and their correlation decays with the distance between the blocks (Figure~\ref{fig:cosdist}).
The cluster bias $\beta_{ib}=r_{ib}-D_b$ is the coherent part of the residual, and the corrections shrink it from 0.73--1.34 to 0.54--0.83 of the residual norm, with the variance correction giving nearly all of this reduction (Table~\ref{tab:beta}).

\begin{table}[!htbp]
\centering
\footnotesize
\setlength{\tabcolsep}{3pt}
\begin{tabular}{l l *{3}{C{1.0cm}}}
\toprule
Quantity & Compensation & CP2.5 & Wan2.1 & DR\\
\midrule
Coherence of $D_b$ & -- & 1.72 & 0.97 & 1.04\\
Coherence of $r_{ib}$ & None & 8.25 & 4.86 & 5.67\\
 & Centroid & 4.48 & 2.76 & 3.36\\
 & Variance only & 2.99 & 2.12 & 2.96\\
 & Value only & 4.20 & 2.55 & 2.62\\
 & Both (ours) & 2.30 & 1.64 & 1.67\\
\midrule
$\|\beta_{ib}\|_2/\|r_{ib}\|_2$ & Centroid & 1.34 & 0.73 & 0.97\\
 & Variance only & 0.82 & 0.53 & 0.71\\
 & Value only & 1.34 & 0.75 & 0.97\\
 & Both (ours) & 0.83 & 0.54 & 0.69\\
\bottomrule
\end{tabular}
\caption{\textbf{Coherence ratio and cluster bias} on saved attention inputs: medians over queries of the coherence ratio of $D_b$ and of the residual $r_{ib}$, and over skipped blocks of $\|\beta_{ib}\|_2/\|r_{ib}\|_2$.}
\label{tab:beta}
\end{table}

\begin{figure*}[!tp]
\centering
\includegraphics[width=\linewidth]{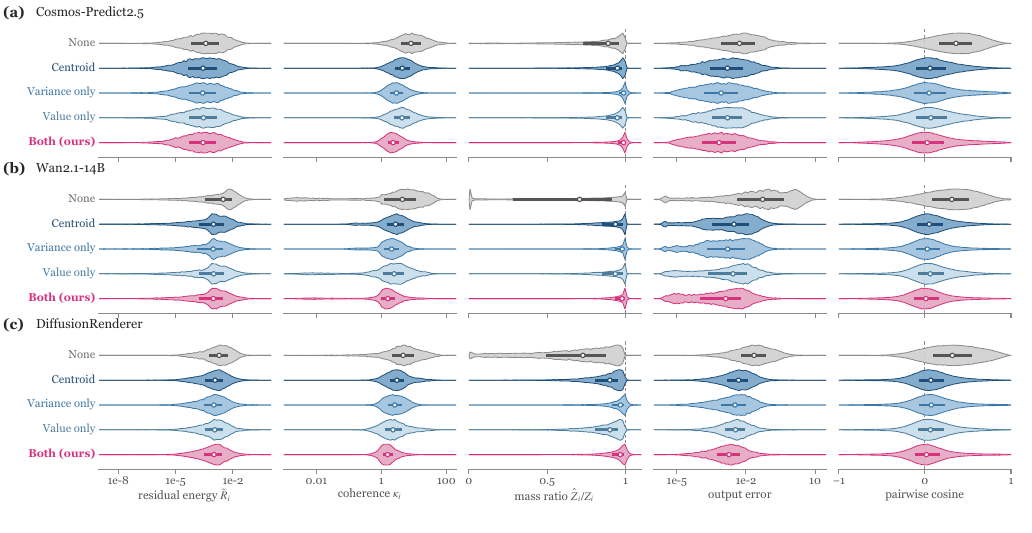}
\caption{\textbf{Distributions of the error factors} of Table~\ref{tab:resid} for its five compensators: residual energy $\tilde R_i$, coherence $\kappa_i$, mass ratio $\hat Z_i/Z_i$, output error, and the cosine between the residuals of two skipped blocks of the same query tile (held-out clips, 10{,}000 queries and 100{,}000 random pairs per compensator). Bars: interquartile range, dots: medians.}
\label{fig:resid}
\end{figure*}

\noindent\textbf{Energy, coherence, and mass.}
Eqs.~\eqref{eq:residual} and~\eqref{eq:coherence} give $\|\hat y_i-y_i\|_2^2=\kappa_iR_i/\hat Z_i^2$ with $R_i=\sum_{b\notin\mathcal S_g}\|r_{ib}\|_2^2$, so the output error splits into residual energy, coherence, and estimated mass.
Table~\ref{tab:resid} reports the three factors and the error for five compensators on the same held-out inputs and masks, next to the gap between fixed blocks and online routing of Table~\ref{tab:selection}.
By Eq.~\eqref{eq:coherence}, $\kappa_i$ grows with the pairwise cosine of the skipped-block residuals, and Figure~\ref{fig:resid} shows that compensation moves the median cosine from 0.31--0.36 to 0.012--0.027.
Unlike Table~\ref{tab:kappa}, $\kappa_i$ is taken per query through the deployed bf16 kernel with the error normalized by the estimated mass $\hat Z_i$, the residual energy is normalized as $\tilde R_i=R_i/(Z_i^2\|y_i\|_2^2)$, and the error is $\|\hat y_i-y_i\|_2^2/\|y_i\|_2^2$.
Compensation lowers all three factors, and the corrections act mainly on coherence and mass (Table~\ref{tab:resid}): from plain centroids to both corrections, the residual energy falls by at most 10\%, while $\kappa_i$ falls from 2.73--4.43 to 1.56--2.29 and the mass error from 0.051--0.100 to 0.016--0.035, so the error falls 2.2--2.6-fold.
The variance correction fixes the mass and also lowers $\kappa_i$, the value correction lowers $\kappa_i$ without changing the mass, and both together reach the lowest error on all three models, as Eq.~\eqref{eq:coherence} predicts.
Figure~\ref{fig:resid} shows these shifts across the whole distributions: with both corrections, $\kappa_i$ and $\hat Z_i/Z_i$ concentrate closer to one, the error falls at every quantile from the 25th to the 99th, and the distribution of the residual energy stays in place.

\begin{table}[!htbp]
\centering
\footnotesize
\setlength{\tabcolsep}{3pt}
\begin{tabular}{l *{5}{C{0.95cm}}}
\toprule
Compensation & $\tilde R_i$ & $\kappa_i$ & $|\hat Z_i/Z_i-1|$ & Error & Gap (dB)\\
\midrule
\rowcolor{groupgray}\multicolumn{6}{l}{\textit{DiffusionRenderer}}\\
None & 0.00191 & 4.71 & 0.270 & 0.0222 & --\\
Centroid & 0.00113 & 3.00 & 0.100 & 0.00478 & --\\
Variance only & 0.00104 & 2.59 & 0.0351 & 0.0034 & --\\
Value only & 0.00112 & 2.32 & 0.100 & 0.00354 & --\\
Both (ours) & 0.00102 & 1.56 & 0.0351 & 0.00186 & 1.87\\
\rowcolor{groupgray}\multicolumn{6}{l}{\textit{Wan2.1-14B}}\\
None & 0.00319 & 4.50 & 0.301 & 0.0570 & --\\
Centroid & 0.000973 & 2.73 & 0.0641 & 0.00295 & --\\
Variance only & 0.000924 & 2.03 & 0.0228 & 0.00166 & --\\
Value only & 0.000994 & 2.46 & 0.0641 & 0.00271 & --\\
Both (ours) & 0.000941 & 1.57 & 0.0228 & 0.00134 & 2.33\\
\rowcolor{groupgray}\multicolumn{6}{l}{\textit{Cosmos-Predict2.5}}\\
None & 0.000378 & 8.1 & 0.107 & 0.00503 & --\\
Centroid & 0.00028 & 4.43 & 0.0512 & 0.00158 & --\\
Variance only & 0.00027 & 2.92 & 0.0162 & 0.000873 & --\\
Value only & 0.00028 & 4.27 & 0.0512 & 0.00154 & --\\
Both (ours) & 0.000267 & 2.29 & 0.0162 & 0.000674 & 3.27\\
\bottomrule
\end{tabular}
\caption{\textbf{Factors of the output error} (medians over queries on held-out clips, deployed bf16 kernel).
Columns are defined in the text above, and the gap is the PSNR of online routing minus fixed blocks under the deployed compensator.}
\label{tab:resid}
\end{table}

\subsection{Reassign Cache}
\label{app:supp-cache}
\begin{figure*}[!tp]
\centering
\includegraphics[width=\linewidth]{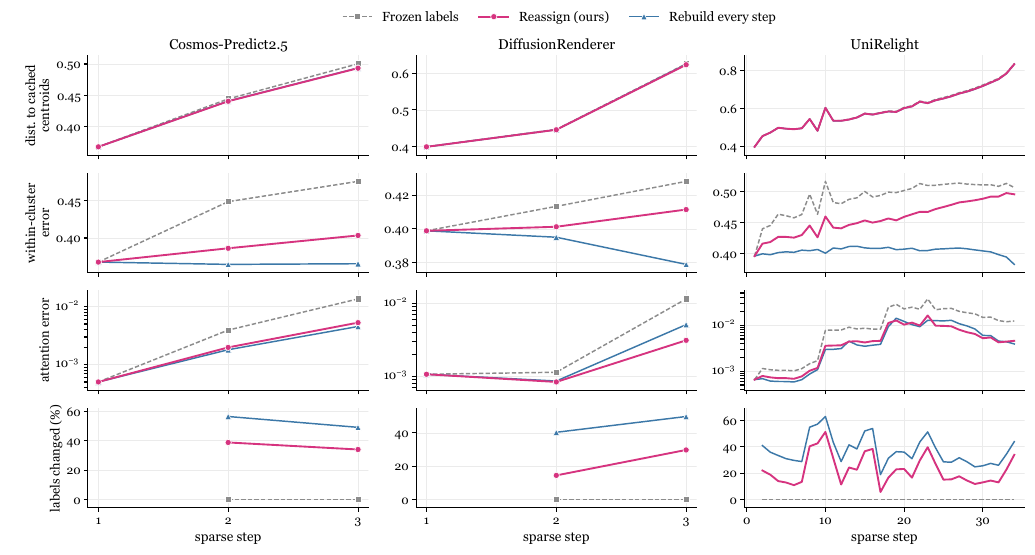}
\caption{\textbf{Assignment schemes across sparse steps} (Cosmos-Predict2.5, DiffusionRenderer, UniRelight). Reassignment stays at or below frozen assignments in distance to the cached centroids and cuts the within-cluster and attention errors that frozen assignments let grow. Step 1 builds the clusters.}
\label{fig:drift}
\end{figure*}

\noindent\textbf{Assignment update schemes.}
Every scheme in Table~\ref{tab:cache} recomputes the cluster statistics from the current keys and values (Eq.~\eqref{eq:comp}) and runs one exact plan; the schemes differ only in the cluster assignments.
Frozen keeps the assignments of the first sparse step, Reassign assigns keys to the cached centroids (Eq.~\eqref{eq:reassign}), Rebuild runs k-means at every sparse step, and Periodic rebuild runs it every 5 steps on the many-step models.
Figure~\ref{fig:drift} follows Frozen, Reassign, and Rebuild across the sparse steps: the distance to the cached centroids (the quantity of \hyperref[res:prop3]{Proposition~3}), the within-cluster error to the current means, the projected attention error, and the share of keys whose assignment changes.
Reassignment keeps the distance to the cached centroids at or below its frozen value, as \hyperref[res:prop3]{Proposition~3} states, and keeps each cluster tight around its current mean, so its attention error stays close to Rebuild's and far below Frozen's.
Reassign recovers most of what Rebuild gains over Frozen at a fraction of its cost (Table~\ref{tab:cache}): on UniRelight it lifts PSNR from 39.13 to 42.75\,dB against 43.18\,dB for Rebuild, changes 22.5\% of the assignments per step against 37.0\%, and adds 4.8\,s of DiT time instead of 36.7\,s; on Cosmos-Predict2.5 it recovers 0.47 of Rebuild's 0.52\,dB for 0.15\,s instead of 1.09\,s.
Periodic rebuild reaches Rebuild's accuracy (43.16\,dB) for 10.0\,s, still twice the cost of Reassign, and on DiffusionRenderer, where stale assignments cost only 0.46\,dB on calibration clips (Table~\ref{tab:rule}), the three schemes lie within 0.07\,dB of each other.

\begin{table*}[!tp]
\centering
\footnotesize
\setlength{\tabcolsep}{3pt}
\begin{tabular}{l *{9}{C{1.2cm}}}
\toprule
 & \multicolumn{3}{c}{Cosmos-Predict2.5} & \multicolumn{3}{c}{DiffusionRenderer} & \multicolumn{3}{c}{UniRelight}\\
\cmidrule(lr){2-4}\cmidrule(lr){5-7}\cmidrule(lr){8-10}
Scheme & PSNR & Changed (\%) & DiT ms & PSNR & Changed (\%) & DiT ms & PSNR & Changed (\%) & DiT ms\\
\midrule
Frozen & 24.23 & 0.0 & 8{,}057 & 50.87 & 0.0 & 3{,}913 & 39.13 & 0.0 & 103{,}836\\
Reassign & 24.70 & 36.5 & 8{,}208 & 50.91 & 22.1 & 4{,}020 & 42.75 & 22.5 & 108{,}671\\
Rebuild & 24.75 & 52.9 & 9{,}148 & 50.94 & 45.1 & 4{,}736 & 43.18 & 37.0 & 140{,}497\\
Periodic rebuild & -- & -- & -- & -- & -- & -- & 43.16 & 22.1 & 113{,}839\\
\bottomrule
\end{tabular}
\caption{\textbf{Assignment update schemes under one exact plan.}
All schemes recompute cluster statistics from current keys and values, and Changed is the share of keys changing assignment per sparse step.
DiT ms: median of interleaved runs (periodic rebuild: mean of two).}
\label{tab:cache}
\end{table*}

\noindent\textbf{Deployment rule.}
Table~\ref{tab:rule} checks the choices of Eq.~\eqref{eq:rules} on test clips: the staleness measured on calibration clips, the scheme the rule selects, the test PSNR gain and latency cost of that scheme over the other, and whether the staleness measured on test clips falls on the same side of $\tau_s$.
On test clips the selected scheme gains 0.14--3.63\,dB on all nine models that reassign, and DiffusionRenderer, which keeps its assignments, gives up 0.06\,dB for 102\,ms.
The calibration staleness overstates the test gain, so the two fall on the same side of $\tau_s$ on five of the ten models.

\begin{table*}[!tp]
\centering
\footnotesize
\setlength{\tabcolsep}{3pt}
\begin{tabular}{l *{5}{C{2.0cm}}}
\toprule
Model & $\delta_{\text{stale}}$ (dB) & Selected & Test gain (dB) & Time cost (ms) & Agrees\\
\midrule
Cosmos-Predict1 & 2.35 & reassign & 1.28 & 7{,}484 & yes\\
Cosmos-Predict2.5 & 3.00 & reassign & 0.47 & 150 & no\\
Wan2.1-14B & 2.73 & reassign & 1.09 & 218 & yes\\
Wan2.2-A14B & 4.35 & reassign & 0.41 & 255 & no\\
HunyuanVideo & 2.86 & reassign & 0.76 & 26{,}463 & no\\
Cosmos3-Nano (I2W) & 2.86 & reassign & 0.31 & 8{,}050 & no\\
Cosmos-Transfer2.5 & 2.13 & reassign & 0.14 & 167 & no\\
DiffusionRenderer & 0.46 & frozen & $-$0.06 & $-$102 & yes\\
AlayaRenderer & 2.43 & reassign & 2.40 & 2{,}850 & yes\\
UniRelight & 1.87 & reassign & 3.63 & 4{,}780 & yes\\
\bottomrule
\end{tabular}
\caption{\textbf{Assignment rule on test clips.} Gain and cost of the selected scheme over the other; agrees: whether the test-clip staleness falls on the same side of $\tau_s$ as the calibration staleness. For HunyuanVideo and Cosmos3-Nano this table re-runs the calibration from scratch, with a reassign PSNR of 27.25 and 21.95\,dB.}
\label{tab:rule}
\end{table*}

\subsection{Efficiency}
\label{app:supp-eff}

\begin{table*}[!tp]
\centering
\footnotesize
\setlength{\tabcolsep}{3pt}
\begin{tabular}{l *{10}{C{1.1cm}}}
\toprule
 & \multicolumn{7}{c}{Attention} & & & \\
\cmidrule(lr){2-8}
Method & Exact & k-means & Assign & Statistics & Counts & Merge & Total & DiT & E2E & Peak GiB\\
\midrule
\rowcolor{groupgray}\multicolumn{11}{l}{\textit{Cosmos-Predict2.5}}\\
Dense & 9{,}352 & -- & -- & -- & -- & -- & 9{,}352 & 12{,}146 & 19{,}682 & 25.4\\
SVG-EAR & 2{,}294 & 4{,}730 & 133 & 23 & 762 & -- & 10{,}687 & 13{,}347 & 20{,}964 & 27.3\\
Ours & 1{,}291 & 620 & 171 & 96 & 46 & 260 & 5{,}484 & 8{,}201 & 15{,}706 & 38.2\\
\rowcolor{groupgray}\multicolumn{11}{l}{\textit{Wan2.1-14B}}\\
Dense & 4{,}842 & -- & -- & -- & -- & -- & 4{,}842 & 13{,}649 & 16{,}399 & 35.9\\
SVG-EAR & 1{,}822 & 7{,}142 & 188 & 28 & 1{,}110 & -- & 11{,}905 & 20{,}520 & 23{,}272 & 37.6\\
Ours & 843 & 923 & 240 & 150 & 62 & 460 & 4{,}266 & 12{,}908 & 15{,}658 & 59.4\\
\rowcolor{groupgray}\multicolumn{11}{l}{\textit{UniRelight}}\\
Dense & 68{,}424 & -- & -- & -- & -- & -- & 68{,}424 & 141{,}099 & 144{,}398 & 34.1\\
SVG-EAR & 29{,}514 & 18{,}435 & 2{,}845 & 444 & 17{,}592 & -- & 77{,}650 & 144{,}762 & 147{,}921 & 36.0\\
Ours & 14{,}157 & 1{,}218 & 5{,}403 & 2{,}534 & 864 & 6{,}466 & 39{,}694 & 108{,}432 & 111{,}718 & 41.1\\
\bottomrule
\end{tabular}
\caption{\textbf{Time breakdown of one video} (ms, median over interleaved runs).
k-means is the first clustering, Counts the block--cluster counts, and E2E covers encoders, DiT, VAE decoding, and transfers.
SVG-EAR has no merge, and its Counts column holds its error map and top-$p$ block selection.
Peak GiB is the maximum allocated GPU memory over the whole generation, measured with one method per process and the text embeddings precomputed.}
\label{tab:timing}
\end{table*}

\begin{table*}[!tp]
\centering
\footnotesize
\setlength{\tabcolsep}{3pt}
\begin{tabular}{l l *{8}{C{1.25cm}}}
\toprule
Model & Method & Subject & Background & Flickering & Smoothness & Dynamic & Imaging & Aesthetic & Overall\\
\midrule
Cosmos-Predict1 & Dense & 0.9451 & 0.9541 & 0.9872 & 0.9945 & 0.2821 & 0.6796 & 0.4644 & 0.2224\\
 & SVG-EAR & 0.9387 & 0.9514 & 0.9860 & 0.9944 & 0.2308 & 0.6835 & 0.4620 & 0.2269\\
 & Ours & 0.9325 & 0.9494 & 0.9870 & 0.9945 & 0.2051 & 0.6775 & 0.4469 & 0.2230\\
\midrule
Cosmos-Predict2.5 & Dense & 0.9430 & 0.9574 & 0.9722 & 0.9878 & 0.5312 & 0.7087 & 0.6047 & 0.2864\\
 & SVG-EAR & 0.9445 & 0.9563 & 0.9715 & 0.9875 & 0.5312 & 0.7124 & 0.6005 & 0.2885\\
 & Ours & 0.9451 & 0.9588 & 0.9732 & 0.9884 & 0.5000 & 0.7168 & 0.6054 & 0.2864\\
\midrule
Wan2.1-14B & Dense & 0.9334 & 0.9537 & 0.9633 & 0.9853 & 0.5625 & 0.7052 & 0.6601 & 0.2586\\
 & SVG-EAR & 0.9335 & 0.9554 & 0.9636 & 0.9852 & 0.5312 & 0.7067 & 0.6604 & 0.2593\\
 & Ours & 0.9343 & 0.9536 & 0.9634 & 0.9854 & 0.5312 & 0.7082 & 0.6588 & 0.2583\\
\midrule
Wan2.2-A14B & Dense & 0.9162 & 0.9457 & 0.9552 & 0.9788 & 0.7500 & 0.7092 & 0.6787 & 0.2671\\
 & SVG-EAR & 0.9181 & 0.9482 & 0.9552 & 0.9787 & 0.7500 & 0.7044 & 0.6759 & 0.2671\\
 & Ours & 0.9172 & 0.9492 & 0.9563 & 0.9795 & 0.7500 & 0.7093 & 0.6718 & 0.2679\\
\midrule
HunyuanVideo & Dense & 0.9165 & 0.9636 & 0.9875 & 0.9946 & 0.6250 & 0.6753 & 0.5578 & 0.2569\\
 & SVG-EAR & 0.9241 & 0.9664 & 0.9881 & 0.9944 & 0.6250 & 0.6654 & 0.5610 & 0.2587\\
 & Ours & 0.9182 & 0.9583 & 0.9881 & 0.9946 & 0.6250 & 0.6647 & 0.5567 & 0.2585\\
\midrule
Cosmos3-Nano (I2W) & Dense & 0.9322 & 0.9466 & 0.9800 & 0.9936 & 0.5625 & 0.7182 & 0.5968 & 0.2406\\
 & SVG-EAR & 0.8440 & 0.8975 & 0.9663 & 0.9852 & 0.5312 & 0.6606 & 0.5064 & 0.2374\\
 & Ours & 0.9318 & 0.9466 & 0.9804 & 0.9938 & 0.5625 & 0.7178 & 0.5992 & 0.2428\\
\bottomrule
\end{tabular}
\caption{\textbf{VBench dimensions.} Subject and background consistency, temporal flickering, motion smoothness, dynamic degree, imaging and aesthetic quality, and overall consistency (text--video alignment).}
\label{tab:vbench}
\end{table*}

\begin{figure*}[!tp]
\centering
\includegraphics[width=\linewidth]{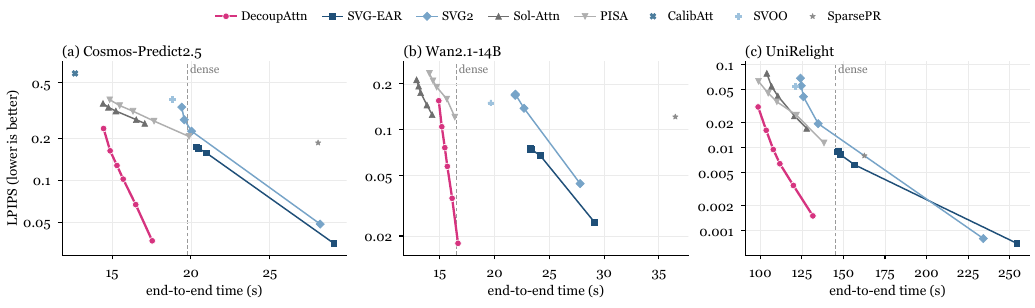}
\caption{\textbf{LPIPS against end-to-end latency} for the operating points of Figure~\ref{fig:tradeoff}b--d; dashed: dense attention.
End-to-end time covers text encoding, DiT, and VAE decoding (UniRelight: the whole generation).}
\label{fig:pareto}
\end{figure*}

\noindent\textbf{Quality--latency curves.}
Each method runs at four or five operating points of its own density knob, with the density of every point measured.
Figure~\ref{fig:tradeoff}b--d plots PSNR against DiT speedup, and Figure~\ref{fig:pareto} plots LPIPS against end-to-end latency for the same points.
In LPIPS the picture is the same: at every latency of DecoupAttn all baselines lie above its curve, and the lower LPIPS of SVG-EAR and SVG2 at top-$p$ 0.99 costs 1.5--1.8 times the dense end-to-end time.
Table~\ref{tab:iso} reads two numbers off them: the PSNR gain at the latency of the strongest baseline, and the speedup at a target PSNR fixed before testing.
At the latency of the strongest baseline, DecoupAttn is 5.74--8.40\,dB more accurate, and the strongest baseline does not reach our deployed PSNR at any of its operating points, so its speedup at equal PSNR is undefined (Table~\ref{tab:iso}).

\begin{table}[!htbp]
\centering
\footnotesize
\setlength{\tabcolsep}{3pt}
\begin{tabular}{l *{4}{C{1.2cm}}}
\toprule
Model & Baseline & $\Delta$PSNR at equal latency & Speedup at equal PSNR & Target PSNR\\
\midrule
CP2.5 & Sol-Attn & +6.40 & -- & 24.70\\
Wan2.1 & PISA & +5.74 & -- & 24.70\\
UniRelight & PISA & +8.40 & -- & 42.75\\
\bottomrule
\end{tabular}
\caption{\textbf{Matched-latency and matched-quality comparison} against the strongest baseline of Figure~\ref{fig:pareto} (CP2.5: Cosmos-Predict2.5; Wan2.1: Wan2.1-14B). --: the baseline does not reach the target PSNR at any operating point.}
\label{tab:iso}
\end{table}

\noindent\textbf{Time breakdown.}
Table~\ref{tab:timing} splits the time of one video into attention (exact blocks, first k-means, assignment, cluster statistics, block--cluster counts, and the fused merge), the DiT, and the full generation (encoders, DiT, VAE decoding, and transfers).
Each entry is the median of five interleaved runs on one NVIDIA H800 GPU with batch size 1 and bf16 precision, without compilation or cold start; Table~\ref{tab:offline} accounts offline calibration, plan solving, and plan loading separately and gives the break-even count
\begin{equation*}
N_{\text{break-even}}=\Big\lceil\frac{T_{\text{calibration}}}{T_{\text{dense, E2E}}-T_{\text{ours, E2E}}}\Big\rceil,
\end{equation*}
the smallest number of videos whose savings repay the calibration, for $T_{\text{ours, E2E}}<T_{\text{dense, E2E}}$.

\noindent\textbf{Where the time goes.}
The gap to SVG-EAR comes mostly from the per-step work (Table~\ref{tab:timing}): SVG-EAR re-clusters at every step and rebuilds its error map and top-$p$ selection, which costs 4.7--18.4\,s of k-means and 0.8--17.6\,s of selection per video, while DecoupAttn clusters once (0.6--1.2\,s) and later only reassigns, counts, and merges.
Its attention therefore takes 1.1--1.7 times less time than dense attention and 1.9--2.8 times less than SVG-EAR's, and every video finishes faster end to end.
Calibration takes 5.3--19.6 GPU-hours and the plan solve under a minute (Table~\ref{tab:offline}), and the break-even count follows the end-to-end saving per video: UniRelight, which saves the most, breaks even after 2.1k videos, and Wan2.1-14B, which saves the least, after 23.7k.

\begin{table}[!htbp]
\centering
\footnotesize
\setlength{\tabcolsep}{2pt}
\begin{tabular}{l *{5}{C{1.0cm}}}
\toprule
Model & Calib.\ (GPU-h) & Peak mem.\ (GiB) & Plan solve (min) & Saving (s / video) & $N_{\text{break-even}}$\\
\midrule
CP2.5 & 5.93 & 38.7 & 0.60 & 4.06 & 5{,}263\\
Wan2.1 & 5.28 & 39.3 & 0.47 & 0.80 & 23{,}664\\
UniRelight & 19.62 & 54.4 & 0.80 & 33.66 & 2{,}099\\
\bottomrule
\end{tabular}
\caption{\textbf{Offline cost and break-even count} (CP2.5: Cosmos-Predict2.5; Wan2.1: Wan2.1-14B). Calib.: block ranking, distortion table, and error calibration of the deployed plan, without the sensitivity measurement. Peak memory: GPU memory during calibration. Saving: end-to-end time saved per video.}
\label{tab:offline}
\end{table}

\subsection{Generalization of the Offline Plan}
\label{app:supp-gen}
\noindent\textbf{Calibration set size.}
Table~\ref{tab:calsize} calibrates on 1, 2, 4, 8, and 16 training clips with three draws each and reports the test PSNR, the intersection over union of the exact blocks with those of the 16-clip plan, and the calibration time.
PSNR rises from 24.59 to 24.87\,dB as the set grows from 1 to 16 clips, and its spread over draws shrinks from 0.22\,dB with one clip to 0.04\,dB with eight (Table~\ref{tab:calsize}); with the deployed four clips, the exact blocks already overlap 94.8\% with the 16-clip plan after 0.33 GPU-hours of error calibration.

\begin{table}[!htbp]
\centering
\footnotesize
\setlength{\tabcolsep}{3pt}
\begin{tabular}{c *{3}{C{1.6cm}}}
\toprule
Clips & PSNR & Block overlap & Calibration (GPU-h)\\
\midrule
1 & 24.59 $\pm$ 0.22 & 0.875 & 0.08\\
2 & 24.67 $\pm$ 0.18 & 0.922 & 0.17\\
4 & 24.80 $\pm$ 0.09 & 0.948 & 0.33\\
8 & 24.87 $\pm$ 0.04 & 0.968 & 0.67\\
16 & 24.87 & 1.000 & 1.34\\
\bottomrule
\end{tabular}
\caption{\textbf{Calibration set size} (Cosmos-Predict2.5; mean $\pm$ standard deviation over three draws). Only the error calibration uses the drawn clips, and the sensitivity and the distortion table are those of the deployed plan. The 4-clip draws are training clips 0--3, 4--7, and 8--11, and the first is the deployed plan. Block overlap: intersection over union of the exact blocks with the 16-clip plan. Calibration counts error calibration alone (301\,s per clip), while Table~\ref{tab:offline} adds block ranking and the distortion table.}
\label{tab:calsize}
\end{table}

\noindent\textbf{Unseen inputs and settings.}
With the plan frozen, Table~\ref{tab:shift} groups held-out prompts by content and reports the mean and the worst decile.
The plan is tied to the token grid of one resolution, so a new resolution runs a new calibration, and Table~\ref{tab:settings} reports its cost for Wan2.1-14B at 720p.
The frozen plan leads SVG-EAR on every content group by 1.22--3.79\,dB, including categories absent from calibration (+3.39\,dB) and complex lighting (+3.79\,dB, Table~\ref{tab:shift}).
Its worst decile stays at 19.4--27.6\,dB in every group (Table~\ref{tab:shift}).
The plan fixes only where exact attention runs, while each video clusters its own keys at its first sparse step, so new content changes the compensation, not the plan.
At 720p the recalibrated plan reaches 23.89\,dB with the same procedure and constants after 18.70 GPU-hours of calibration, 3.5 times that of the 480p plan (Table~\ref{tab:settings}).

\begin{table}[!htbp]
\centering
\footnotesize
\setlength{\tabcolsep}{3pt}
\begin{tabular}{l *{4}{C{1.1cm}}}
\toprule
Content & Clips & Mean & Worst 10\% & vs.\ SVG-EAR\\
\midrule
Unseen categories & 6 & 24.77 & 22.68 & +3.39\\
Fast motion & 7 & 23.35 & 19.39 & +3.25\\
Several subjects & 7 & 24.95 & 19.39 & +3.26\\
Camera motion & 7 & 22.22 & 19.39 & +1.22\\
Fine texture & 6 & 24.64 & 20.64 & +1.52\\
Complex lighting & 6 & 29.72 & 27.62 & +3.79\\
\bottomrule
\end{tabular}
\caption{\textbf{Frozen plan on held-out content} (Cosmos-Predict2.5; PSNR, dB). vs.\ SVG-EAR: mean paired difference.}
\label{tab:shift}
\end{table}

\begin{table}[!htbp]
\centering
\footnotesize
\setlength{\tabcolsep}{3pt}
\begin{tabular}{l *{2}{C{1.8cm}}}
\toprule
Resolution & PSNR (dB) & Calibration (GPU-h)\\
\midrule
480p (deployed) & 24.70 & 5.28\\
720p (recalibrated) & 23.89 & 18.70\\
\bottomrule
\end{tabular}
\caption{\textbf{Recalibration cost for a new resolution} (Wan2.1-14B). 480p: the deployed plan of Table~\ref{tab:main}, with its cost from Table~\ref{tab:offline}.}
\label{tab:settings}
\end{table}

\noindent\textbf{Cluster count and k-means rounds.}
Table~\ref{tab:clusters} varies the clusters per head over 220, 440, 880, and 1{,}760 and the k-means rounds of the first sparse step.
More clusters per head raise PSNR steadily, from 23.88 to 25.18\,dB, because smaller clusters have members closer to their centroid and leave smaller residuals (\hyperref[res:A6]{Proposition~A.6}), while k-means time grows from 404 to 927\,ms and DiT time from 7.78 to 8.78\,s; 880 clusters keep all but 0.48\,dB of the 1{,}760-cluster gain at 0.59\,s less DiT time.
The k-means rounds matter little, 3 rounds losing 0.06\,dB and 20 rounds gaining 0.01\,dB, since every later call recomputes the cluster statistics.

\begin{table}[!htbp]
\centering
\footnotesize
\setlength{\tabcolsep}{3pt}
\begin{tabular}{cc *{4}{C{1.05cm}}}
\toprule
$C$ & Rounds & PSNR & LPIPS & k-means ms & DiT ms\\
\midrule
220 & 10 & 23.88 & 0.116 & 404 & 7{,}778\\
440 & 10 & 24.24 & 0.109 & 468 & 7{,}901\\
880 (ours) & 10 & 24.70 & 0.102 & 625 & 8{,}187\\
1{,}760 & 10 & 25.18 & 0.095 & 927 & 8{,}778\\
880 & 3 & 24.64 & 0.103 & 278 & 7{,}879\\
880 & 20 & 24.71 & 0.101 & 1{,}090 & 8{,}633\\
\bottomrule
\end{tabular}
\caption{\textbf{Clusters per head and k-means rounds} (Cosmos-Predict2.5). k-means ms: the first clustering per video.}
\label{tab:clusters}
\end{table}

\subsection{Task Quality}
\label{app:supp-task}
\noindent\textbf{VBench dimensions.}
Table~\ref{tab:vbench} reports each VBench dimension~\cite{huang2024vbench} together with dynamic degree, the amount of motion, and overall consistency, the text--video alignment score.
DecoupAttn tracks dense attention in every dimension on the four-step models (Table~\ref{tab:vbench}), and on Cosmos3-Nano it keeps every dimension within 0.003 of dense attention, while SVG-EAR drops subject consistency from 0.932 to 0.844 and imaging quality from 0.718 to 0.661.
Dynamic degree stays within one video of the dense value on every model but Cosmos-Predict1, where it drops by three of 39 videos (SVG-EAR: two), so the sparse videos keep nearly all the motion of the dense outputs.

\noindent\textbf{Conditions and ground truth.}
Table~\ref{tab:cond} measures edge adherence on Cosmos-Transfer2.5, the F1 score between the input edges and the edges of the output, and the error against ground truth where the rendering test sets provide it.
DecoupAttn keeps edge adherence at the dense level (0.423 against 0.421) and matches dense attention against ground truth on both renderers (27.48 against 27.49\,dB and 12.08 against 12.17\,dB), above SVG-EAR on all three (Table~\ref{tab:cond}), so its fidelity to the dense output carries over to the input conditions and to the ground truth.

\begin{table}[!htbp]
\centering
\footnotesize
\setlength{\tabcolsep}{3pt}
\begin{tabular}{l *{3}{C{1.2cm}}}
\toprule
 & Edge F1 & \multicolumn{2}{c}{PSNR to ground truth}\\
\cmidrule(lr){3-4}
Method & T2.5 & DR & Alaya\\
\midrule
Dense & 0.421 & 27.49 & 12.17\\
SVG-EAR & 0.410 & 27.43 & 11.94\\
Ours & 0.423 & 27.48 & 12.08\\
\bottomrule
\end{tabular}
\caption{\textbf{Condition adherence and ground-truth error.} T2.5: Cosmos-Transfer2.5; DR: DiffusionRenderer; Alaya: AlayaRenderer.}
\label{tab:cond}
\end{table}

\FloatBarrier
\section{Case Studies}
\label{app:cases}

Four clips show where the method is stressed: two are preselected by a measurable property before generation (fast motion with fine texture, and occlusion), and two are the clips on which DecoupAttn scores lowest.
Each figure shows the same frames from dense attention, SVG-EAR, and DecoupAttn, generated with the main-table configuration at density 0.20 and seed 1000.

\begin{figure*}[p]
\centering
\includegraphics[width=0.9\linewidth]{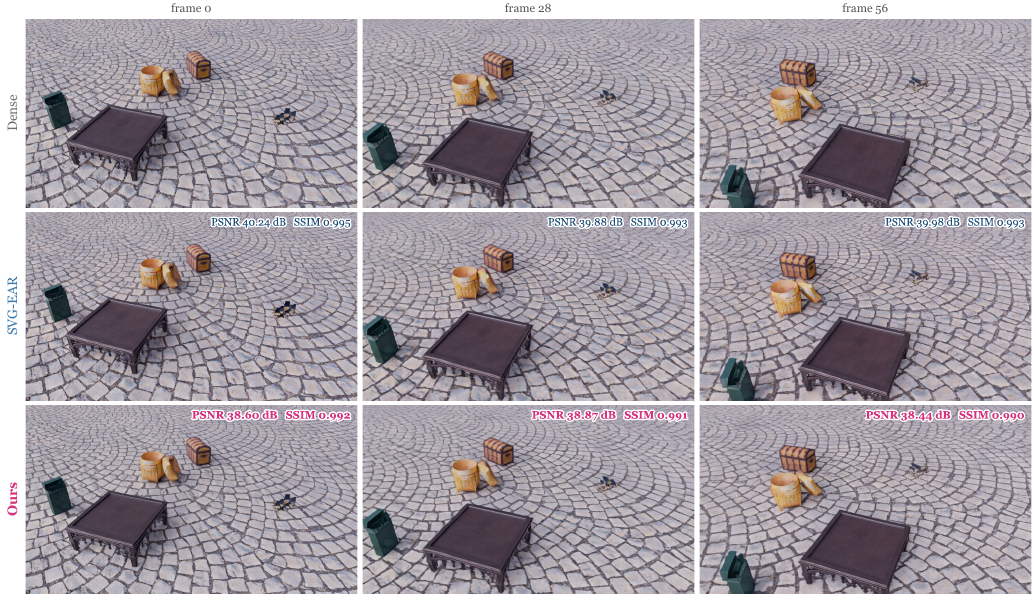}
\caption{\textbf{Case 1: fast motion} (UniRelight, syn\_test\_0004, frames 0, 28, and 56). Rows: dense attention, SVG-EAR, and DecoupAttn, with per-frame PSNR and SSIM against dense attention.}
\label{fig:case1}
\end{figure*}
\begin{figure*}[p]
\centering
\includegraphics[width=0.9\linewidth]{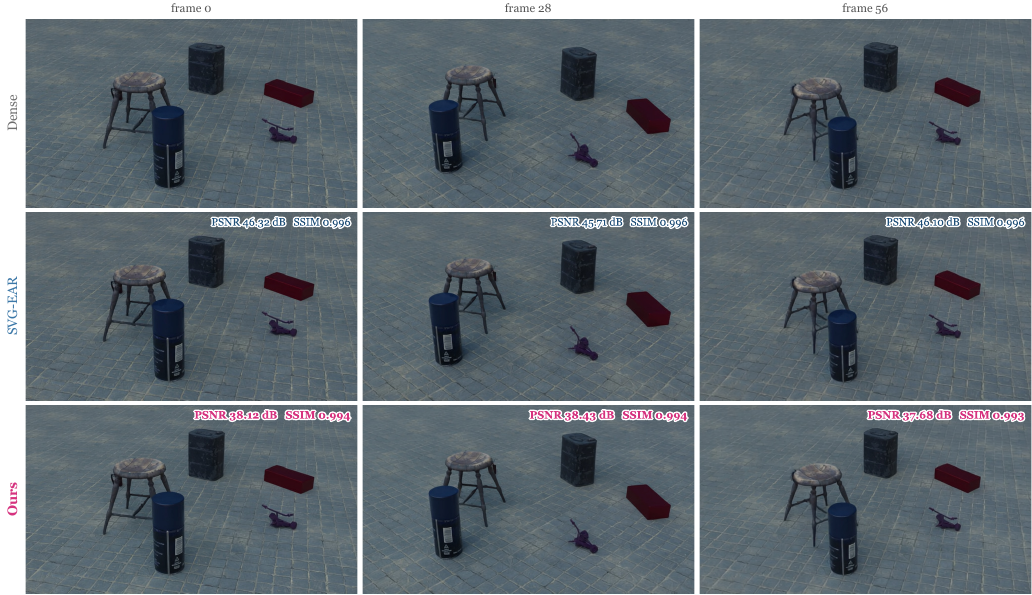}
\caption{\textbf{Case 2: occlusion} (UniRelight, syn\_test\_0006, frames 0, 28, and 56). Rows: dense attention, SVG-EAR, and DecoupAttn, with per-frame PSNR and SSIM against dense attention.}
\label{fig:case2}
\end{figure*}
\begin{figure*}[p]
\centering
\includegraphics[width=0.9\linewidth]{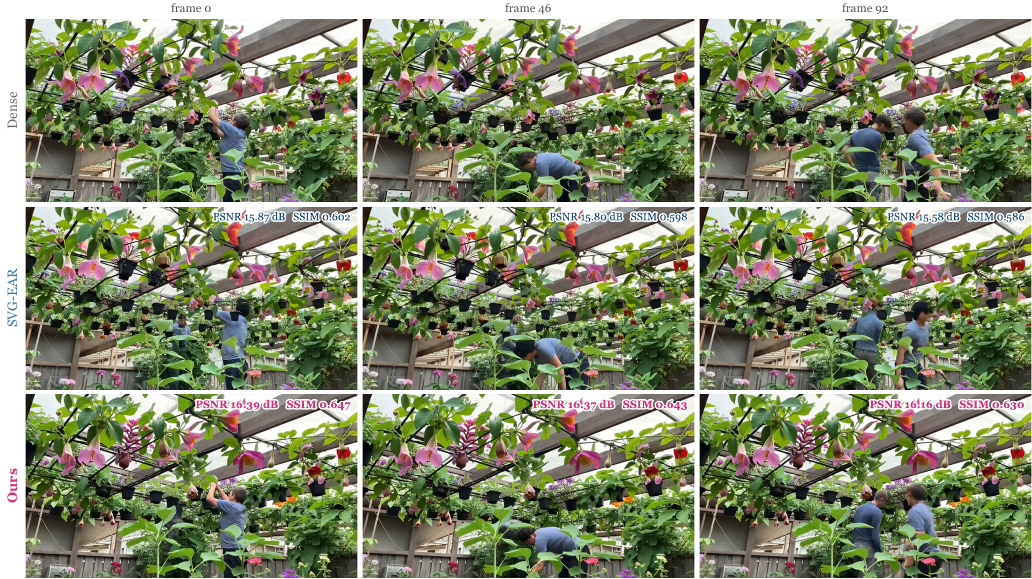}
\caption{\textbf{Case 3: worst clip of a few-step model} (Cosmos-Predict2.5, test\_19, frames 0, 46, and 92). Rows: dense attention, SVG-EAR, and DecoupAttn, with per-frame PSNR and SSIM against dense attention.}
\label{fig:case3}
\end{figure*}
\begin{figure*}[p]
\centering
\includegraphics[width=0.9\linewidth]{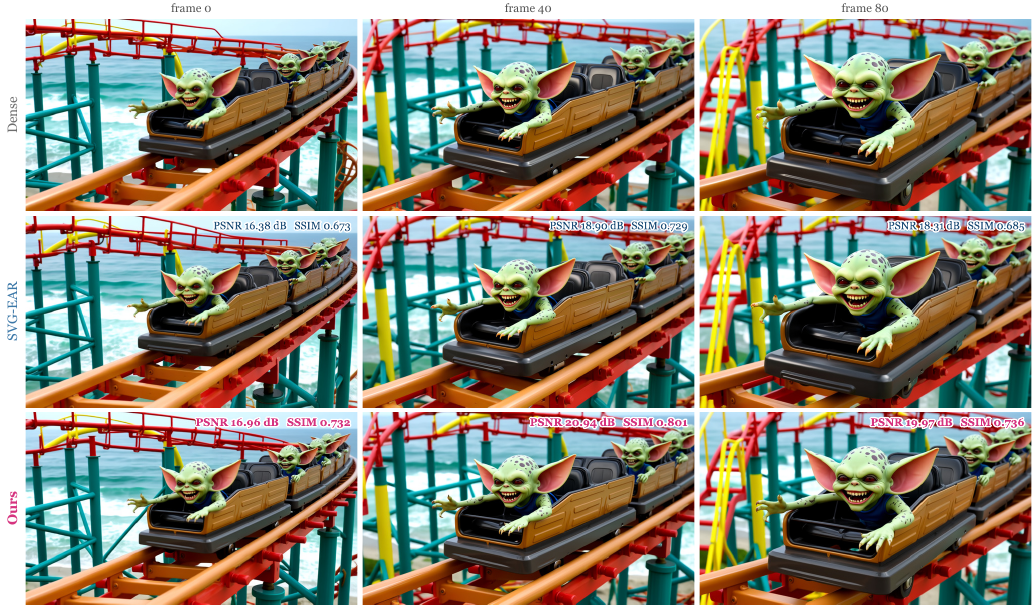}
\caption{\textbf{Case 4: worst clip of a distilled 14B model} (Wan2.1-14B, test\_26, frames 0, 40, and 80). Rows: dense attention, SVG-EAR, and DecoupAttn, with per-frame PSNR and SSIM against dense attention.}
\label{fig:case4}
\end{figure*}

\noindent\textbf{Case 1: Fast motion (UniRelight, syn\_test\_0004).}
It has the largest mean absolute frame difference of the dense output among the 39 test clips, and the largest mean Laplacian variance, so it combines fast motion with fine texture.
Figure~\ref{fig:case1} shows frames 0, 28, and 56 of this clip.

\noindent\textbf{Key Observations:} Both sparse outputs are indistinguishable from the dense frames: DecoupAttn scores 38.4--38.9\,dB per frame and SVG-EAR 39.9--40.2\,dB (38.76 and 39.94\,dB over the clip).
The remaining error of both methods lies on the edges of the cobblestones and the furniture, the fine texture that the motion carries across blocks.
SSIM stays at 0.990--0.992 on every frame for DecoupAttn and 0.993--0.995 for SVG-EAR, so both keep the structure of the dense frames.
The error of DecoupAttn has no block shape although its exact blocks are fixed: the compensated residuals of the skipped blocks cancel like dithered quantization errors (\hyperref[res:prop1]{Proposition~1}, Figure~\ref{fig:obs}a--c), so what remains spreads as fine noise over the texture instead of gathering at block borders.
UniRelight is also the model on which keeping the clusters current matters most (Table~\ref{tab:cache}), and on its fastest clip DecoupAttn, which reassigns on this model, keeps every frame above 38\,dB.

\noindent\textbf{Case 2: Occlusion (UniRelight, syn\_test\_0006).}
It has the largest share of pixels whose G-buffer depth changes by more than 10\% of its range between frames, so objects keep covering and uncovering each other.
Figure~\ref{fig:case2} shows frames 0, 28, and 56 of this clip.

\noindent\textbf{Key Observations:} SVG-EAR stays closest to dense here (45.7--46.3\,dB per frame against 37.7--38.4\,dB for DecoupAttn, 46.02 and 38.18\,dB over the clip), the largest gap of the four cases.
The difference of DecoupAttn is a faint brightness offset spread evenly over the floor, together with thin lines along the outlines of the stool, the cans, and the box where the objects occlude one another.
The objects and the floor texture match the dense output in every frame.
The brightness offset is the visible form of an error in the estimated mass: the exact and the compensated parts share one softmax denominator (Eq.~\eqref{eq:decoup}), so a small error in $\hat Z_i$ scales a whole region evenly instead of leaving a local artifact, and SSIM stays at 0.993--0.994 on every frame.
The variance correction of Eq.~\eqref{eq:comp} targets exactly this mass error and cuts it about threefold on the three models of Table~\ref{tab:resid}, so the offset stays faint.

\noindent\textbf{Case 3: Worst clip of a few-step model (Cosmos-Predict2.5, test\_19).}
It is the clip on which DecoupAttn has its lowest PSNR among the 32 test clips (16.39\,dB).
Figure~\ref{fig:case3} shows frames 0, 46, and 92 of this clip.

\noindent\textbf{Key Observations:} SVG-EAR scores lower still on this clip (15.85\,dB over the clip, 15.6--15.9\,dB per frame against 16.2--16.4\,dB for DecoupAttn).
Both sparse outputs keep the greenhouse, the hanging baskets, and the person bending over the plants, but they redraw the dense canopy of small leaves and flowers at shifted positions, so the error covers the whole canopy evenly.
The four-step sampler amplifies early differences the most (Figure~\ref{fig:obs}d), and a prompt with this much fine texture turns them into a low PSNR for every sparse method.
DecoupAttn keeps a higher SSIM on every frame (0.630--0.647 against 0.586--0.602), so even where both methods redraw the leaves it keeps more of the dense layout.
The even spread of its error is what the cancellation of \hyperref[res:prop1]{Proposition~1} predicts for compensated residuals, and it keeps the worst clip of Cosmos-Predict2.5 free of local failures.

\noindent\textbf{Case 4: Worst clip of a distilled 14B model (Wan2.1-14B, test\_26).}
It is the clip on which DecoupAttn has its lowest PSNR among the 32 test clips (19.95\,dB).
Figure~\ref{fig:case4} shows frames 0, 40, and 80 of this clip.

\noindent\textbf{Key Observations:} DecoupAttn again stays ahead of SVG-EAR (19.95 against 18.24\,dB over the clip).
The first frame is the hardest for both (16.96 and 16.38\,dB), and later frames recover (20.94 and 18.90\,dB at frame 40).
Both outputs keep the goblin, the car, and the coaster and shift the thin rails and struts of the background, while SVG-EAR also departs more on the goblin's face and the front car, which DecoupAttn keeps close to the dense frames.
Per frame, DecoupAttn leads by 0.58, 2.04, and 1.66\,dB and keeps SSIM at 0.732--0.801 against 0.673--0.729, so its lead holds on every frame of the clip.

\noindent\textbf{Summary.}
Across the four cases, from two video models and a renderer, the error of DecoupAttn is spread, not local: a fine texture shift under fast motion, an even brightness offset under occlusion, and redrawn leaves on the hardest prompt, never a broken object or a block-shaped artifact.
DecoupAttn leads SVG-EAR on every frame of both four-step cases, and on both UniRelight cases its frames exceed 37.6\,dB with SSIM $\geq0.99$ and match the dense frames.
These are the behaviors that decoupling predicts: the residuals of the fixed blocks cancel and spread (Section~\ref{sec:decoup}), and the clusters of each video follow its own content as objects move and occlude each other (Appendix~\ref{app:supp-gen}).

\noindent\textbf{Components at work.}
The cases show the three components of DecoupAttn in the frames.
Decoupling lets fixed exact blocks coexist with semantic clusters that cover every key, so the textures of Cases~1 and~3 change as spread noise, not at block borders (Section~\ref{sec:decoup}).
One offline plan per model sets the exact blocks of all four clips and each clip clusters its own keys, so a single frozen plan serves fast motion, occlusion, and dense texture, as it serves held-out content (Table~\ref{tab:shift}).
Distortion Budget matters on the four-step models of Cases~3 and~4, where removing its sensitivity costs 2.30 and 2.14\,dB on average (Table~\ref{tab:abl}).
On UniRelight, the model of Cases~1 and~2, reassignment gains 3.63\,dB over frozen assignments on test clips, the largest gain of the ten models (Table~\ref{tab:rule}), so the clusters follow the moving and occluding objects.

\noindent\textbf{Stability along the clip.}
The per-frame PSNR of DecoupAttn varies by at most 0.75\,dB within Cases~1--3 (38.44--38.87, 37.68--38.43, and 16.16--16.39\,dB), so its error does not build up from first to last frame, and in Case~4 it is highest after the first frame (20.94\,dB, frame 40).
SVG-EAR is as flat (39.88--40.24, 45.71--46.32, and 15.58--15.87\,dB), so the per-frame scores follow the prompt, not the position of the frame in the clip.

\noindent\textbf{Hard prompts.}
The prompts of Cases~3 and~4 are hard for every sparse method: SVG-EAR falls on them from its test means of 21.85 and 22.95\,dB to 15.85 and 18.24\,dB (Table~\ref{tab:main}), and DecoupAttn leads on both (16.39 and 19.95\,dB).
Their dense canopy of small leaves and their thin rails and struts are fine, repeated structures, and on them DecoupAttn keeps the layout, the people, and the vehicles of the dense video: the gardener and the goblin keep their poses and positions in every frame (Figures~\ref{fig:case3} and~\ref{fig:case4}).

\noindent\textbf{Structure and quality.}
DecoupAttn leads in SSIM on every frame of Cases~3 and~4, by 0.04--0.07, so it keeps more of the dense layout where both methods redraw fine detail, and its LPIPS is the lowest of all sparse methods on the three models of the cases (0.102, 0.057, and 0.006, Table~\ref{tab:main}).
VBench stays within 0.007 of dense attention on every model (Table~\ref{tab:full}), and the error against ground truth matches that of dense attention on both renderers that provide it (Table~\ref{tab:cond}), so the fidelity seen in the four cases holds across the test sets of all ten models.

\end{document}